\documentclass{article}
\usepackage{ijcai23}

\usepackage{times}
\usepackage{soul}
\usepackage{url}
\usepackage[hidelinks]{hyperref}
\usepackage[utf8]{inputenc}
\usepackage[small]{caption}
\usepackage{graphicx}
\usepackage{amsmath}
\usepackage{amsthm}
\usepackage{booktabs}
\usepackage{algorithm}
\usepackage{algorithmic}
\usepackage[switch]{lineno}
\usepackage{color}
\usepackage{amsfonts}
\usepackage{comment}
\usepackage{multirow}

\usepackage{graphicx}
\usepackage{subcaption}

\title{Adversary-Guided Motion Retargeting for Skeleton Anonymization}

\author{
Thomas Carr\and
Depeng Xu\and
Aidong Lu
\affiliations
University of North Carolina at Charlotte
\emails
\{tcarr23, depeng.xu, aidong.lu\}@charlotte.edu
}

\begin{document}

\maketitle

\begin{abstract}
        Skeleton-based motion visualization is a rising field in computer vision, especially in the case of virtual reality (VR). With further advancements in human-pose estimation and skeleton extracting sensors, more and more applications that utilize skeleton data have come about. These skeletons may appear to be anonymous but they contain embedded personally identifiable information (PII). In this paper we present a new anonymization technique that is based on motion retargeting, utilizing adversary classifiers to further remove PII embedded in the skeleton. Motion retargeting is effective in anonymization as it transfers the movement of the user onto the a dummy skeleton. In doing so, any PII linked to the skeleton will be based on the dummy skeleton instead of the user we are protecting. We propose a Privacy-centric Deep Motion Retargeting model (PMR) which aims to further clear the retargeted skeleton of PII through adversarial learning. In our experiments, PMR achieves motion retargeting utility performance on par with state of the art models while also reducing the performance of privacy attacks.
\end{abstract}

\section{Introduction}
The rapid advancement of virtual reality (VR) technologies has brought in a new era of digital interaction, where the boundaries between physical and virtual worlds are blending together. At the core of this advancement is skeleton-based data, a critical component for creating immersive and interactive VR experiences. Captured through motion capture systems like the Microsoft Xbox Kinect v2, this data translates the movement of users into digital avatars for various applications, playing a pivotal role in enhancing VR experiences \cite{TranWTRLP18,FanelloGMO13,SaggeseSVP19,GalANNS15}. However, the benefits of skeleton data in VR are accompanied by privacy concerns that have not been fully addressed, marking a crucial area for ongoing research.

Privacy in skeleton data is important due to the sensitive nature of the information it can reveal. Unlike traditional data types, skeleton data captures the unique physical movements and attributes of individuals. These attributes can include not only basic biometrics like height and limb length but also more intimate details like gait patterns and posture \cite{OladeFL20}. Such information could be used to infer personal characteristics, health conditions, and even emotional states \cite{s21010205}. For instance, variations in gait and posture could potentially indicate a person's age, physical fitness, or medical conditions such as arthritis or Parkinson's disease. Moreover, the distinctiveness of these attributes makes them highly valuable for identification purposes, raising concerns about unauthorized surveillance \cite{10.1145/3491101.3519645}, identity theft \cite{10.3389/frvir.2022.974652}, or discrimination \cite{linkage}.
Protecting the privacy of skeleton data is therefore critical not only for safeguarding personal identifiable information (PII) but also for maintaining trust in skeleton-based technologies.

\begin{figure}
    \centering
    \includegraphics[width=0.8\linewidth]{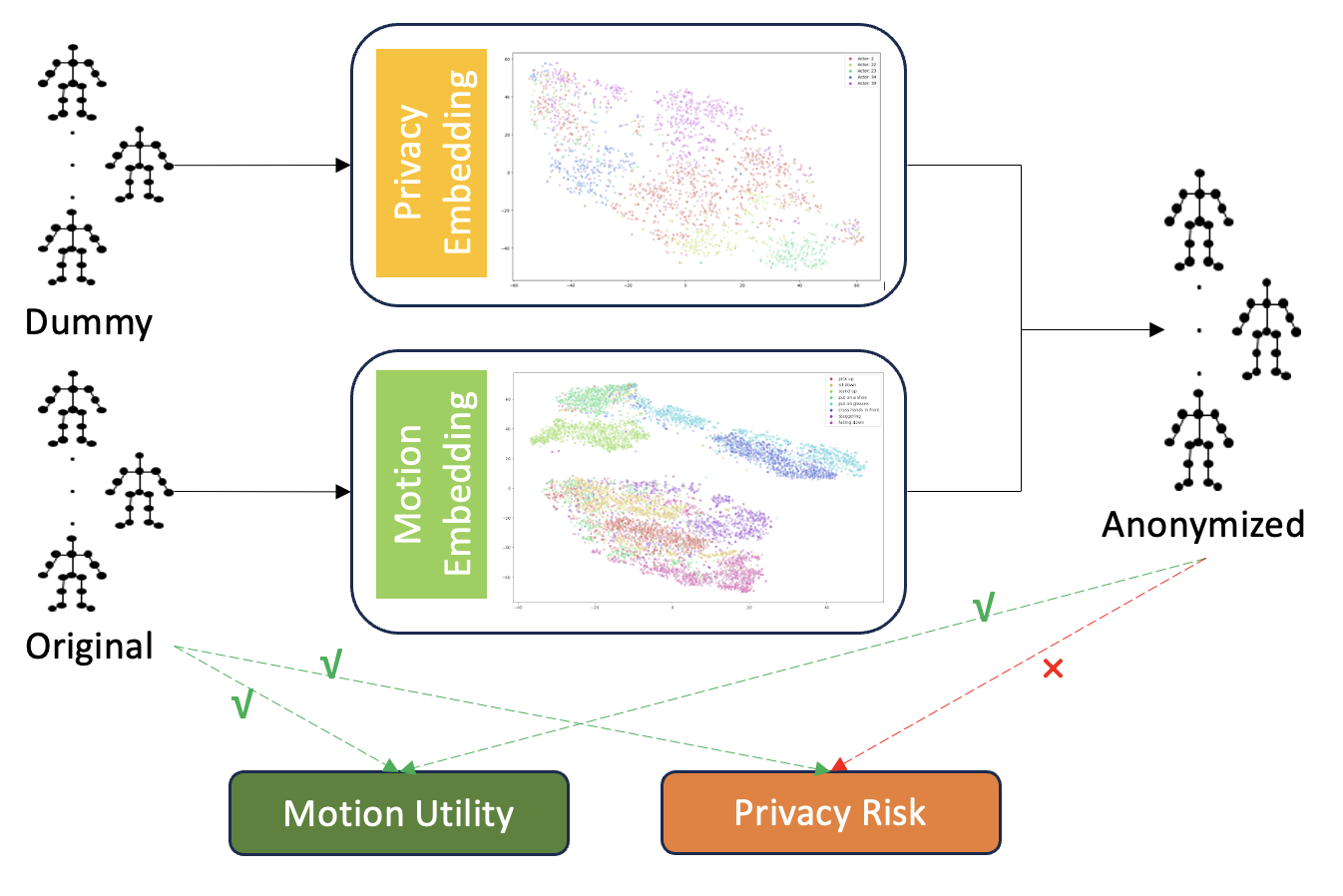}
    \caption{Motion retargeting for anonymization of skeleton data.}
    \label{fig:Intro}
\end{figure}

Motion retargeting involves transferring the motion data from one skeleton to another, effectively decoupling the actions from the original actor's skeletal structure \cite{AbermanWLCC19,AbermanLLSCC20}. This technique presents technical potential for privacy protection for skeleton-based data as it addresses the private information inherent in skeleton data such as height or wingspan. 
By casting actions from one skeleton onto another, motion retargeting may anonymize the data, removing identifiers that are tied to the specific physical attributes of the original subject. This process of retargeting not only preserves the utilities and nuances of the original motion but also ensures that the unique biometric signatures, specifically proportions specific to an individual, are not directly associated with the skeleton-data. Essentially, motion retargeting acts as a form of an anonymizer, creating a layer of abstraction between the individual and their actions. This makes it more challenging to reverse-engineer the data to identify the original subject, thereby enhancing privacy.

In this work, we explore privacy-centric deep motion retargeting for anonymization of skeleton data. The anonymized skeleton should maintain the motion utility but mitigate the privacy risk of re-identification. We use an autoencoder system to retarget the original skeleton onto a dummy skeleton. The generated skeleton keeps the character-agnostic motion from the original but replaces the private information with those of the dummy, thus it is anonymized. To make the deep motion retargeting model privacy-centric, we add motion and privacy classifiers to enhance the embedding representations learned by the encoders. One encoder extracts the motion embedding with no PII encoded,  which is passed to the anonymized skeleton. The other encoder extracts the privacy embedding with all PII of the original skeleton, which is removed and replaced by the dummy's. Using a real-world skeleton dataset, we evaluate our proposed model on motion utility and privacy risk of the anonymized data.

We summarize our contributions as follows:
(1) Developing a novel Privacy-centric Deep Motion Retargeting model (PMR) that removes the privacy information of the original data;
(2) Enhancing the representation learning on skeleton data with cooperative and adversarial learnings from a privacy-centric view;
(3) Demonstrating through experiments that PMR achieves state-of-the-art motion retargeting while effectively concealing PII.

\section{Related Works}
\noindent \textbf{Motion Retargeting.}
Recent advancements in motion retargeting address the challenges of adapting motion across diverse skeletal structures. 
The Neural Kinematic Networks for unsupervised motion retargeting \cite{NKN}, employs a recurrent neural network architecture with a Forward Kinematics layer, which captures the high-level properties of an input motion and adapts them to a target character with different skeleton bone lengths. They use the cycle consistency-based adversarial training to solve the Inverse Kinematics problem in an unsupervised manner. 
The work from \cite{AbermanWLCC19} extracts a high-level latent motion representation directly from 2D videos, which separates motion from the specific characteristics of the performer's skeleton, allowing for the retargeting of motion between different performers while remaining invariant to the skeleton geometry and camera view. 
In their later work, \cite{AbermanLLSCC20} further addresses the challenges of retargeting motion captured from individuals with varying skeletal structures. Their approach focuses on developing an abstract, character-and camera-agnostic, latent representation of human motion. 
These studies focus on perfectly aligned animation data. Whereas our work focuses on real-world motions and characters, which is more challenging, and we also add in privacy protection techniques.

\noindent \textbf{Privacy of Skeleton Data.}
The study by \cite{KimM2021} is the first to explore the field of privacy attack and anonymization on skeleton data. They propose actor re-identification and gender classification attacks, which demonstrates that skeleton data is susceptible to privacy leakages. 
Another work by \cite{linkage} proposes a linkage attack model based on the Siamese Network. It detects whether a target and a reference skeleton belong to the same individual.
The study by \cite{KimM2021} also uses adversarial learning for frame-by-frame anonymization. It modifies the skeleton data to confuse a personal ID classifier and a gender classifier while  maintaining the performance of an action recognition model, which limits its performance to these specific models used during training.
Their anonymization method is effective against privacy attacks at the cost of generalized motion utility. 
In our work, we focus on full video level anonymization through motion retargeting and evaluate on general motion utility with offline models.

\section{Methodology}
\subsection{Problem Statement}
A 3D skeleton $\mathbf{s}\in\mathbb{R}^{J \times T \times 3}$ captures the human motion with 3D coordinates $\mathbf{s} = (x_{i,j},y_{i,j},z_{i,j})^{J\times T}$ of $J$ joins over $T$ frames. The skeleton data is visualized in VR so the motion action information can be recognized. The skeleton can be used to predict PII indicators such as age and gender. The attacker can use public skeleton data to train models to predict and misuse these seemingly anonymous attributes. Privacy retargeting will move the motion information from the original skeleton to a dummy skeleton to hide the PII attributes. The goal of privacy retargeting is to successfully transition the motion such that the motion utility of the skeleton is still present, but the PII indicators can no longer be predicted.

\begin{figure}
    \centering
    \includegraphics[width=0.85\linewidth]{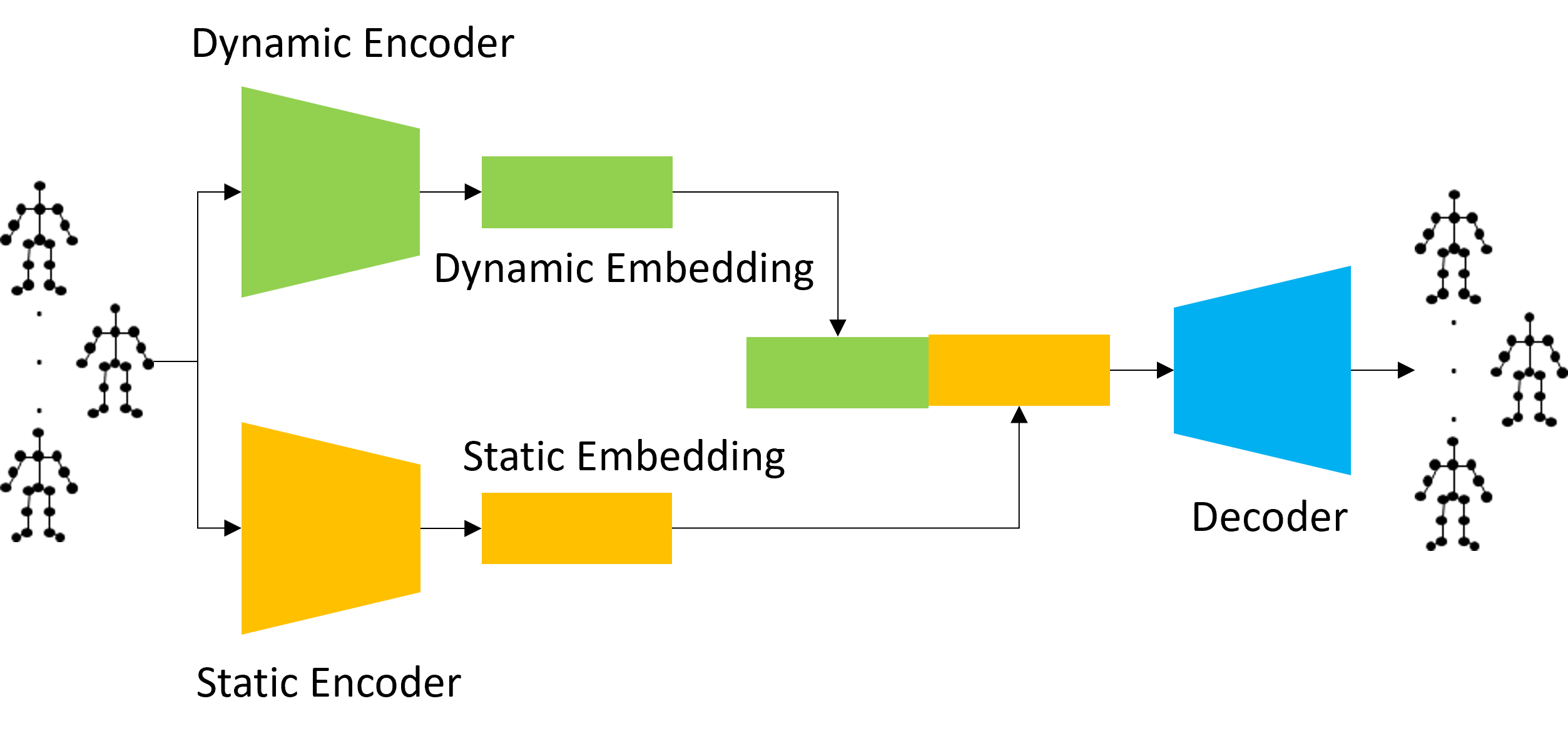}
    \caption{The architecture of DMR}
    \label{fig:DMR}
\end{figure}

\begin{figure}
    \centering
    \includegraphics[width=\linewidth]{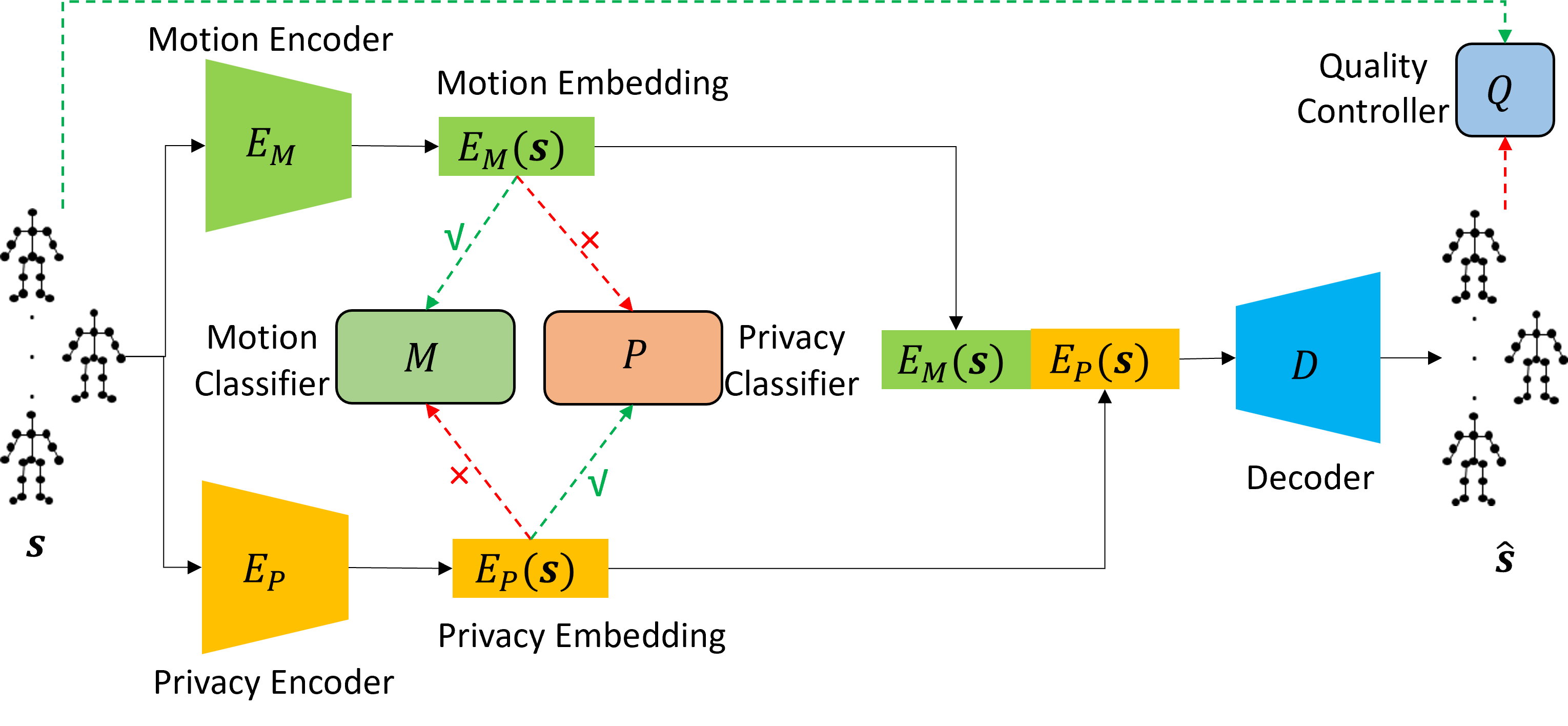}
    \caption{The architecture of PMR in the unpaired setting.}
    \label{fig:PMR}
\end{figure}

\begin{figure}
    \centering
    \includegraphics[width=0.95\linewidth]{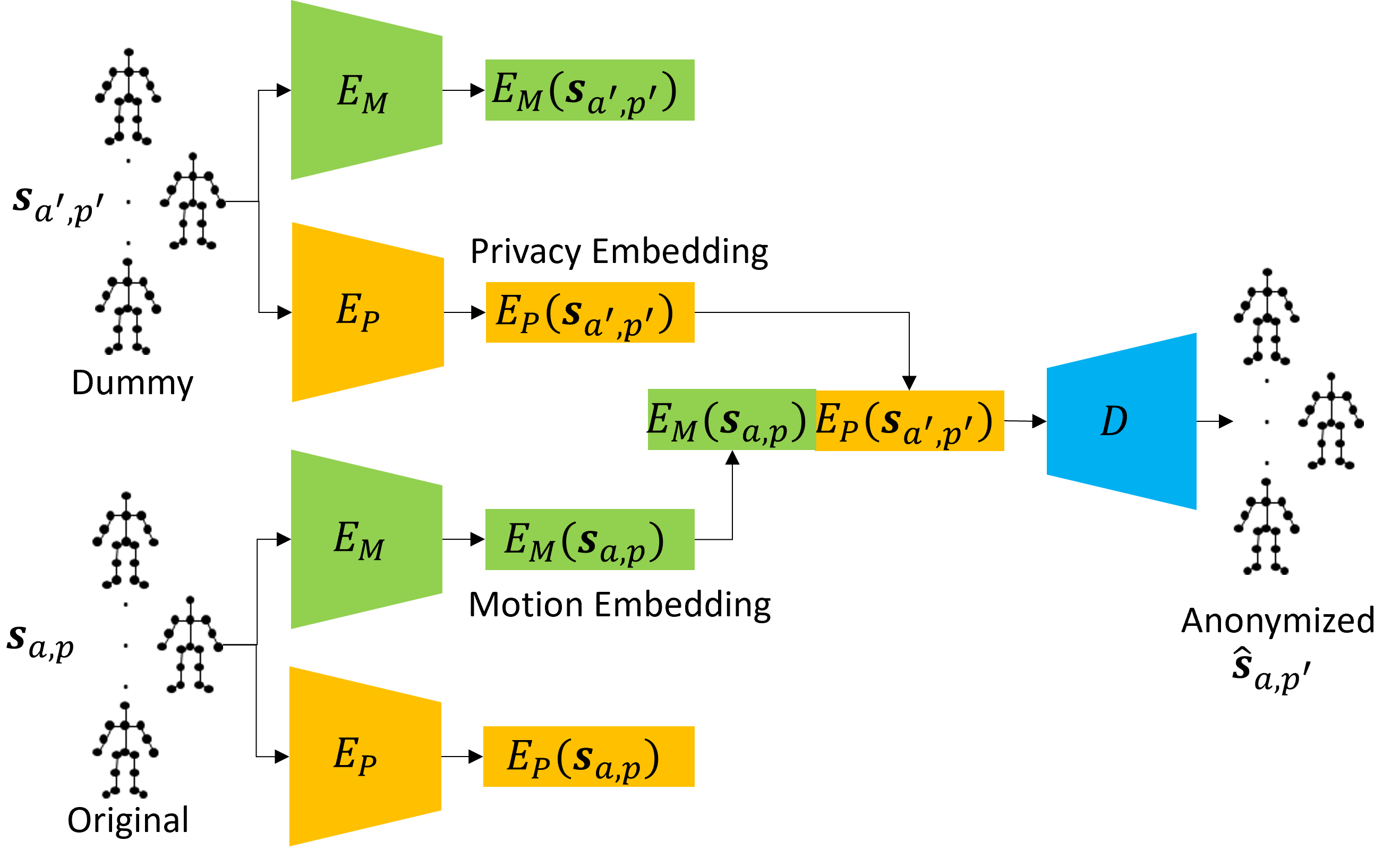}
    \caption{The architecture of PMR for anonymization in the paired setting. The classifiers are not shown.}
    \label{fig:cross}
\end{figure}

\subsection{Overview}

To achieve the goal of privacy in skeleton-based data, we propose a Privacy-centric Deep Motion Retargeting model (PMR). PMR is developed based on the deep motion retargeting model (DMR) introduced in \cite{AbermanWLCC19} and \cite{AbermanLLSCC20}. In DMR, a multi-encoder/single-decoder neural network is trained to decompose and recompose the temporal sequences of joint positions, as shown in Figure \ref{fig:DMR}. One encoder encodes the input into a dynamic (duration-dependent) motion representation. The other encodes into a static (duration-independent) skeleton representation. The decoder recomposes the extracted motion representation and skeleton representation from the same input (or different inputs) to reconstruct (or cross-reconstruct/retarget) the motion. We re-purpose DMR for privacy protection and add several modifications to enhance the quality of the representation learning and skeleton anonymization. 

The proposed PMR architecture is shown in Figure \ref{fig:PMR}. PMR also has two encoders and one decoder. The motion encoder $E_M$ encodes the temporal information about the action being performed into motion embedding $E_M(\mathbf{s})$. The privacy encoder $E_M$ encodes all PII including the size of the skeleton, gait, and the how motions are performed by that actor into privacy embedding $E_M(\mathbf{s})$.
The decoder $D$ recomposes the motion sequence from the concatenated motion and privacy embeddings. 
With the goal of privacy, we employ two classifiers on both of the embeddings with the purpose of working as both adversaries and cooperatives for different encoders. 
The motion classifier $M$ is trained to predict the action label from the motion embedding $E_M(\mathbf{s})$, while the privacy classifier $P$ is trained to predict the actor ID from the privacy embedding $E_M(\mathbf{s})$. We also add a quality controller $Q$, which operates as a real/fake discriminator, at the end to ensure the generated skeleton by decoder is realistic.

\subsection{Unpaired vs. Paired}
In our approach to motion retargeting, we employ both paired and unpaired settings. In the unpaired setting, shown in Figure \ref{fig:PMR}, a single input sample $\mathbf{s}$ serves as the foundation for the initial training of the auto-encoder component of our model. The encoders decompose the input sequence into two distinct latent embeddings. The decoder reconstructs the original input from these embeddings. This phase is critical for establishing a baseline understanding of motion and skeleton data. The unpaired setting provides a diverse range of motion patterns for the model to learn from. Once the auto-encoder is sufficiently trained on the unpaired setting, we introduce the paired setting to fine-tune the model for motion retargeting.

In the paired setting, shown in Figure \ref{fig:cross}, the model is trained with carefully matched sets of skeleton motions. This setting is designed to provide the model with a clear understanding of how motion can be transferred between different skeletons while preserving the action's essence. We construct paired samples where two distinct actors ($p$ and $p'$) perform the same two actions ($a$ and $a'$) under an identical camera view. 
This strategic pairing is essential for our model to learn subtleties of motion transfer while maintaining the integrity of the action. The paring scheme ensures that each actor has done both actions all under the same camera view.
This configuration is a fundamental part of motion retargeting, providing the model with rich, comparative data, essential to understanding how differing skeletal structures can perform the same actions differently.

The paired setting is used for anonymizing the skeleton data. The original skeleton $\mathbf{s}_{a,p}$ is paired with a dummy skeleton  $\mathbf{s}_{a',p'}$ as inputs. Both $\mathbf{s}_{a,p}$ and $\mathbf{s}_{a',p'}$ are decomposed into motion embeddings $E_M(\mathbf{s}_{a,p}), E_M(\mathbf{s}_{a',p'})$ and privacy embeddings $E_M(\mathbf{s}_{a,p}), E_M(\mathbf{s}_{a',p'})$. The decoder takes the original skeleton's motion embedding $E_M(\mathbf{s}_{a,p})$ and the dummy skeleton's privacy embedding $ E_M(\mathbf{s}_{a',p'})$ to cross-reconstruct an anonymized skeleton sequence $\hat{\mathbf{s}}_{a,p'}$.

\subsection{Encoders}
The architecture of our model comprises two distinct encoders, each playing a critical role in the process of motion retargeting. The privacy encoder $E_P$, and the motion encoder $E_M$, are designed to capture different aspects of the skeleton data. 
Their implementations are identical and involve a series of convolutional layers, each followed by activation and pooling layers to progressively refine the input data into meaningful feature representations. These layers are designed to handle the intricacies of motion data, ensuring that crucial details are not lost in the encoding process. The final output of each encoder are motion and privacy embeddings which encapsulate the characteristics of the skeleton's motion and PII features. The design of embedding classifiers and loss functions ensures these embeddings are distinct  for their purposes.

\subsection{Decoder}
The decoder, represented as $D$, takes the concatenated motion embedding $E_M(\mathbf{s})$ and privacy embedding $E_M(\mathbf{s})$ as inputs and reconstructs the skeleton motion. The decoder, consisting of a series of transposed convolutional layers, effectively reverses the encoding process. It gradually upsamples and refines the combined embeddings, reconstructing the motion in a new context. The output of the decoder, denoted as $\hat{\mathbf{s}}$, represents the generated skeleton motion sequence, i.e., the reconstructed skeleton in the unpaired setting or the cross-reconstructed skeleton in the paired setting.  The generated motion maintains the original action utility but applied to a different skeletal structure. The cross-reconstructed motion is the essence of our privacy-centric approach, ensuring that the identity of the individual is obscured while preserving the integrity of the motion itself.

\subsection{Embedding Classifiers}
The embedding classifiers in our model play a pivotal role in enhancing privacy while preserving the utility of motion retargeting. We have two embedding classifiers: motion ($M$) and privacy ($P$). The motion classifier $M$ aims to predict action classes from the embeddings. The privacy classifier $P$ aims to predict the actor's identity from the embeddings. 

These classifiers engage in a min-max game with the encoders, utilizing adversarial and cooperative training to enhance the representation learning on the skeleton data. 
The motion classifier $M$ ensures that despite anonymization, the action's essential characteristics are preserved in the utility embedding. It provides guidance to the motion encoder $E_M$ to ensure that essential motion information is retained in $E_M(\mathbf{s})$. It works as an adversary to the privacy encoder $E_P$ to ensure $E_P(\mathbf{s})$ has no motion utility information. In contrast, the privacy classifier $P$ works to strip away identifiable traits from the privacy embedding, focusing on aspects like skeletal structure and size that could be linked to specific individuals. It provides guidance to the privacy encoder $E_P$ to ensure all PII is extracted into $E_P(\mathbf{s})$. It works as an adversary to the motion encoder $E_M$ to ensure $E_M(\mathbf{s})$ has no PII, which can be passed along to the anonymized skeleton. 

Unlike DMR, the embedding classifiers in PMR provide enhancement on representation learning to achieve distinct decomposed embeddings from the encoders. One has all the private information that requires protection and the other has all the nonsensitive information that can be shared freely. 
This approach creates a balanced tug-of-war, where the utility of the data is maintained for accurate motion retargeting, while personal identifiers are obscured to protect privacy. The training involves an iterative back-and-forth adjustment process, with each one refining its strategies based on the other's performance. This method ensures that neither aspect—utility nor privacy—dominates, leading to an effective equilibrium where both objectives are met. This is integral to our model's ability to navigate the balance between maintaining the integrity and utility of motion data and ensuring the anonymity and privacy of the individuals represented in the skeleton data.

\subsection{Quality Controller}
The quality controller $Q$ in our model is a discriminator inspired by the principles of Generative Adversarial Networks (GANs) \cite{gan}. The quality controller's goal is to differentiate whether a skeleton is real or generated. This binary classification plays a crucial role in refining the quality of the generated skeletons, ensuring they are indistinguishable from real ones. By continuously attacking the rest of the PMR model, it increases its ability to generate realistic skeleton data. 
This is especially useful when working with real-world imperfect data.

\subsection{Training and Loss}
Due to the complexity of the model, we break the training down into 4 stages: (1) Pre-training the Auto-Encoder ($E_M$, $E_M$, $D$), (2) Pre-training the Embedding Classifiers ($M$, $P$), (3) Unpaired Training, and (4) Paired Training.

\subsubsection{Pre-training the Auto-Encoder}

We first pre-train the auto-encoder in the unpaired setting for reconstruction. 
The encoders $E_M, E_P$ take $\mathbf{s}$ as input and decompose it into two distinct latent spaces. The decoder $D$ recompose the embeddings $E_M(\mathbf{s}), E_P(\mathbf{s})$ into a reconstructed skeleton $\hat{\mathbf{s}} $.
\begin{equation}
\nonumber
    \hat{\mathbf{s}} = D(E_M(\mathbf{s}), E_P(\mathbf{s})).
    \label{eq:ae}
\end{equation}

The goal of reconstruction is to make it as close to the original skeleton as possible, following the \textbf{reconstruction loss} $L_{rec}$ below.
\begin{equation}
\nonumber
    L_{rec} = \mathbb{E}_{\mathbf{s} \sim \mathcal{S}} \left[ \left\| D(E_M(\mathbf{s}), E_P(\mathbf{s})) - \mathbf{s} \right\|^2 \right].
    \label{eq:loss_rec}
\end{equation}

On top of the reconstruction loss, we add a \textbf{smooth loss}  $L_{smooth}$ to improve temporal consistency of the skeleton motion. The position shift between frames for each joint should be consistent between the original skeleton and the generated skeleton. This prevents the sequence from jumping around or stuttering.  

\resizebox{.8\linewidth}{!}{
\begin{minipage}{\linewidth}
\begin{equation}
\nonumber
    L_{smooth} = \mathbb{E}_{\mathbf{s} \sim \mathcal{S}} \left[ \dfrac{ \sqrt{ \sum_{i}^{J} \left| \sum_{j}^{T} \left( \hat{s}_{i,j} - \hat{s}_{i,j+1} \right)^2 - \sum_{j}^{T} \left( \mathbf{s}_{i,j} - \mathbf{s}_{i,j+1} \right)^2 \right| }}{J \times T} \right],
\end{equation}
\end{minipage}
}

\noindent where $J$ is the number of joints and $T$ is the number of frames.

The total loss for the autoencoder pre-training is 
\begin{equation}
\nonumber
    L_{ae} = \alpha_{rec} L_{rec} + \alpha_{smooth} L_{smooth},
    \label{eq:loss_ae}
\end{equation}
where $\alpha_{rec}, \alpha_{smooth}$ are hyperparameters.

\subsubsection{Pre-training the Embedding Classifiers}

After the autoencoder has a base performance, we add in the embedding classifiers $M, P$ for pre-training. The encoders $E_M,E_P$ are fixed for this stage. 

The motion classifier $M$ is trained to predict action label $a$ from both embeddings. For $M$ to predict $a$ from the motion embedding, it helps with the cooperative training between $M$ and $E_M$ in the later stages. For $M$ to predict $a$ from the privacy embedding, it helps with the adversarial training between $M$ and $E_P$ in the later stages. 

The \textbf{motion classifier loss}  $L_{M}$ to train $M$ is 
\begin{align}
    L_{M} = \mathbb{E}_{\mathbf{s}_{a,p}\sim \mathcal{S}} \bigg[  CE \left(M(E_M\left(\mathbf{s}_{a,p}\right)), a \right) + \nonumber \\
     CE \left(M(E_P\left(\mathbf{s}_{a,p}\right)), a \right) \bigg], \nonumber
    \label{eq:loss_M}
\end{align}
where CE denotes cross entropy. $E_M$ will  cooperate with $M$ by minimizing the first term. $E_P$ will try to fool $M$ by maximizing the second term.

Similarly, the privacy classifier $P$ is trained to predict actor ID $p$ from both embeddings. For $P$ to predict $p$ from the privacy embedding, it helps with the cooperative training between $M$ and $E_P$ in the later stages. For $M$ to predict $p$ from the privacy embedding, it helps with the adversarial training between $M$ and $E_M$ in the later stages. 

The \textbf{privacy classifier loss}  $L_{P}$ to train $P$ is 
\begin{equation}
    \begin{aligned}
    L_{P} = \mathbb{E}_{\mathbf{s}_{a,p}\sim \mathcal{S}} \bigg[  CE \left(P(E_M\left(\mathbf{s}_{a,p}\right)), p \right) + \nonumber \\
     CE \left(P(E_P\left(\mathbf{s}_{a,p}\right)), p \right) \bigg]. \nonumber 
\end{aligned}
\label{eq:loss_P}
\end{equation}
$E_P$ will  cooperate with $P$ by minimizing the first term. $E_M$ will try to fool $P$ by maximizing the second term.

In this stage, we also set up pre-training for the quality controller $Q$. $Q$ aims to predict whether a skeleton is authentic or generated. It predicts the original skeleton $\mathbf{s}$ as 1 and the generated skeleton $\hat{\mathbf{s}}$ as 0. The \textbf{quality controller loss} $L_{qc}$ is
\begin{equation}
    L_{qc} = \mathbb{E}_{\mathbf{s} \sim \mathcal{S}} \left[ \log Q(\mathbf{s}) + \log \left( 1-  Q \left( D(E_M(\mathbf{s}), E_P(\mathbf{s})) \right) \right) \right].
    \label{eq:loss_discrim}
    \nonumber
\end{equation}
The decoder $D$ will try to fool $Q$ by maximizing the second term.

\subsubsection{Unpaired Training}
After all network components are warmed up with pre-training, we start training the whole model in the unpaired  setting. 
In this stage, we enhance the quality of representation learning by $E_M,E_P$  through cooperative and adversarial learning with the embedding classifiers $M,P$. We also enhance the quality of generated skeletons by $D$ though adversarial learning with the quality controller $Q$.

We iteratively train the autoencoder components $E_M,E_P,D$ and the classifiers $M,P,Q$. The latter is fixed while the former is training, and vice versa. 

For the \textbf{cooperative training}, the motion encoder $E_M$ cooperates with $M$ to improve the quality of motion embedding $E_M(\mathbf{s})$ on extracting action information. The privacy  encoder $E_P$ cooperates with $P$ to improve the quality of privacy embedding $E_P(\mathbf{s})$ on extracting PII.  
The autoencoder training loss adopts a new component, \textbf{cooperative loss} $L_{coop}$.
\begin{align}
    L_{coop} = \mathbb{E}_{\mathbf{s}_{a,p}\sim \mathcal{S}} \bigg[  CE \left(M(E_M\left(\mathbf{s}_{a,p}\right)), a \right) + \nonumber \\
     CE \left(P(E_P\left(\mathbf{s}_{a,p}\right)), p \right) \bigg]. \nonumber
    \label{eq:loss_coop}
\end{align}

For the \textbf{adversarial training}, the motion encoder $E_M$ plays a minimax game with $P$ to ensure the motion embedding $E_M(\mathbf{s})$ contains no PII. The privacy  encoder $E_P$ plays a minimax game with $M$ to ensure there is no motion utility information left in the privacy embedding $E_P(\mathbf{s})$.  Meanwhile, the decoder $D$ plays an adversarial game with $Q$ to ensure the generated skeletons $\hat{\mathbf{s}}$ are indistinguishable from the real skeletons.
Thus, the autoencoder training loss adopts another component, \textbf{adversarial loss} $L_{adv}$.
\begin{align}
    L_{adv} = \mathbb{E}_{\mathbf{s}_{a,p}\sim \mathcal{S}}\bigg[ - 
     CE \left(M(E_P\left(\mathbf{s}_{a,p}\right)), a \right)  \nonumber \\
     - CE \left(P(E_M\left(\mathbf{s}_{a,p}\right)), p \right) \nonumber \\
     - \log \left( 1-  Q \left( D(E_M(\mathbf{s}), E_P(\mathbf{s})) \right) \right) \bigg], \nonumber
\end{align}
where the first two terms are for the encoders $E_M, E_P$ and the third term is for the decoder $D$.

The total loss for the autoencoder unpaired training is 
\begin{align}
    L_{unpaired} &=  L_{ae} + \alpha_{emb} L_{coop} + \alpha_{emb} L_{adv},  \nonumber
\end{align}
where $\alpha_{emb}$ is a hyperparameter.

The embedding classifiers $M,P$ and quality controller $Q$ have the same loss in the iterative training as pre-training.

\subsubsection{Paired Training}
After the unpaired training stage, the two encoders decompose the original skeleton into two distinct latent representations. The PII and motion information are completely separated. We transit into the paired setting for motion retargeting. 

In the paired training, the model takes two skeletons as input, where $\mathbf{s}_{a,p}$ serves as the original skeleton and $\mathbf{s}_{a',p'}$ serves as the dummy skeleton. The goal is to retarget the action of the original skeleton to the dummy skeleton without a trace of the original actor's PII. 
The retarget skeleton $\hat{\mathbf{s}_{a,p'}} $ is the anonymized output of PMR.
\begin{equation}
\nonumber
    \hat{\mathbf{s}_{a,p'}} = D(E_M(\mathbf{s}_{a,p}), E_P(\mathbf{s}_{a',p'})).
\end{equation}

The autoencoder performs cross-reconstruction, where the \textbf{cross-reconstruction loss} $L_{cross}$ is \\
\resizebox{0.98\linewidth}{!}{
\begin{minipage}{\linewidth}
\begin{equation}
    L_{cross} = \mathbb{E} _{\mathbf{s}_{a,p},  \mathbf{s}_{a',p'}\sim \mathcal{S}}
             \left\| D(E_M(\mathbf{s}_{a,p}), E_P(\mathbf{s}_{a',p'})) - \mathbf{s}_{a,p'} \right\|^2.
    \label{eq:loss_cross}
    \nonumber
\end{equation}
\end{minipage}
}

In this stage, we take advantage of the paired setting to add several components to improve the quality of representation learning and retargeted skeleton generation.

To further ensure the distinctness of embeddings, we add a \textbf{triplet loss} $L_{trip}$ to explicitly require the separation of two latent spaces encoded in $E_M, E_P$. It ensures motion embeddings are close in an embedding space while privacy embeddings are close in a distant embedding space. 

\resizebox{.96\linewidth}{!}{
\begin{minipage}{\linewidth}
\begin{align}
    L_{trip} &= \mathbb{E}_{\mathbf{s}_{a,p},  \mathbf{s}_{a',p'}\sim \mathcal{S}} \bigg[ \nonumber \\
        &\quad \max \left( 0, \left\| E_M(\mathbf{s}_{a,p}) - E_M(\mathbf{s}_{a',p}) \right\|^2 - \left\| E_M(\mathbf{s}_{a,p}) - E_M(\mathbf{s}_{a,p'}) \right\|^2 + \gamma \right) + \nonumber \\
        &\quad \max \left( 0, \left\| E_P(\mathbf{s}_{a,p}) - E_P(\mathbf{s}_{a,p'}) \right\|^2 - \left\| E_P(\mathbf{s}_{a,p}) - E_P(\mathbf{s}_{a',p}) \right\|^2 + \gamma \right) \bigg] ,
    \nonumber
\end{align}
\end{minipage}
}
\\
where $\gamma$ is the triplet loss margin.

We also add \textbf{latent consistency loss} $L_{latent}$ to ensure the same action by different actors has consistent motion embedding in the latent space. It also ensures the same actor performing different actions has consistent privacy embedding in the latent space. 

\resizebox{.95\linewidth}{!}{
\begin{minipage}{\linewidth}
\begin{align}
    L_{latent} = \mathbb{E}_{\mathbf{s}_{a,p},  \mathbf{s}_{a',p'}\sim \mathcal{S}} \bigg[ MSE \left(E_M\left(\mathbf{s}_{a,p}\right), E_M\left({\mathbf{s}}_{a,p'}\right) \right) + \nonumber \\
     MSE \left(E_P\left(\mathbf{s}_{a,p}\right), E_P\left({\mathbf{s}}_{a',p}\right) \right) \bigg],
    \nonumber
\end{align}
\end{minipage}
}\\
where MSE denotes mean squared error.

To handle the skeleton spacial size difference, we add a \textbf{end-effectors loss} $L_{ee}$ to the generated skeleton. It requires the end-effectors of the original and the retargeted  skeleton to have the same normalized velocity. It prevents foot sliding and ensures steady positioning when the original skeleton is still.  
\begin{equation}
    L_{ee} = \mathbb{E}_{\mathbf{s} \sim \mathcal{S}} \sum_{e \in E} \left\| \frac{V_{\mathbf{s}_e} - V_{\hat{\mathbf{s}}_e}}{h_{e}}\right\|^2,
    \label{eq:loss_ee}
    \nonumber
\end{equation}
where $E$ is the set of end-effectors, $h_e$ is the length of the kinematic chain for end-effector $e$, $V_{s_e}$ is the velocity of skeleton's end-effector $e$ kinematic chain.

The total loss for the autoencoder paired training is 
\begin{align}
    L_{paired} &= L_{unpaired} + \alpha_{cross} L_{cross} + \nonumber \\
    & \alpha_{ee} L_{ee} + \alpha_{trip} L_{trip} + \alpha_{latent} L_{latent},
    \nonumber
\end{align}
where $\alpha_{cross},\alpha_{ee}, \alpha_{trip} ,\alpha_{latent}$ are hyperparameters.

\section{Experiment Setup}
In the experiment section we explain the implementation of the PMR model and demonstrate its performance against both motion retargeting and anonymization models.

\subsection{Implementation Details}

\begin{table}[]
\small
\centering
\begin{tabular}{|l|l|}
\hline
Hyperparameter    & Value \\ \hline
$\alpha_{rec}$    & 2     \\ \hline
$\alpha_{cross}$  & 0.1   \\ \hline
$\alpha_{ee}$     & 1     \\ \hline
$\alpha_{trip}$   & 1     \\ \hline
$\alpha_{smooth}$ & 3     \\ \hline
$\alpha_{latent}$ & 10    \\ \hline
$\alpha_{emb}$    & 10    \\ \hline
\end{tabular}
\caption{Hyperparameters}
\label{tbl:hyper}
\end{table}

Our model was implemented in PyTorch and experiments were conducted on a PC with a NVIDIA GeForce RTX 3090 Ti (24GB) and an AMD Ryzen 7 5800k CPU (64GB RAM). 
The values of hyperparameters are shown in Table \ref{tbl:hyper}.
We utilize the Adam optimizer for the auto-encoder portion, the embedding classifiers, and the discriminator.

\subsubsection{Dataset}
In our experiment we utilize the NTU RGB+D 60 dataset from \cite{ntu60}. It consists of 60 action classes and 40 actors. This dataset contains skeleton data captured from the Microsoft Kinect v2. Each skeleton is captured at 30fps and consists of 25 joints. For the purpose of our experiment, we only focus on the XYZ coordinates of each joint. 
We denoise the raw skeleton data as well as remove the skeleton files that contain poor data. 
Since our main focus is on retargeting a single actor, we remove any files that contain two actors. This removes 11 action classes. The dataset provides 3 camera views. We use the first camera view for evaluations, and the other two for training. Due to the varying frame count of all of the data, we find the sweet spot to be around the average frame length of $T=75$. For all recordings longer than 75 frames, we cut the data. For all recordings shorter, we repeat the final frame until it reaches $T$.

A major difference in our experiment versus those in standard motion retargeting is that our data is imperfect. Other datasets like Mixamo, as used in \cite{AbermanWLCC19} and \cite{AbermanLLSCC20}, are great for motion retargeting training as all the actors complete actions in the exact same way.

\subsubsection{Metrics}
We evaluate the anonymized skeletons on two aspects. It should keep the maximum motion utility while avoiding any potential privacy risk. 

For motion utility, we use mean squared error (MSE) for quantitative analysis and example visualization for qualitative analysis. 

For privacy risk, we adapted an action recognition model to perform a privacy attack on skeletons. By changing the label from action to actor, Semantic-Guided Neural Network (SGN) \cite{sgn}, was able to achieve 85\% top-1 accuracy and 95.87\% top-5 accuracy on re-identifying who was completing an action in the original skeleton. We use this same model to evaluate the privacy risk of the anonymized skeletons.

\subsubsection{Baselines}

We compare our PMR model against the frame-level anonymization models (\textbf{UNet} and \textbf{ResNet}) proposed by \cite{KimM2021} for motion utility  and privacy risk. We also compare with \textbf{DMR} \cite{AbermanWLCC19,AbermanLLSCC20}\footnote{These models only run on Mixamo dataset. We implemented a modified version to perform evaluation on the NTU dataset.}.

For both DMR and PMR, to evaluate the anonymization performance, we run two experiments, where one uses a constant dummy skeleton for all test samples, and the other uses a random dummy skeleton for each test sample.

\subsection{Anonymization Results}

\begin{table}[]

\centering
\begin{tabular}{|l|l|ll|}
\hline
             & Utility & \multicolumn{2}{c|}{Re-Identification} \\ \hline
Method             & MSE     & \multicolumn{1}{l|}{Top-1}   & Top-5   \\ \hline
Original          & -       & \multicolumn{1}{l|}{0.8500} & 0.9587 \\ \hline
UNet         & 0.0834 & \multicolumn{1}{l|}{0.0303} & 0.2666  \\ \hline
ResNet       & 0.2988 & \multicolumn{1}{l|}{0.0897} & 0.3406 \\ \hline
DMR Constant & 0.0072 & \multicolumn{1}{l|}{0.1926} & 0.5066 \\ \hline
DMR Random   & 0.0067 & \multicolumn{1}{l|}{0.2006} & 0.5057 \\ \hline
PMR Constant & 0.0120 & \multicolumn{1}{l|}{0.0648} & 0.2451 \\ \hline
PMR Random   & 0.0109 & \multicolumn{1}{l|}{0.1202} & 0.3645 \\ \hline
\end{tabular}
\caption{Motion utility and privacy risk comparison}
\label{tab:main}
\end{table}

Table \ref{tab:main} shows the motion utility and privacy risk comparison across different models. UNet  and ResNet reduce the success rate of the re-identification attack. However, the anonymized skeleton has a relatively larger MSE. DMR has the lowest MSE but it still contains PII traits of the original actor. The re-identification attack still has a relatively higher success rate on DMR. Our PMR retargeting works the best as an anonymization method. The MSE is very low, close to DMR, which indicates high motion utility. The re-identification attack has low accuracy on PMR, indicating low privacy risk.

\subsection{Qualitative Analysis}
\begin{figure}[!tb]
    \centering

    \begin{subfigure}{\linewidth}
        \centering
        \includegraphics[width=.24\linewidth]{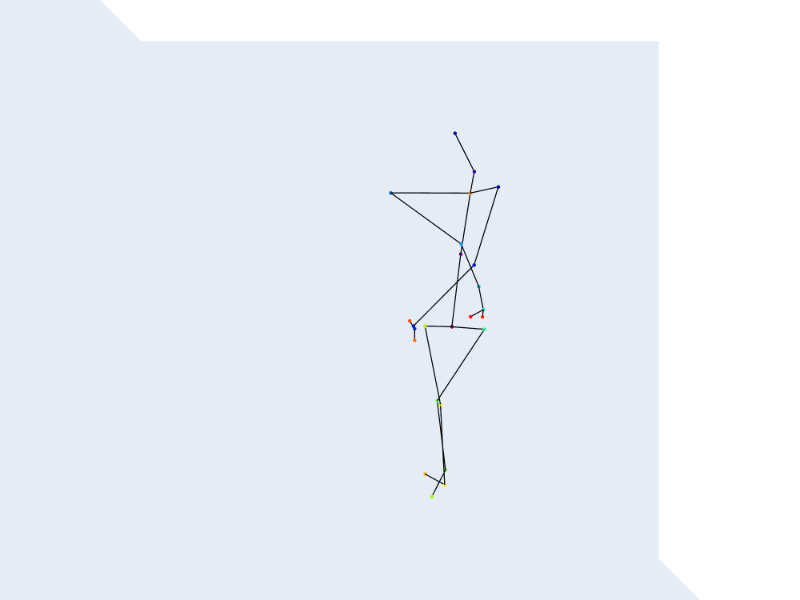}
        \includegraphics[width=.24\linewidth]{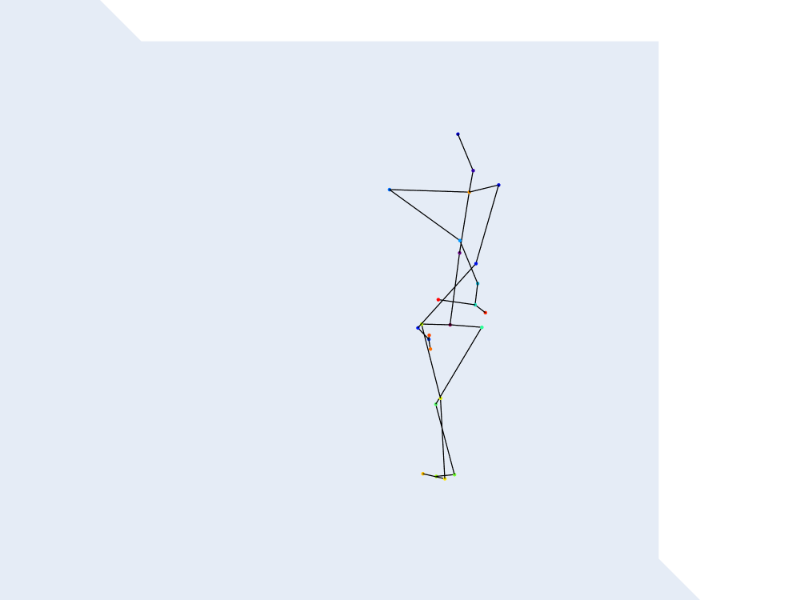}
        \includegraphics[width=.24\linewidth]{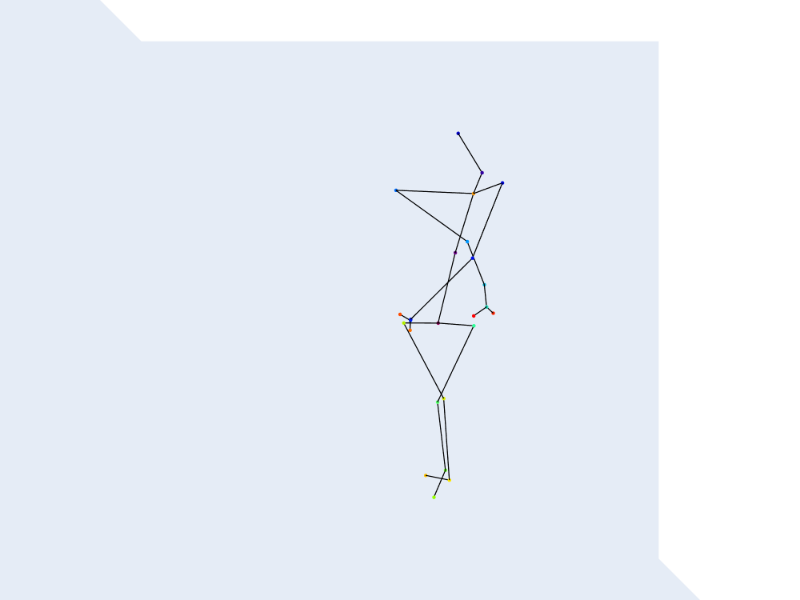}
        \includegraphics[width=.24\linewidth]{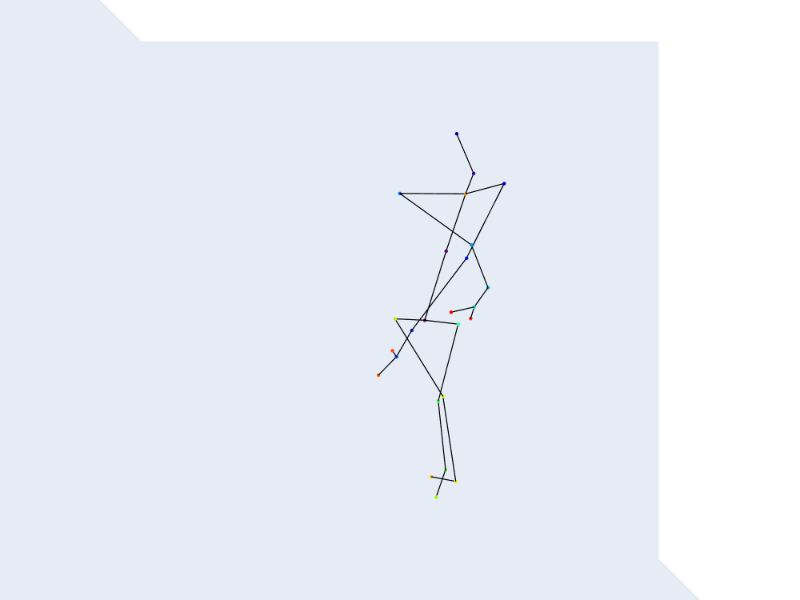}
        \caption{Original}
    \end{subfigure}

    \begin{subfigure}{\linewidth}
        \centering
        \includegraphics[width=.24\linewidth]{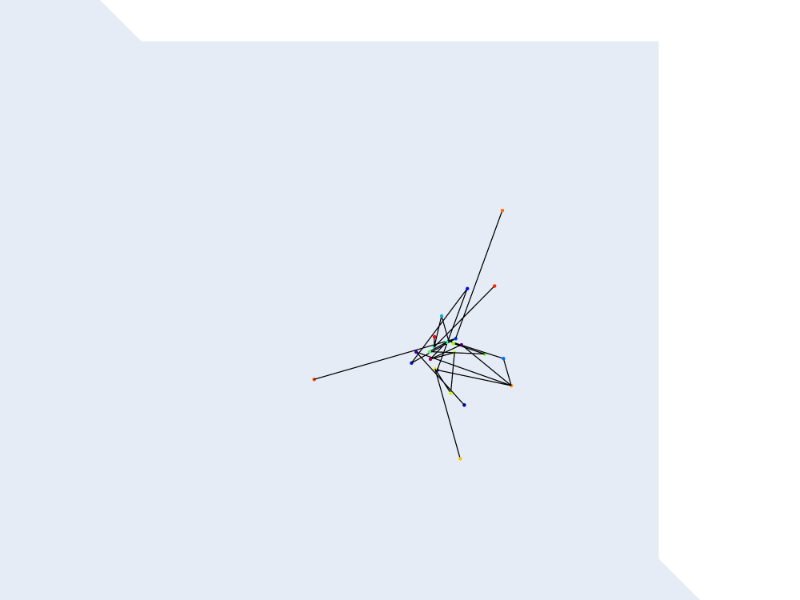}
        \includegraphics[width=.24\linewidth]{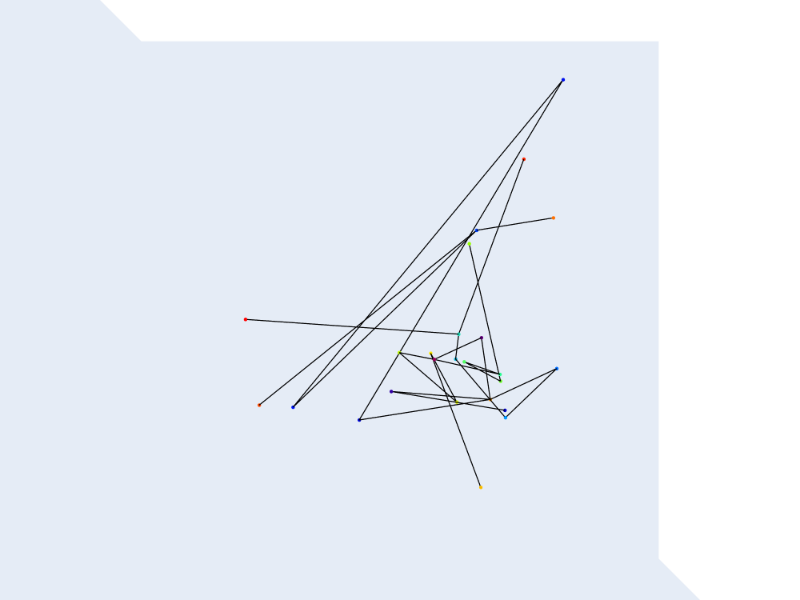}
        \includegraphics[width=.24\linewidth]{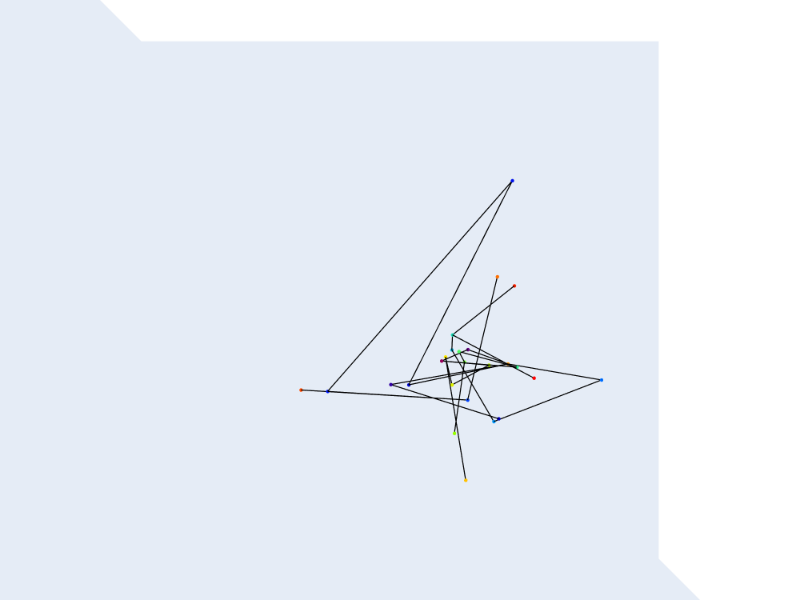}
        \includegraphics[width=.24\linewidth]{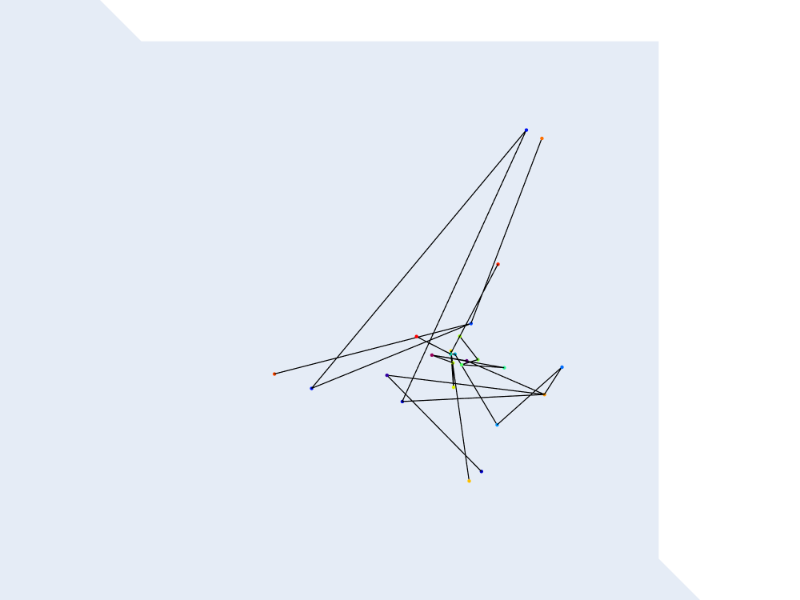}
        \caption{UNet}
    \end{subfigure}

    \begin{subfigure}{\linewidth}
        \centering
        \includegraphics[width=.24\linewidth]{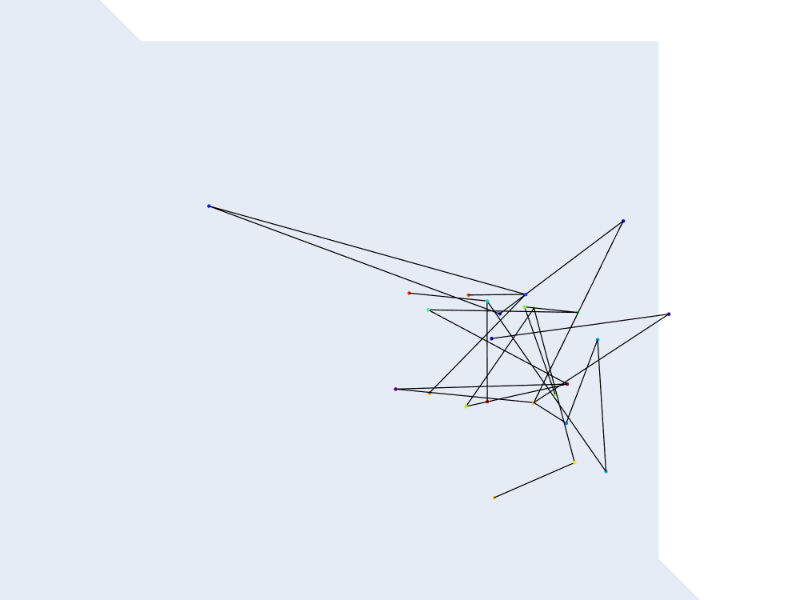}
        \includegraphics[width=.24\linewidth]{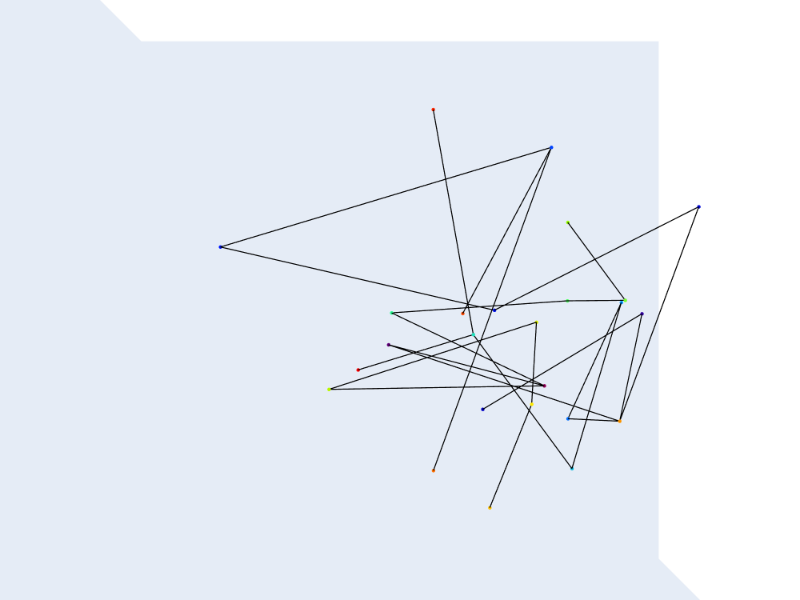}
        \includegraphics[width=.24\linewidth]{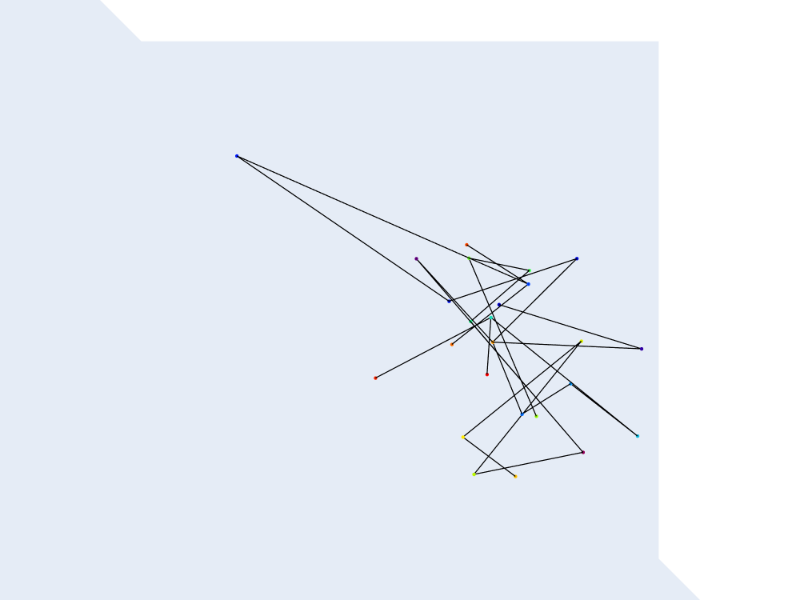}
        \includegraphics[width=.24\linewidth]{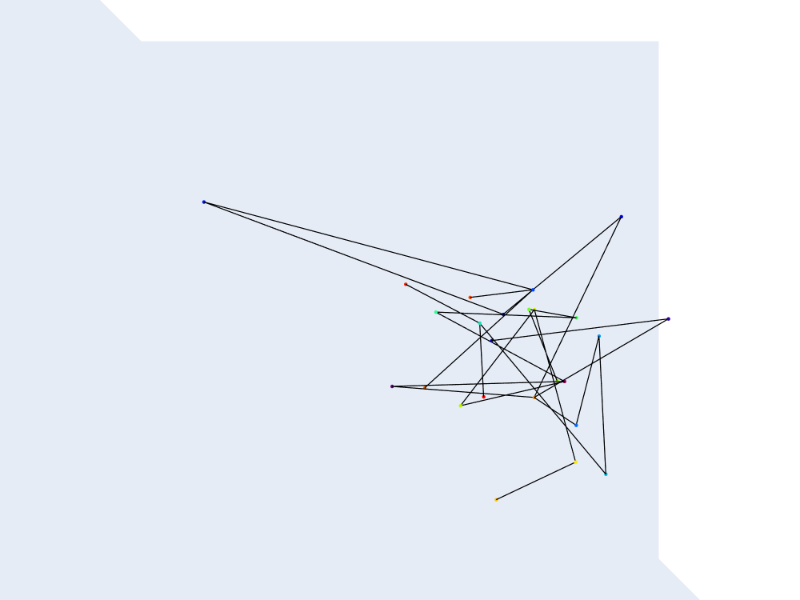}
        \caption{ResNet}
    \end{subfigure}

    \begin{subfigure}{\linewidth}
        \centering
        \includegraphics[width=.24\linewidth]{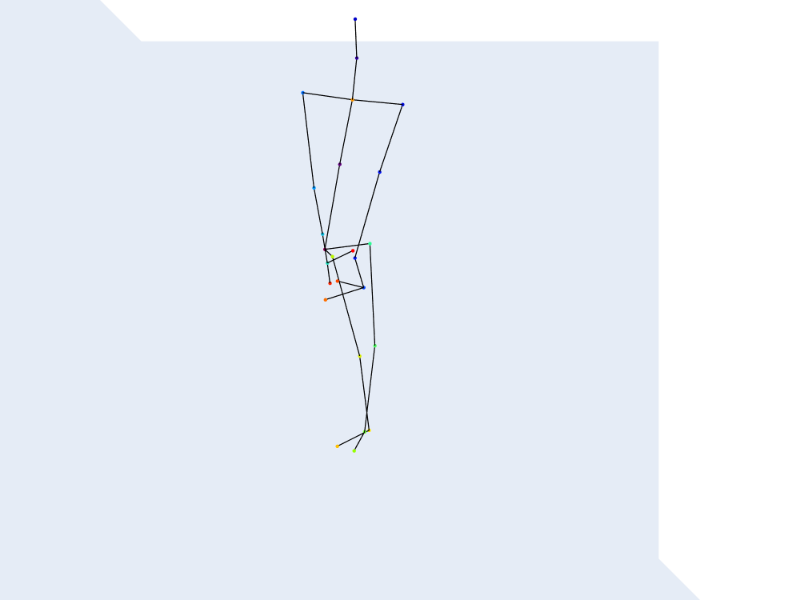}
        \includegraphics[width=.24\linewidth]{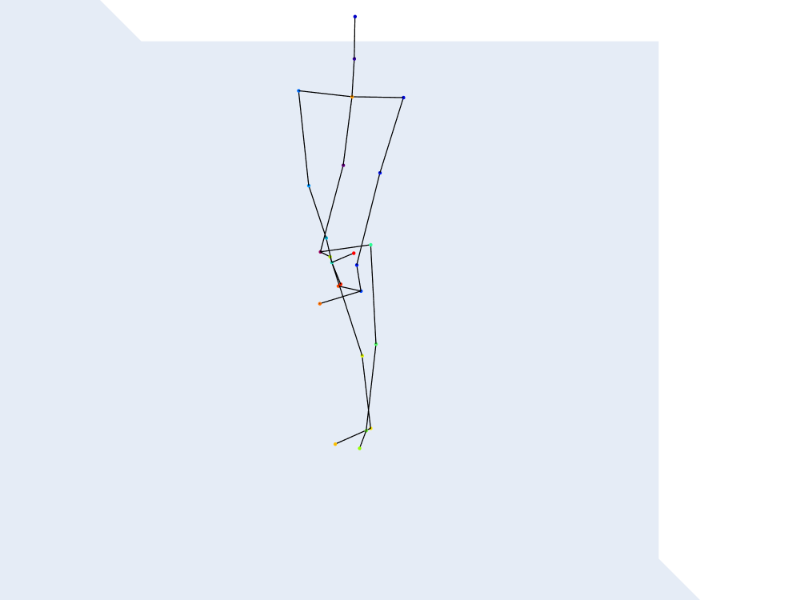}
        \includegraphics[width=.24\linewidth]{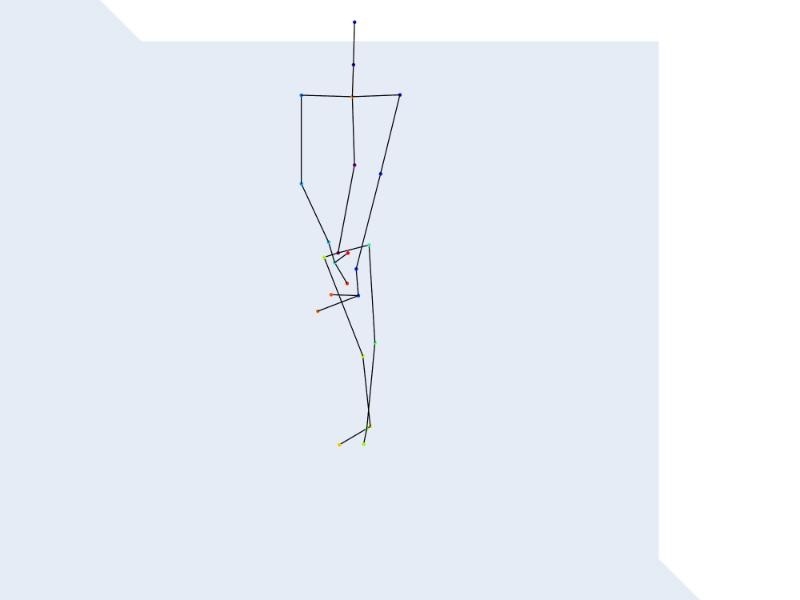}
        \includegraphics[width=.24\linewidth]{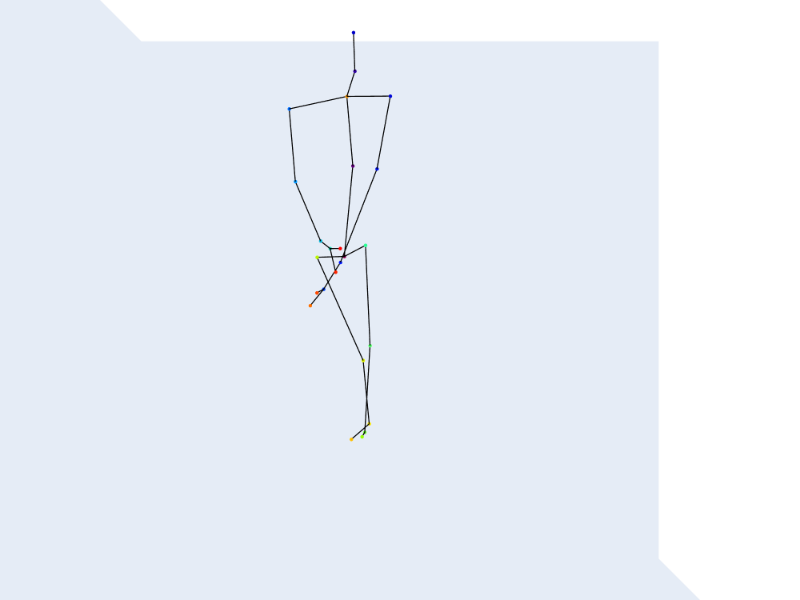}
        \caption{DMR}
    \end{subfigure}

    \begin{subfigure}{\linewidth}
        \centering
        \includegraphics[width=.24\linewidth]{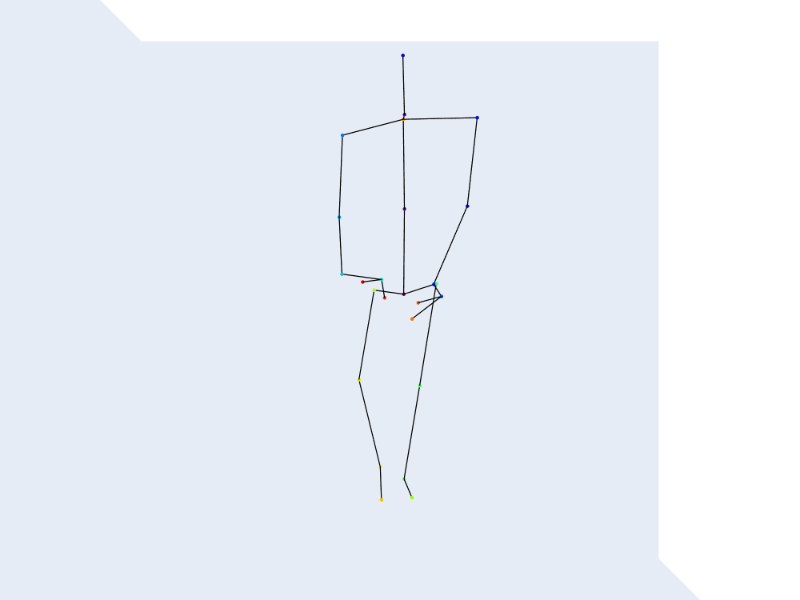}
        \includegraphics[width=.24\linewidth]{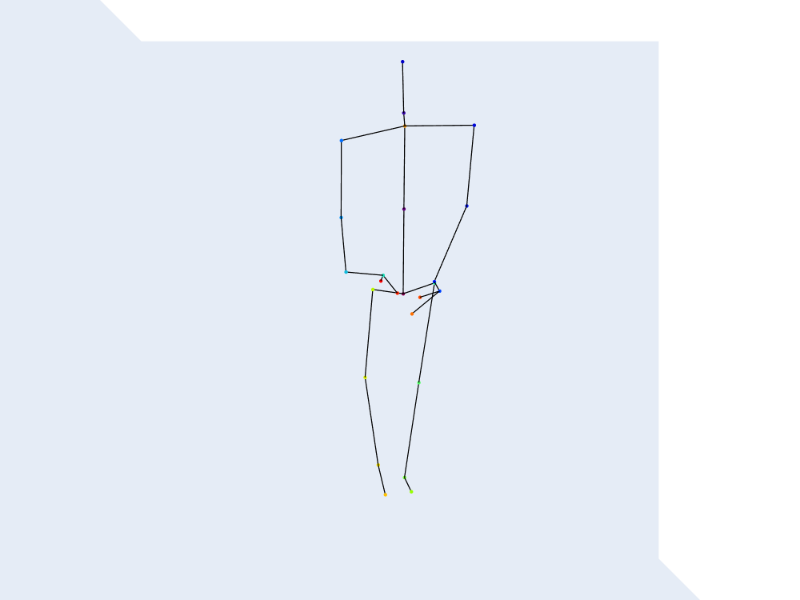}
        \includegraphics[width=.24\linewidth]{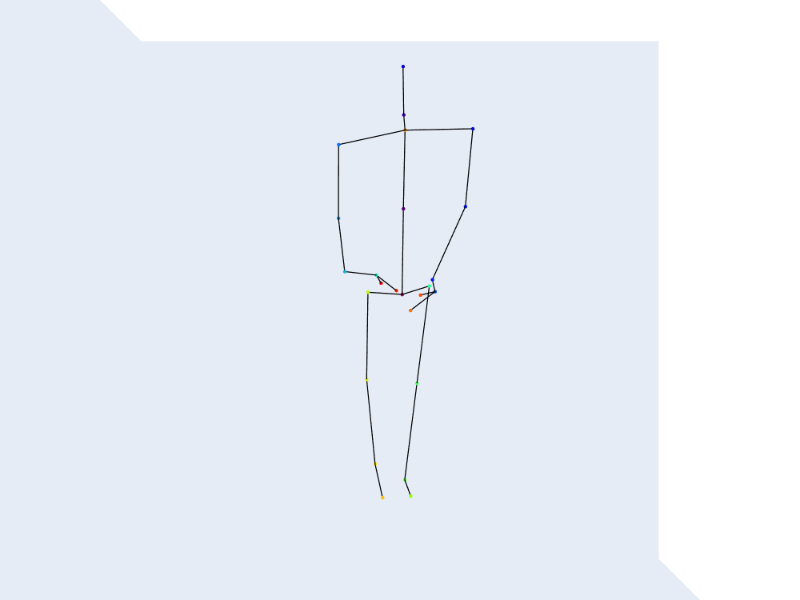}
        \includegraphics[width=.24\linewidth]{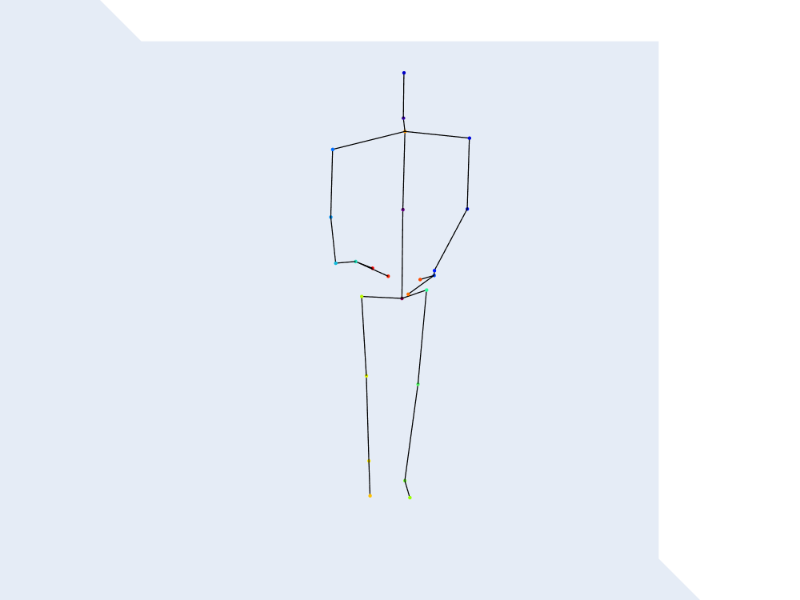}
        \caption{PMR}
    \end{subfigure}

    \caption{Example visualization: Actor 40 (Female) performing the ``Cross Hands in Front" action.} 
    \label{fig:vis}
\end{figure}

We further evaluate the models on qualitative analysis. Figure \ref{fig:vis} visualizes the anonymized skeletons, where the original skeleton is actor 40 (female) performing the ``Cross Hands in Front" action (more visualizations in the Supplementary Material Figure \ref{fig:vis2} and \ref{fig:vis3}). Although UNet and ResNet have the lowest privacy risk, it is at the cost of destroying the motion utility. The visualization is not human-like. Because it is only frame-level anonymization, the consistency between frames is bad. DMR has the closest visualization to the original. The action is retargeted to a male actor, thus the skeleton spatial size is larger than the original actor. However, there is still PII in how the original actor performs the action. It suggests a male actor is mimicking actor 40's action style. In PMR, the retargeted skeleton acts in a different manner. It no longer shows the action style of the original actor. The privacy risk is low, yet it still keeps the motion information that it is performing the ``Cross Hands in Front" action.

\subsection{Trade-off Analysis}
The hyper parameter $\alpha_{emb}$ controls the strength of cooperative and adversarial learning in PMR. Table \ref{tab:alpha} shows the trade-pff analysis on different values of $\alpha_{emb}$.  A stronger adversarial learning puts more restrictions on what information can be passed along to the retargeted skeleton.  When $\alpha_{emb} =40$, re-identification attack accuracy is lower but the MSE is higher. When $\alpha_{emb} = 10$, the privacy risk is still relatively low while the motion utility is high. When $\alpha_{emb} = 0$, PMR becomes similar to DMR.

\begin{table}[]
\small
\centering
\begin{tabular}{|l|l|ll|}
\hline
$\alpha_{emb}$ & Utility & \multicolumn{2}{c|}{Re-identification}           \\ \hline
Constant & MSE    & \multicolumn{1}{l|}{Top-1}  & Top-5  \\ \hline
0        & 0.0072 & \multicolumn{1}{l|}{0.1926} & 0.5066 \\ \hline
1        & 0.0116 & \multicolumn{1}{l|}{0.0417} & 0.2913 \\ \hline
5        & 0.0192 & \multicolumn{1}{l|}{0.0859} & 0.2738 \\ \hline
10       & 0.0140 & \multicolumn{1}{l|}{0.0794} & 0.2666 \\ \hline
20       & 0.0125 & \multicolumn{1}{l|}{0.0733} & 0.2262 \\ \hline
40       & 0.0261 & \multicolumn{1}{l|}{0.0559} & 0.2214 \\ \hline
Random   & MSE    & \multicolumn{1}{l|}{Top-1}  & Top-5  \\ \hline
0        & 0.0067 & \multicolumn{1}{l|}{0.2006} & 0.5057 \\ \hline
1        & 0.0116 & \multicolumn{1}{l|}{0.1135} & 0.4009 \\ \hline
5        & 0.0173 & \multicolumn{1}{l|}{0.0777} & 0.2800  \\ \hline
10       & 0.0126 & \multicolumn{1}{l|}{0.1119} & 0.3385 \\ \hline
20       & 0.0131 & \multicolumn{1}{l|}{0.0738} & 0.2573 \\ \hline
40       & 0.0241 & \multicolumn{1}{l|}{0.0600} & 0.2366 \\ \hline
\end{tabular}
\caption{Motion utility and privacy risk trade-off }
\label{tab:alpha}
\end{table}

\subsection{Representation Learning}
\begin{figure}
    \centering
    \begin{subfigure}{.49\linewidth}
        \centering
        \includegraphics[width=\linewidth]{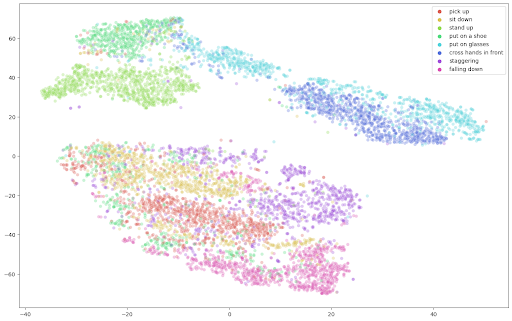}
        \caption{Motion Embedding}
    \end{subfigure}
    \label{fig:me}
    \begin{subfigure}{.49\linewidth}
        \centering
        \includegraphics[width=\linewidth]{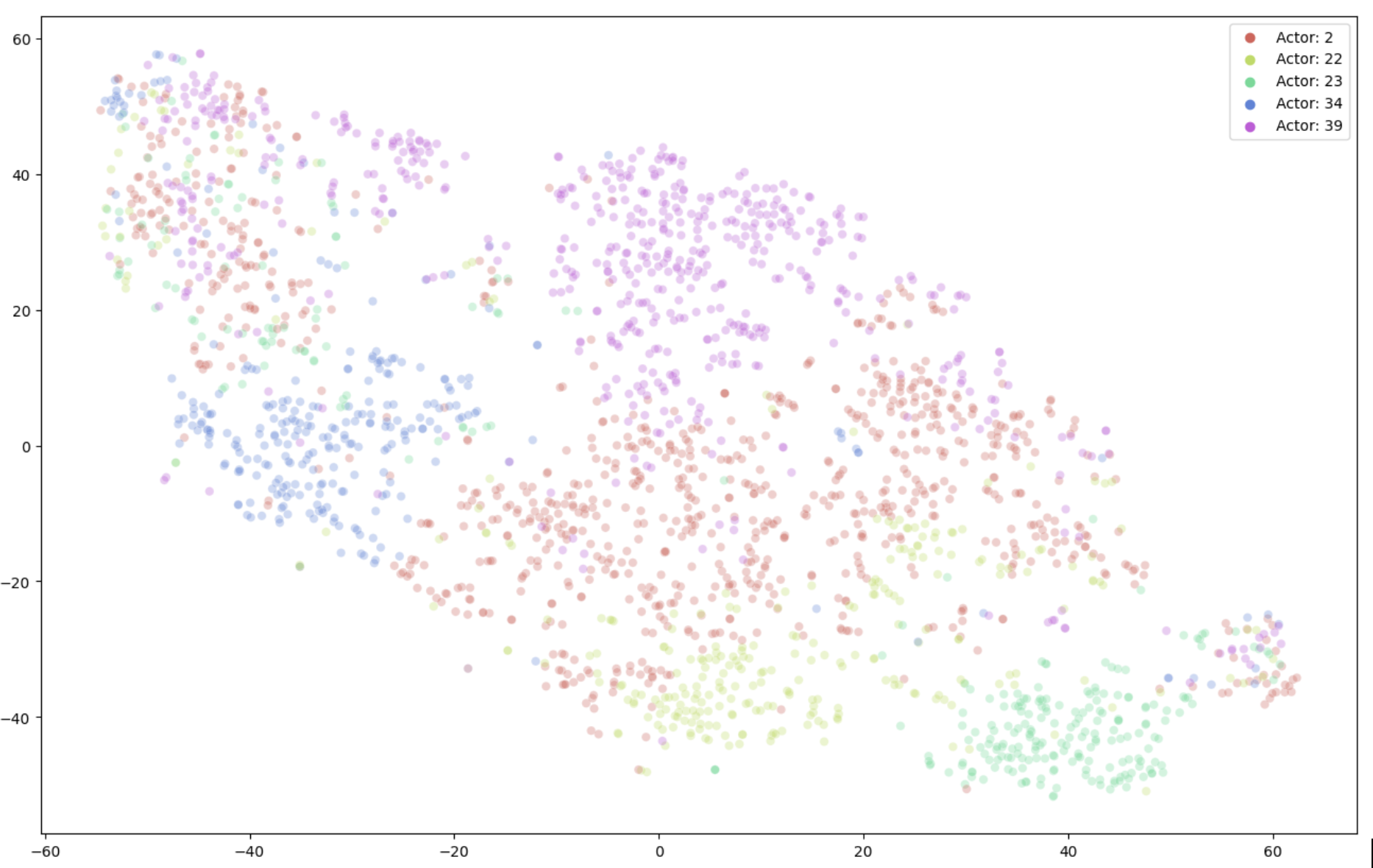}
        \caption{Privacy Embedding}
    \end{subfigure}
    \label{fig:pe}
    \caption{Latent space clustering on learned representations}
    \label{fig:clustering}
\end{figure}

One of the major differences between DMR and PMR is the enhancement on representation learning.  
To examine the quality of representation learning, we use a smaller subset of samples to run clustering analysis on motion embedding and privacy embedding. 
Figure \ref{fig:clustering} shows the clustering visualization using t-SNE. For motion embedding, the embeddings are clustered based on action labels. This shows the motion embedding is effective at capturing diverse motion information in the latent space. For privacy embedding, the embeddings are clustered based on actor ID. This shows the privacy embedding is effective at capturing PII for different actors, where the static skeleton plays a dominant role.

\section{Conclusion}
In this work, we propose a novel motion retargeting model to anonymize skeleton data. The original skeleton data contains PII, which exposes them to re-identification attacks. Using an autoencoder-based neural network, we decompose the skeleton into motion embedding and privacy embedding. The privacy embedding is substituted with one from a dummy skeleton. The decoder recompose a retargeted skeleton, which is anonymous. We use cooperative and adversarial learning with embedding classifiers to enhance the representation learning. Our evaluation shows the effectiveness of our proposed model and a balanced trade-off between motion utility and privacy risk in our approach. In the future, we will explore transformer-based models for skeleton anonymization. It has a strong representation learning ability to potentially separate PII from motion information.

\section*{Acknowledgements}
This work was supported in part by UNC Charlotte startup fund and NSF grant 1840080.

\clearpage
\newpage

\bibliographystyle{named}
\bibliography{main}

\begin{thebibliography}{}

\bibitem[\protect\citeauthoryear{Aberman \bgroup \em et al.\egroup
  }{2019}]{AbermanWLCC19}
Kfir Aberman, Rundi Wu, Dani Lischinski, Baoquan Chen, and Daniel Cohen{-}Or.
\newblock Learning character-agnostic motion for motion retargeting in 2d.
\newblock {\em {ACM} Trans. Graph.}, 38(4):75:1--75:14, 2019.

\bibitem[\protect\citeauthoryear{Aberman \bgroup \em et al.\egroup
  }{2020}]{AbermanLLSCC20}
Kfir Aberman, Peizhuo Li, Dani Lischinski, Olga Sorkine{-}Hornung, Daniel
  Cohen{-}Or, and Baoquan Chen.
\newblock Skeleton-aware networks for deep motion retargeting.
\newblock {\em {ACM} Trans. Graph.}, 39(4):62, 2020.

\bibitem[\protect\citeauthoryear{Carr \bgroup \em et al.\egroup
  }{2023}]{linkage}
Thomas Carr, Aidong Lu, and Depeng Xu.
\newblock Linkage attack on skeleton-based motion visualization.
\newblock In {\em Proceedings of the 32nd {ACM} International Conference on
  Information and Knowledge Management, {CIKM} 2023, Birmingham, United
  Kingdom, October 21-25, 2023}, pages 3758--3762. {ACM}, 2023.

\bibitem[\protect\citeauthoryear{Doula \bgroup \em et al.\egroup
  }{2022}]{10.1145/3491101.3519645}
Achref Doula, Alejandro Sanchez~Guinea, and Max M\"{u}hlh\"{a}user.
\newblock Vr-surv: A vr-based privacy preserving surveillance system.
\newblock In {\em Extended Abstracts of the 2022 CHI Conference on Human
  Factors in Computing Systems}, CHI EA '22, New York, NY, USA, 2022.
  Association for Computing Machinery.

\bibitem[\protect\citeauthoryear{Fanello \bgroup \em et al.\egroup
  }{2013}]{FanelloGMO13}
Sean~Ryan Fanello, Ilaria Gori, Giorgio Metta, and Francesca Odone.
\newblock Keep it simple and sparse: real-time action recognition.
\newblock {\em J. Mach. Learn. Res.}, 14(1):2617--2640, 2013.

\bibitem[\protect\citeauthoryear{Gal \bgroup \em et al.\egroup
  }{2015}]{GalANNS15}
Norbert Gal, Diana Andrei, Dan~Ion Nemes, Emanuela Nadasan, and Vasile
  Stoicu{-}Tivadar.
\newblock A kinect based intelligent e-rehabilitation system in physical
  therapy.
\newblock In Ronald Cornet, Lacramioara Stoicu{-}Tivadar, Alexander
  H{\"{o}}rbst, Carlos Luis~Parra Calder{\'{o}}n, Stig~Kj{\ae}r Andersen, and
  Mira Hercigonja{-}Szekeres, editors, {\em Digital Healthcare Empowering
  Europeans - Proceedings of MIE2015, Madrid Spain, 27-29 May, 2015}, volume
  210 of {\em Studies in Health Technology and Informatics}, pages 489--493.
  {IOS} Press, 2015.

\bibitem[\protect\citeauthoryear{Goodfellow \bgroup \em et al.\egroup
  }{2014}]{gan}
Ian~J. Goodfellow, Jean Pouget{-}Abadie, Mehdi Mirza, Bing Xu, David
  Warde{-}Farley, Sherjil Ozair, Aaron~C. Courville, and Yoshua Bengio.
\newblock Generative adversarial nets.
\newblock In Zoubin Ghahramani, Max Welling, Corinna Cortes, Neil~D. Lawrence,
  and Kilian~Q. Weinberger, editors, {\em Advances in Neural Information
  Processing Systems 27: Annual Conference on Neural Information Processing
  Systems 2014, December 8-13 2014, Montreal, Quebec, Canada}, pages
  2672--2680, 2014.

\bibitem[\protect\citeauthoryear{Lin and
  Latoschik}{2022}]{10.3389/frvir.2022.974652}
Jinghuai Lin and Marc~Erich Latoschik.
\newblock Digital body, identity and privacy in social virtual reality: A
  systematic review.
\newblock {\em Frontiers in Virtual Reality}, 3, 2022.

\bibitem[\protect\citeauthoryear{Moon \bgroup \em et al.\egroup
  }{2023}]{KimM2021}
Saemi Moon, Myeonghyeon Kim, Zhenyue Qin, Yang Liu, and Dongwoo Kim.
\newblock Anonymization for skeleton action recognition.
\newblock {AAAI} Press, 2023.

\bibitem[\protect\citeauthoryear{Olade \bgroup \em et al.\egroup
  }{2020}]{OladeFL20}
Ilesanmi Olade, Charles Fleming, and Hai{-}Ning Liang.
\newblock Biomove: Biometric user identification from human kinesiological
  movements for virtual reality systems.
\newblock {\em Sensors}, 20(10):2944, 2020.

\bibitem[\protect\citeauthoryear{Saggese \bgroup \em et al.\egroup
  }{2019}]{SaggeseSVP19}
Alessia Saggese, Nicola Strisciuglio, Mario Vento, and Nicolai Petkov.
\newblock Learning skeleton representations for human action recognition.
\newblock {\em Pattern Recognit. Lett.}, 118:23--31, 2019.

\bibitem[\protect\citeauthoryear{Shahroudy \bgroup \em et al.\egroup
  }{2016}]{ntu60}
Amir Shahroudy, Jun Liu, Tian{-}Tsong Ng, and Gang Wang.
\newblock {NTU} {RGB+D:} {A} large scale dataset for 3d human activity
  analysis.
\newblock In {\em 2016 {IEEE} Conference on Computer Vision and Pattern
  Recognition, {CVPR} 2016, Las Vegas, NV, USA, June 27-30, 2016}, pages
  1010--1019. {IEEE} Computer Society, 2016.

\bibitem[\protect\citeauthoryear{Shi \bgroup \em et al.\egroup
  }{2021}]{s21010205}
Jiaqi Shi, Chaoran Liu, Carlos~Toshinori Ishi, and Hiroshi Ishiguro.
\newblock Skeleton-based emotion recognition based on two-stream self-attention
  enhanced spatial-temporal graph convolutional network.
\newblock {\em Sensors}, 21(1), 2021.

\bibitem[\protect\citeauthoryear{Tran \bgroup \em et al.\egroup
  }{2018}]{TranWTRLP18}
Du~Tran, Heng Wang, Lorenzo Torresani, Jamie Ray, Yann LeCun, and Manohar
  Paluri.
\newblock A closer look at spatiotemporal convolutions for action recognition.
\newblock In {\em 2018 {IEEE} Conference on Computer Vision and Pattern
  Recognition, {CVPR} 2018, Salt Lake City, UT, USA, June 18-22, 2018}, pages
  6450--6459. Computer Vision Foundation / {IEEE} Computer Society, 2018.

\bibitem[\protect\citeauthoryear{Villegas \bgroup \em et al.\egroup
  }{2018}]{NKN}
Ruben Villegas, Jimei Yang, Duygu Ceylan, and Honglak Lee.
\newblock Neural kinematic networks for unsupervised motion retargetting.
\newblock In {\em 2018 {IEEE} Conference on Computer Vision and Pattern
  Recognition, {CVPR} 2018, Salt Lake City, UT, USA, June 18-22, 2018}, pages
  8639--8648. Computer Vision Foundation / {IEEE} Computer Society, 2018.

\bibitem[\protect\citeauthoryear{Zhang \bgroup \em et al.\egroup }{2020}]{sgn}
Pengfei Zhang, Cuiling Lan, Wenjun Zeng, Junliang Xing, Jianru Xue, and Nanning
  Zheng.
\newblock Semantics-guided neural networks for efficient skeleton-based human
  action recognition.
\newblock In {\em 2020 {IEEE/CVF} Conference on Computer Vision and Pattern
  Recognition, {CVPR} 2020, Seattle, WA, USA, June 13-19, 2020}, pages
  1109--1118. Computer Vision Foundation / {IEEE}, 2020.

\end{thebibliography}

\clearpage
\newpage

\appendix
\section{Supplementary Material}
\subsubsection{Embedding Classifier Accuracy}

\begin{figure}[!htb]
    \centering
    \includegraphics[width=\linewidth]{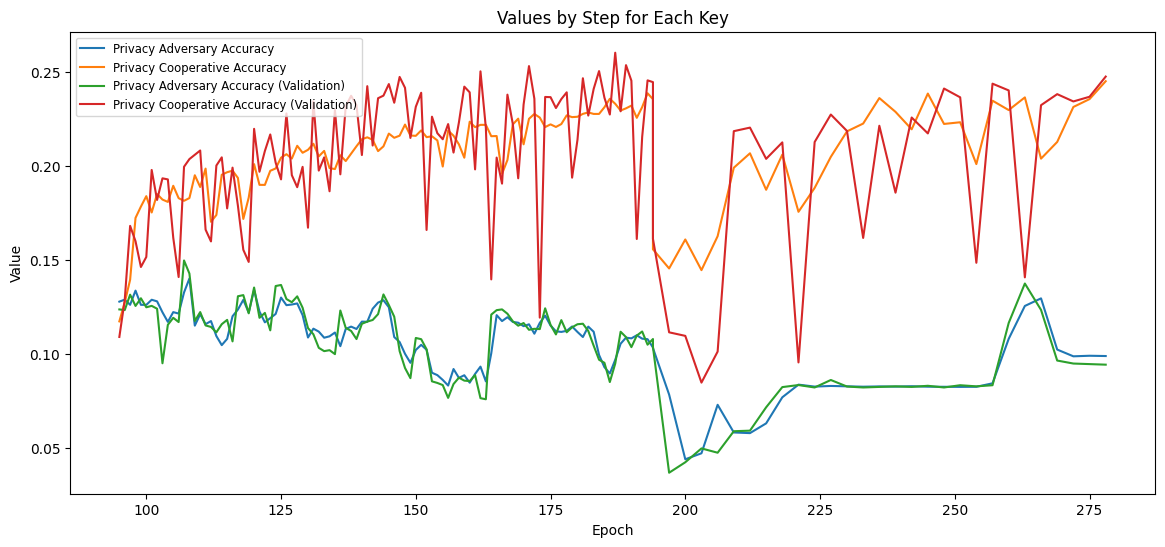}
    \caption{Privacy Classifier Accuracy}
    \label{fig:emb_priv_acc}
\end{figure}

Figure~\ref{fig:emb_priv_acc} illustrates the privacy classifier's accuracy on training and validation. Red and green lines represent accuracy on the privacy embedding, whereas blue and purple lines show accuracy on the motion embedding. The aim is low performance on the motion embedding, indicating minimal PII presence.

\begin{figure}[!htb]
    \centering
    \includegraphics[width=\linewidth]{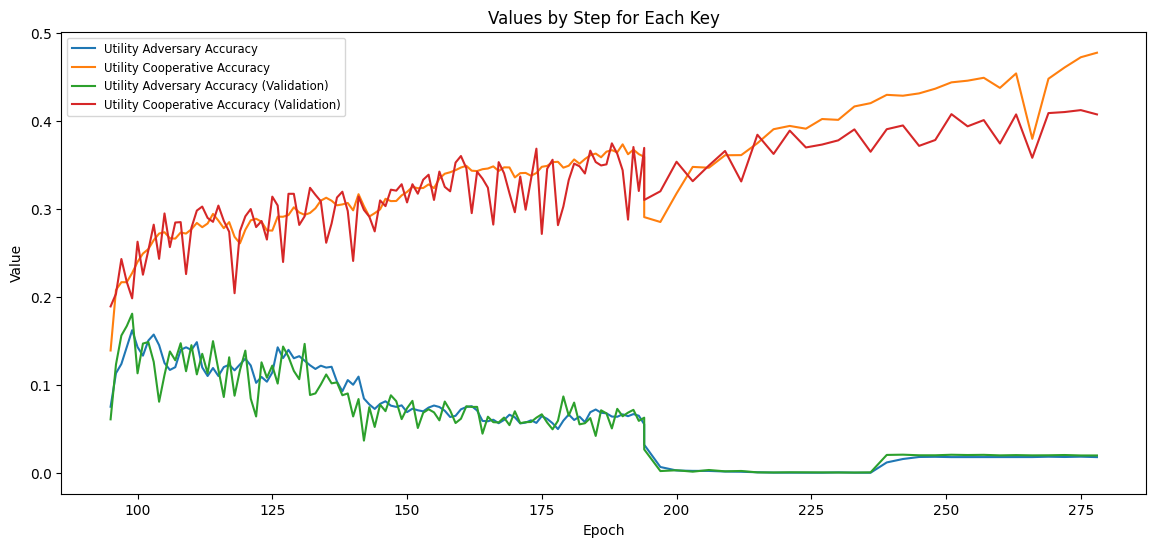}
    \caption{Utility Classifier Accuracy}
    \label{fig:emb_util_acc}
\end{figure}

Figure~\ref{fig:emb_util_acc} displays the utility classifier's accuracy in training and validation. Here, red and green lines indicate accuracy on the motion embedding, focusing on the skeleton's utility. Blue and purple lines represent performance on the privacy embedding. The goal is high accuracy on the motion embedding and low on the privacy embedding, ensuring motion information exclusivity.

\subsubsection{Training Stages}

\begin{table}[!htb]
\centering
\begin{tabular}{|l|l|l|}
\hline
\textbf{Stage}                                 & \textbf{Paired} & \textbf{Epochs} \\ \hline
Pre-training the Auto-Encoder                  & Yes    & 5      \\ \hline
Pre-training the Auto-Encoder                  & No     & 20     \\ \hline
Pre-training the Embedding Classifiers         & Yes    & 20     \\ \hline
Pre-training the Embedding Classifiers         & No     & 50     \\ \hline
Unpaired Training                              & No     & 100    \\ \hline
Paired Training                                & Yes    & 80    \\ \hline
\end{tabular}
\caption{Training stages used for final model}
\label{tab:stages}
\end{table}

Table~\ref{tab:stages} details the epochs and stages of training. Initial pre-training primes the models briefly. The first two stages utilize paired and unpaired data for embedding separation. The next two stages ready the embedding classifiers for iterative cooperative and adversarial training. In the unpaired stage, the model is learning how motion data works and learns how to breakdown and reconstruct the skeletons. Most of the emphasis is put on the paired training, where the motion retargeting is fine tuned.

\subsubsection{Model Architecture}

\begin{table}[!htb]
\small
\centering
\begin{tabular}{cccccc}
\hline
\multicolumn{1}{|c|}{Name} &
  \multicolumn{1}{c|}{Layers} &
  \multicolumn{1}{c|}{$k$} &
  \multicolumn{1}{c|}{$s$} &
  \multicolumn{1}{c|}{$p$} &
  \multicolumn{1}{c|}{in/out} \\ \hline
                      &                       &     &   &   &                               \\ \hline
\multicolumn{1}{|c}{\multirow{4}{*}{Encoder}} &
  C2D + LR + MP + RP2D &
  3x3 &
  1 &
  - &
  \multicolumn{1}{c|}{75/12} \\
\multicolumn{1}{|c}{} & C2D + LR + MP + RP2D  & 3x3 & 1 & - & \multicolumn{1}{c|}{12/24}    \\
\multicolumn{1}{|c}{} & C2D + LR + MP + RP2D  & 3x3 & 1 & - & \multicolumn{1}{c|}{24/32}    \\
\multicolumn{1}{|c}{} & C2D + LR + MP + RP2D  & 3x3 & 1 & - & \multicolumn{1}{c|}{32/256}   \\ \hline
                      &                       &     &   &   &                               \\ \hline
\multicolumn{1}{|c}{\multirow{4}{*}{Decoder}} &
  CT2D + LR + Up + RP2D &
  3x3 &
  1 &
  1 &
  \multicolumn{1}{c|}{512/ 256} \\
\multicolumn{1}{|c}{} & CT2D + LR + Up + RP2D & 3x3 & 1 & 1 & \multicolumn{1}{c|}{256/128}  \\
\multicolumn{1}{|c}{} & CT2D + LR + Up + RP2D & 3x3 & 1 & 1 & \multicolumn{1}{c|}{128/96}   \\
\multicolumn{1}{|c}{} & CT2D + LR + Up + RP2D & 3x3 & 1 & 1 & \multicolumn{1}{c|}{96/75}    \\ \hline
                      &                       &     &   &   &                               \\ \hline
\multicolumn{1}{|c}{\multirow{6}{*}{\begin{tabular}[c]{@{}c@{}}Embedding\\ Classifier\end{tabular}}} &
  CT1D + BN + R &
  3 &
  1 &
  1 &
  \multicolumn{1}{c|}{256/128} \\
\multicolumn{1}{|c}{} & CT1D + BN + R         & 3   & 1 & 1 & \multicolumn{1}{c|}{128/256}  \\
\multicolumn{1}{|c}{} & CT1D + BN + R + AP    & 3   & 1 & 1 & \multicolumn{1}{c|}{256/512}  \\
\multicolumn{1}{|c}{} & FL + LR + R           & -   & - & - & \multicolumn{1}{c|}{512/1024} \\
\multicolumn{1}{|c}{} & LR + R                & -   & - & - & \multicolumn{1}{c|}{1024/512} \\
\multicolumn{1}{|c}{} & LR + R + SM           & -   & - & - & \multicolumn{1}{c|}{512/Y}    \\ \hline
                      &                       &     &   &   &                               \\ \hline
\multicolumn{1}{|c}{\multirow{6}{*}{\begin{tabular}[c]{@{}c@{}}Quality\\ Control\\ (Discrim)\end{tabular}}} &
  CT1D + LR + Up + RP1D &
  3 &
  1 &
  1 &
  \multicolumn{1}{c|}{T/64} \\
\multicolumn{1}{|c}{} & CT1D + LR + Up + RP1D & 3   & 1 & 1 & \multicolumn{1}{c|}{64/32}    \\
\multicolumn{1}{|c}{} & CT1D + LR + Up + RP1D & 3   & 1 & 1 & \multicolumn{1}{c|}{32/16}    \\
\multicolumn{1}{|c}{} & CT1D + LR + Up + RP1D & 3   & 1 & 1 & \multicolumn{1}{c|}{16/8}     \\
\multicolumn{1}{|c}{} & FL + LR + R           & -   & - & - & \multicolumn{1}{c|}{80/32}    \\
\multicolumn{1}{|c}{} & LR + Sig              & -   & - & - & \multicolumn{1}{c|}{32/1}     \\ \hline
\end{tabular}
\caption{Model Implementations. $k$ represents kernel size, $s$ represents stride, and $p$ represents padding}
\label{tab:arch}
\end{table}

Table~\ref{tab:arch} shows our pytorch implementation of the encoders, decoder, embedding classifiers, and the quality controller. The embedding were of shape (256, 32) and in our codebase is a tunable hyperparameter. Table~\ref{tab:acronyms} list of acronyms used:

\begin{table}[!htb]
\centering
\small
\begin{tabular}{|l|l|}
\hline
\textbf{Acronym} & \textbf{Definition}                              \\ \hline
C2D              & Convolutional 2D                                 \\ \hline
CT2D/1D          & Convolutional Transpose 2D/1D                    \\ \hline
LR               & Leaky ReLU                                       \\ \hline
R                & ReLU                                             \\ \hline
Up               & Upsample                                         \\ \hline
MP/AP            & Max/Average Pooling                              \\ \hline
RP/2D            & Reflection Pad/2D                                \\ \hline
MLP              & Multi-Layer Perceptron                           \\ \hline
BN               & Batch Normalization                              \\ \hline
Y                & Number of classes                                \\ \hline
FL               & Flatten                                          \\ \hline
SM               & Softmax                                          \\ \hline
Sig              & Sigmoid                                          \\ \hline
\end{tabular}
\caption{Definitions of Acronyms Used}
\label{tab:acronyms}
\end{table}

\begin{figure*}[!htb]
    \centering

    \begin{subfigure}{\linewidth}
        \centering
        \includegraphics[width=.195\linewidth]{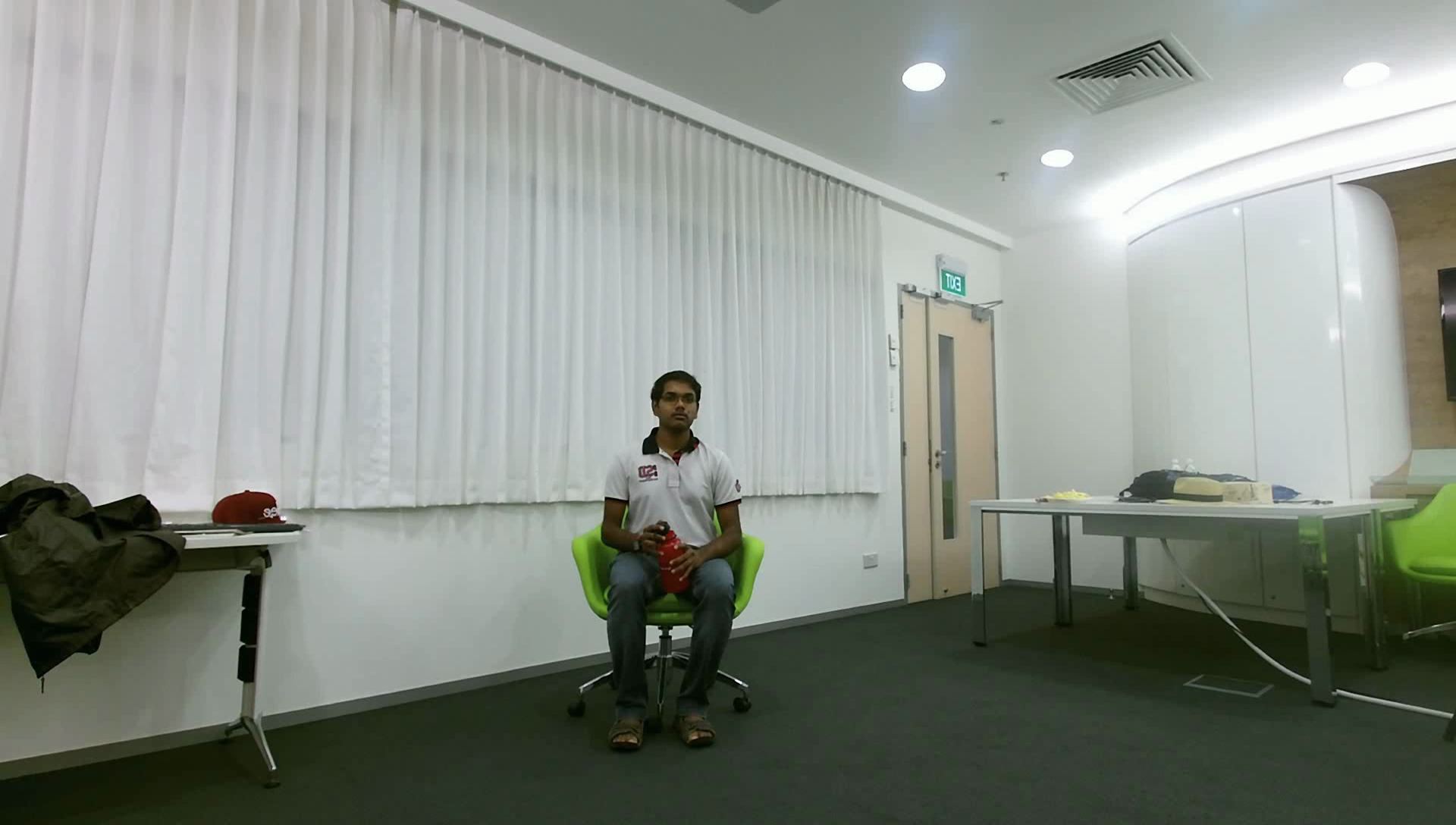}
        \includegraphics[width=.195\linewidth]{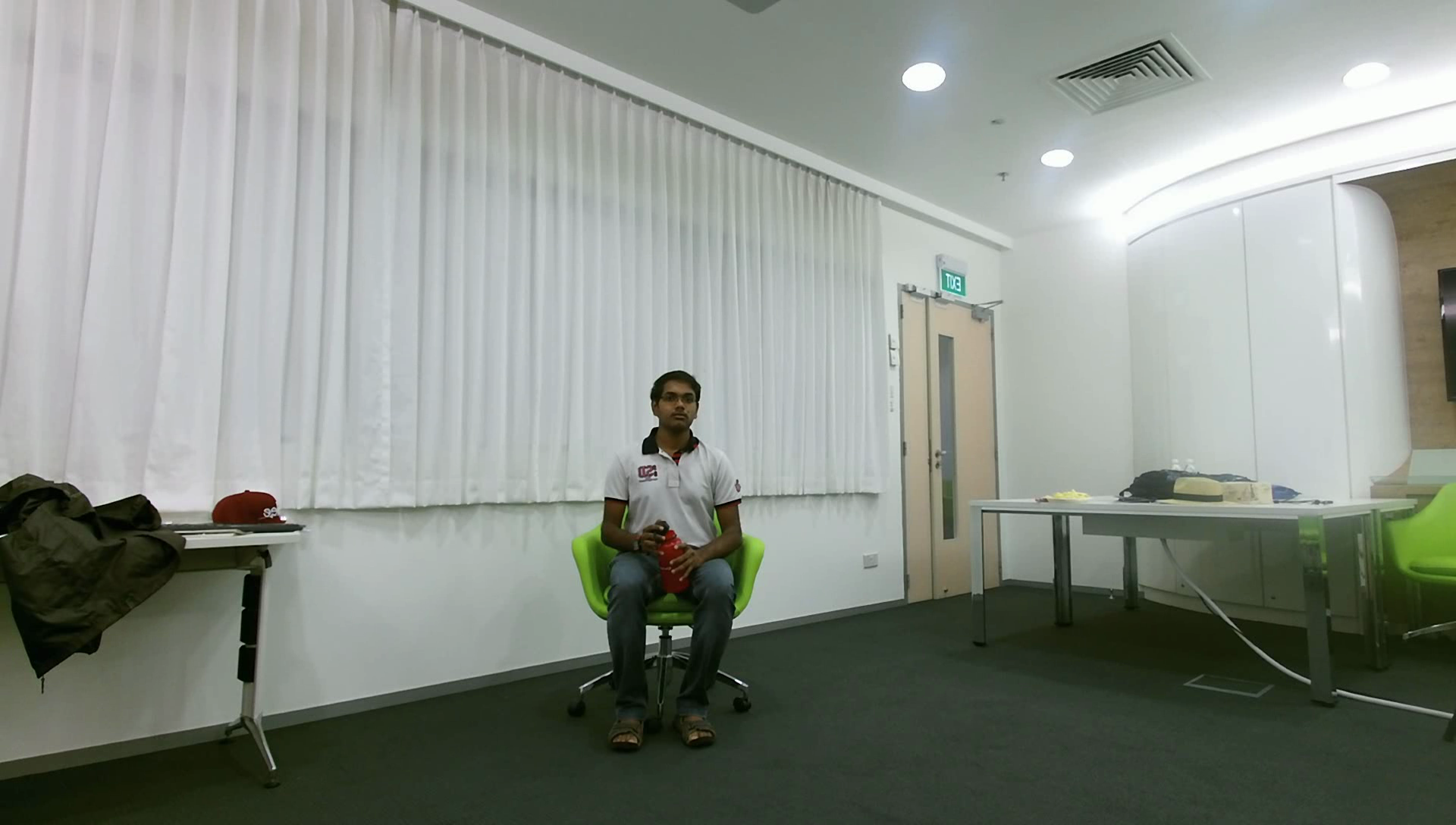}
        \includegraphics[width=.195\linewidth]{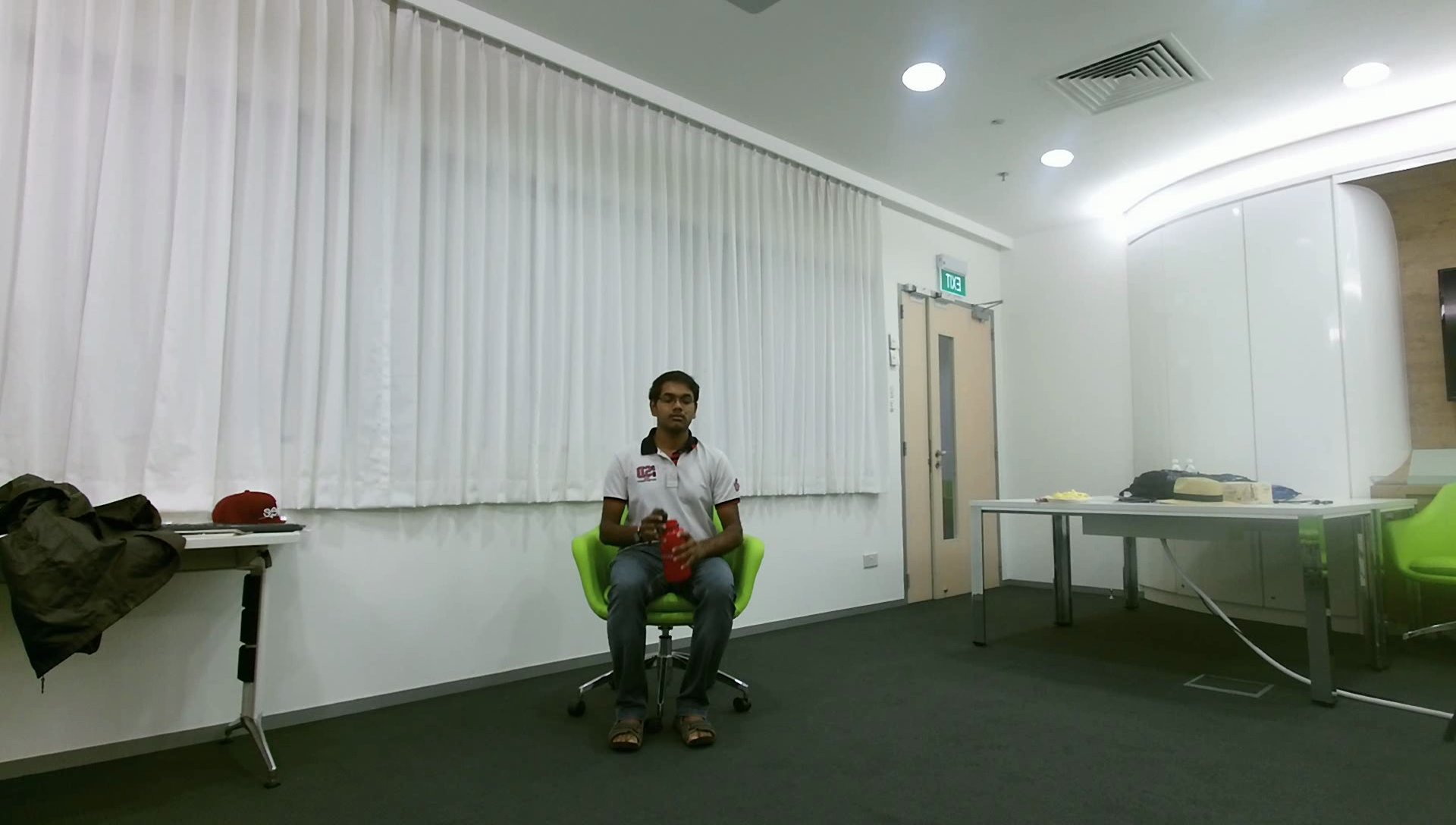}
        \includegraphics[width=.195\linewidth]{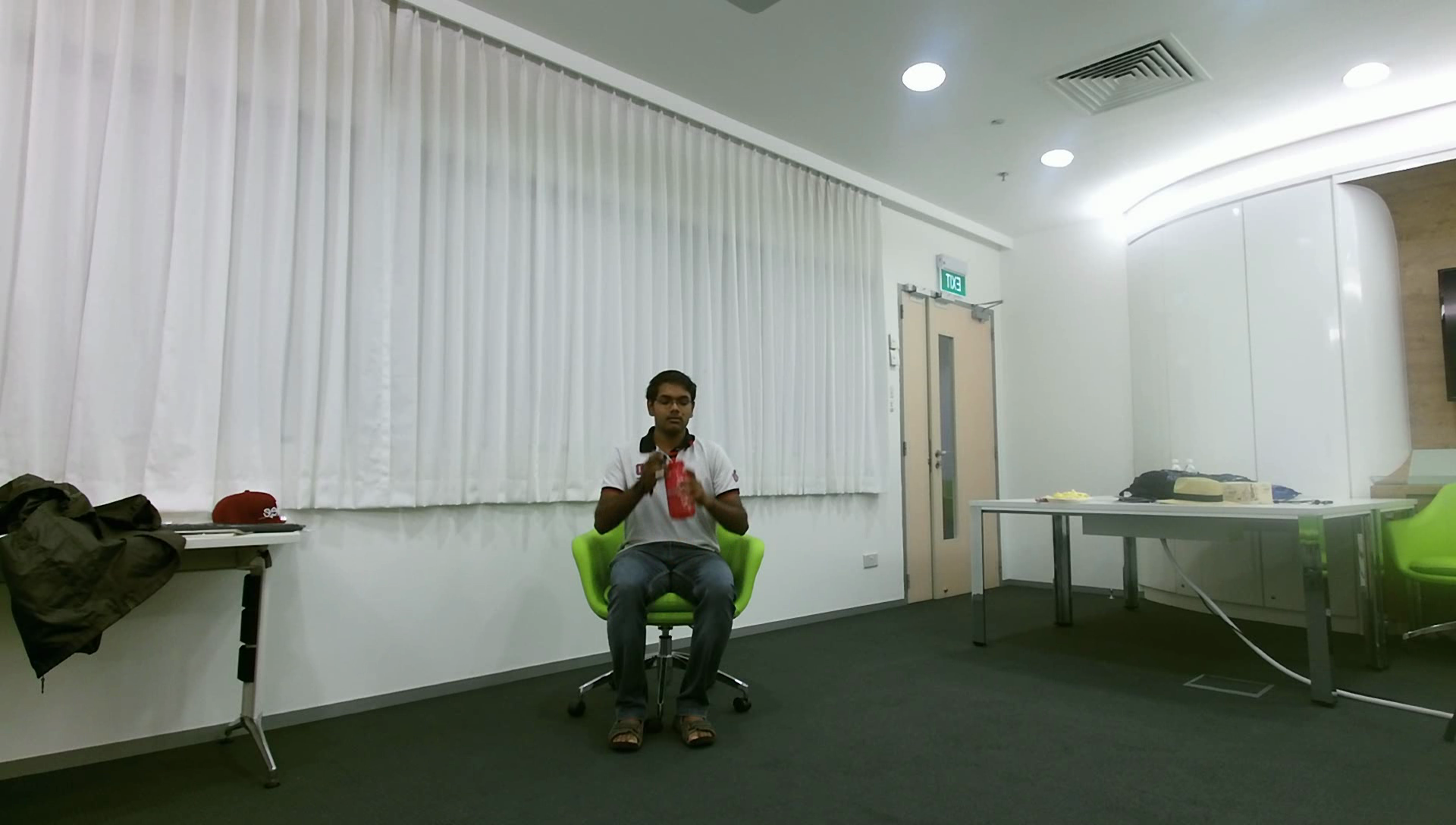}
        \includegraphics[width=.195\linewidth]{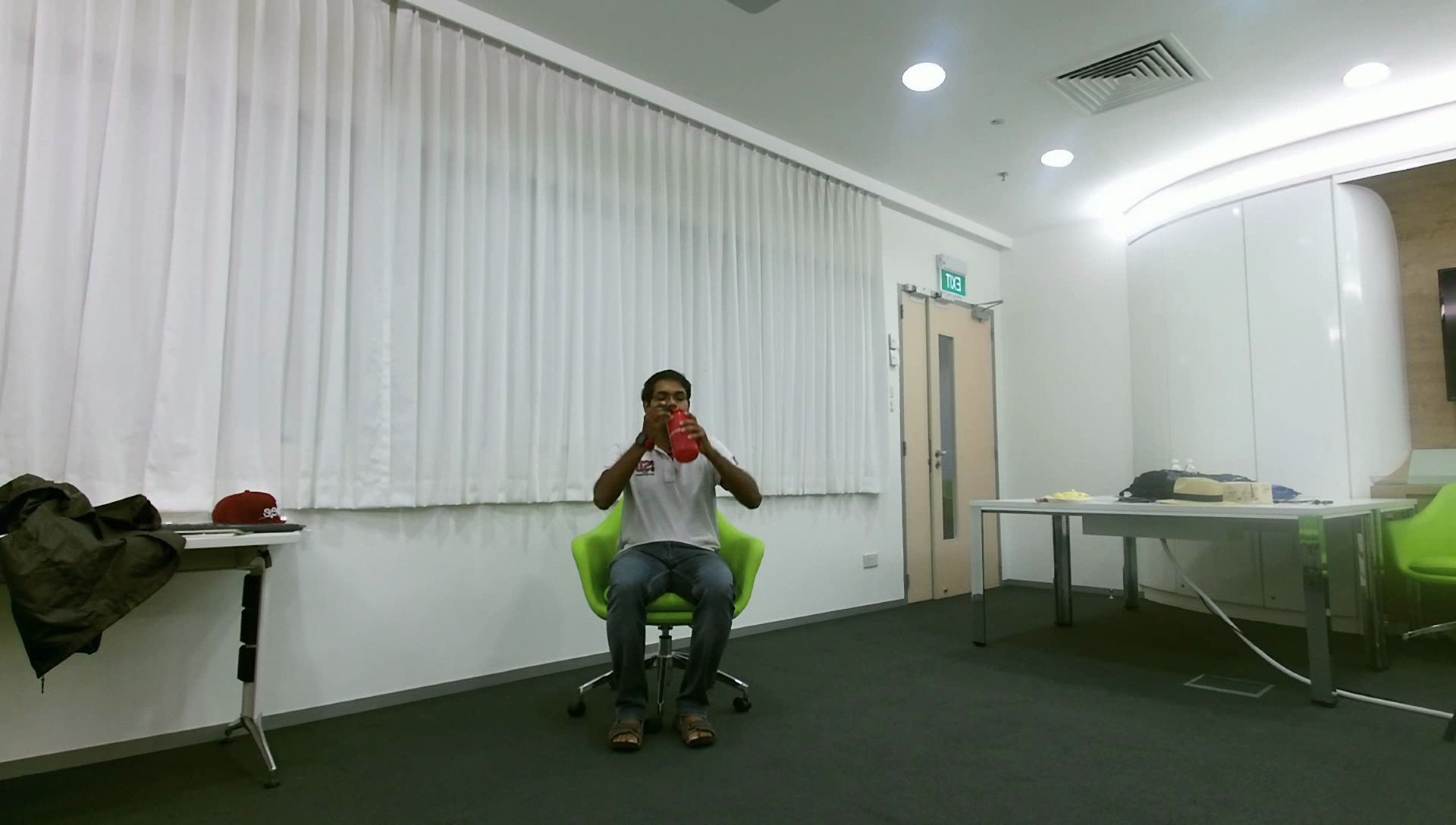}
        \caption{Original RGB Video}
    \end{subfigure}

    \begin{subfigure}{\linewidth}
        \centering
        \includegraphics[width=.195\linewidth]{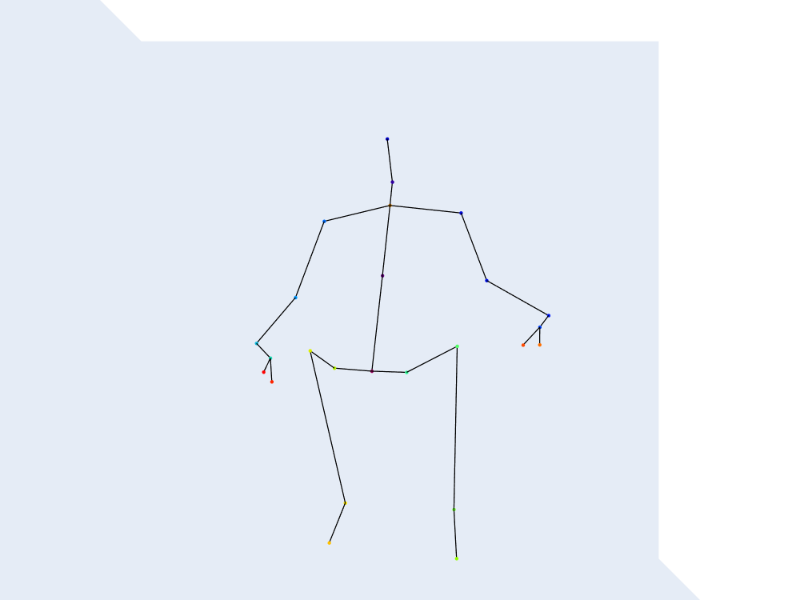}
        \includegraphics[width=.195\linewidth]{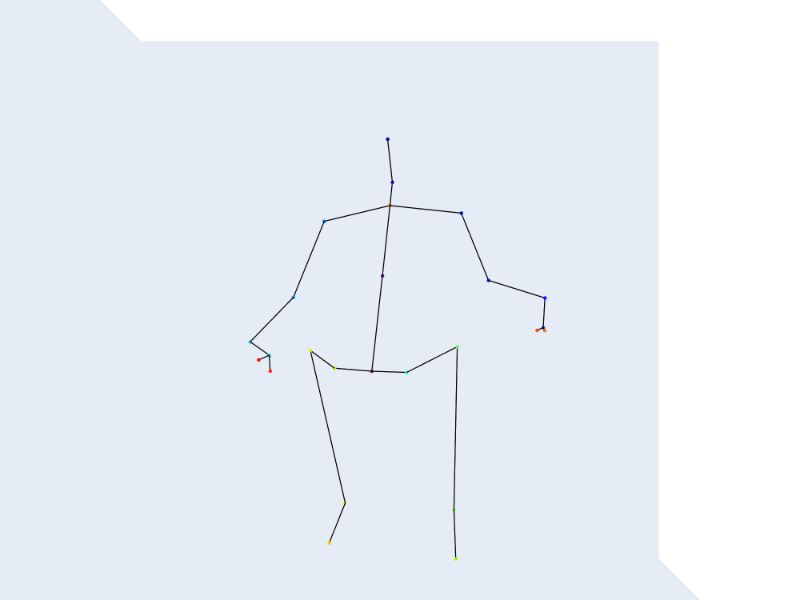}
        \includegraphics[width=.195\linewidth]{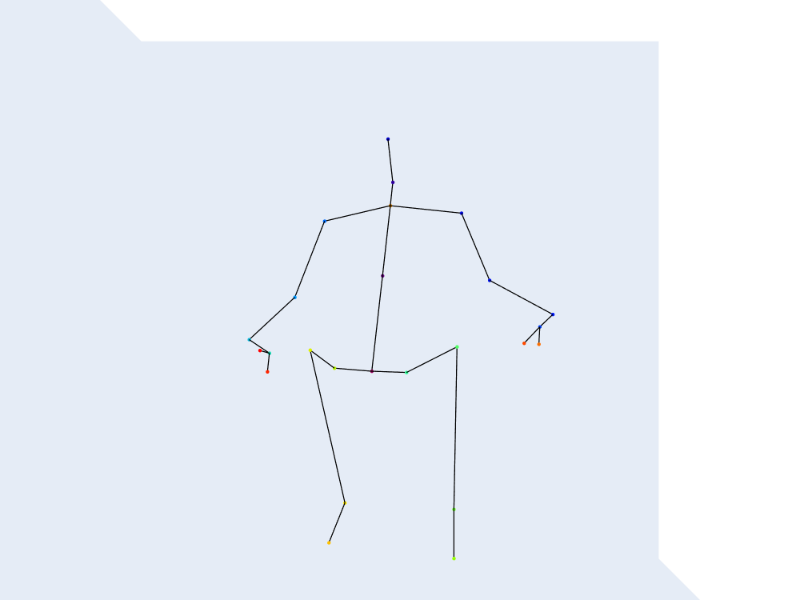}
        \includegraphics[width=.195\linewidth]{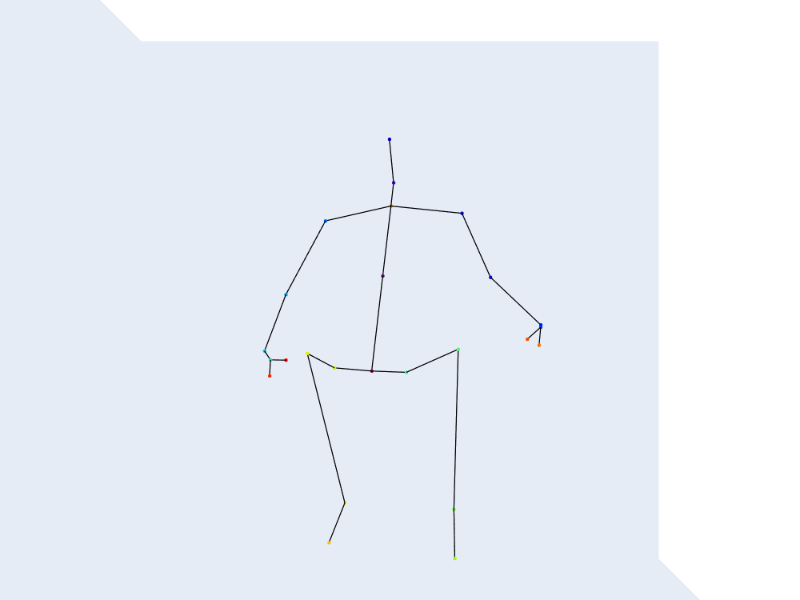}
        \includegraphics[width=.195\linewidth]{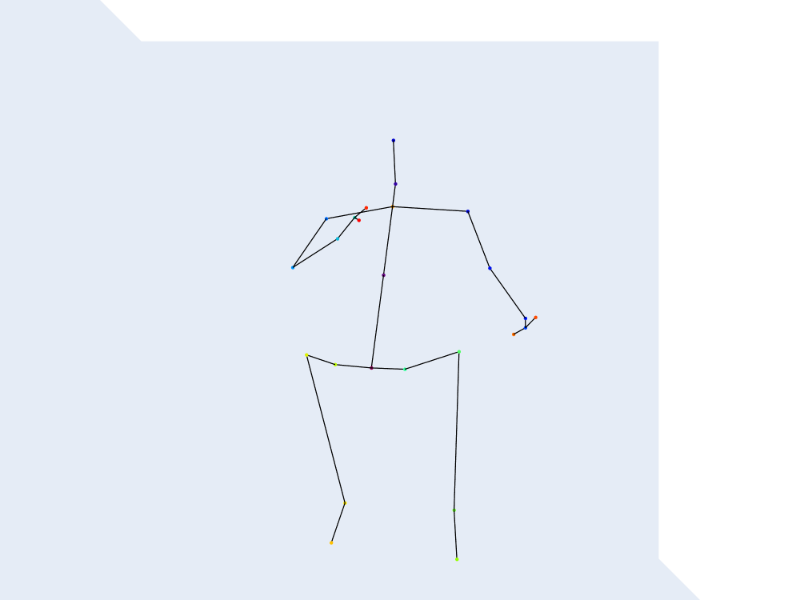}
        \caption{Original}
    \end{subfigure}

     \begin{subfigure}{\linewidth}
         \centering
         \includegraphics[width=.195\linewidth]{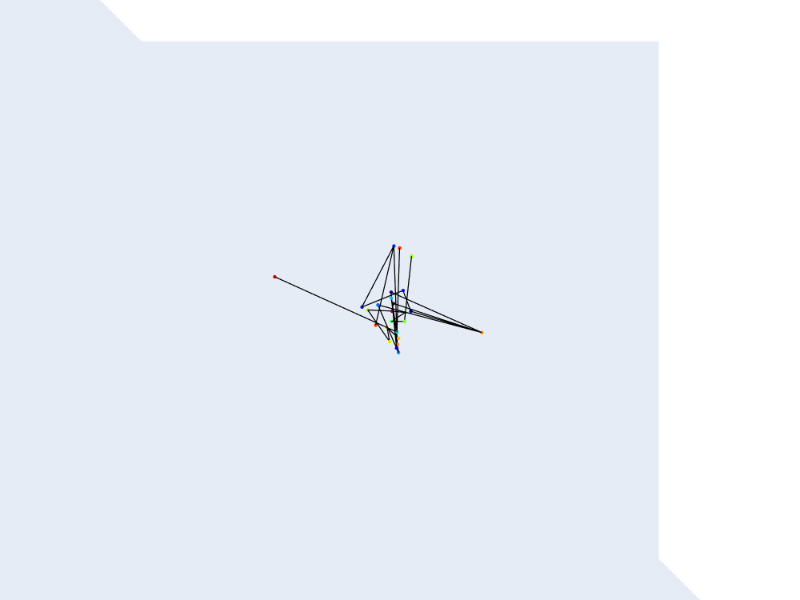}
         \includegraphics[width=.195\linewidth]{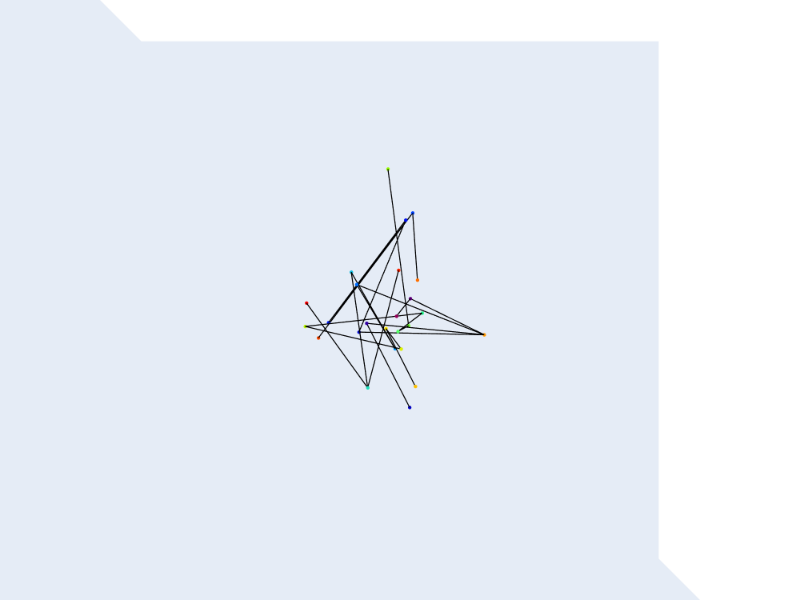}
         \includegraphics[width=.195\linewidth]{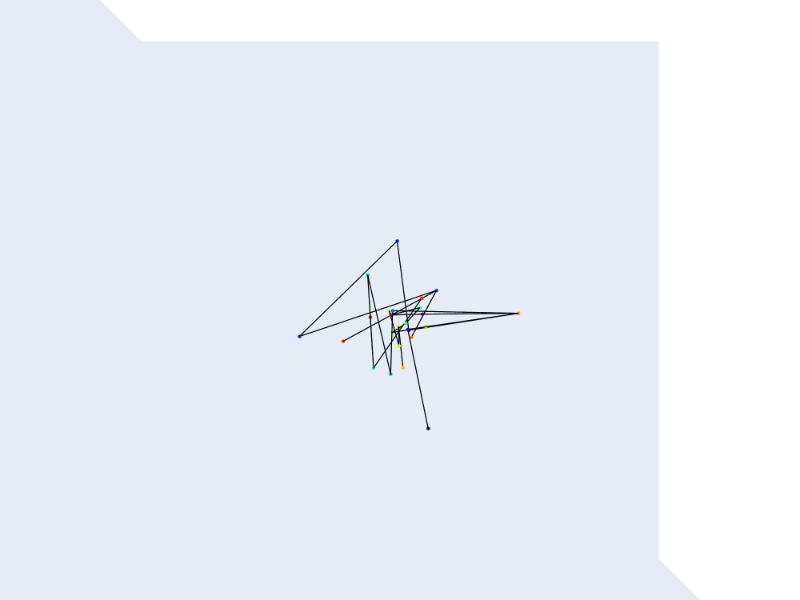}
         \includegraphics[width=.195\linewidth]{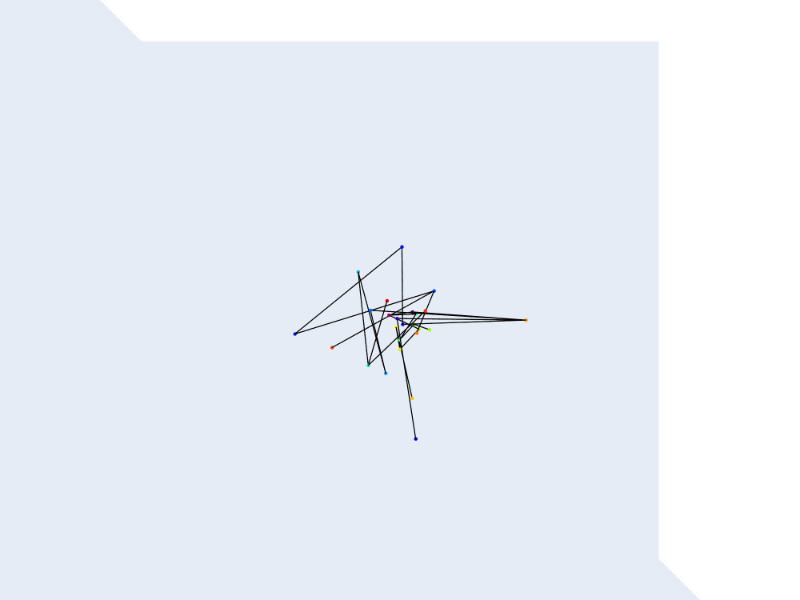}
         \includegraphics[width=.195\linewidth]{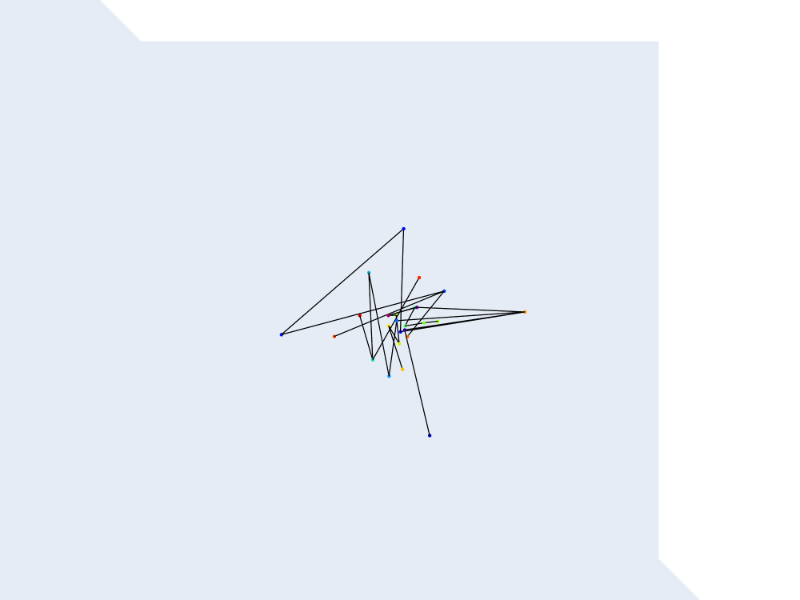}
         \caption{UNet}
     \end{subfigure}

     \begin{subfigure}{\linewidth}
         \centering
         \includegraphics[width=.195\linewidth]{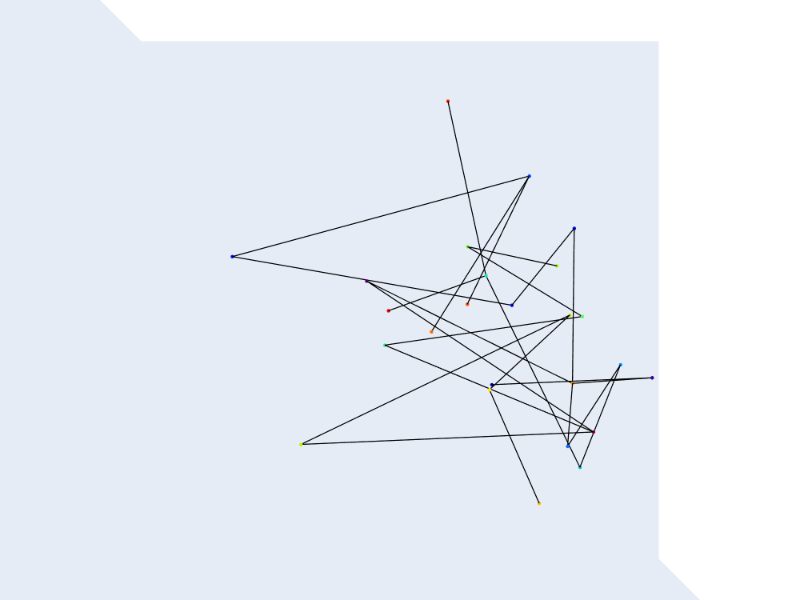}
         \includegraphics[width=.195\linewidth]{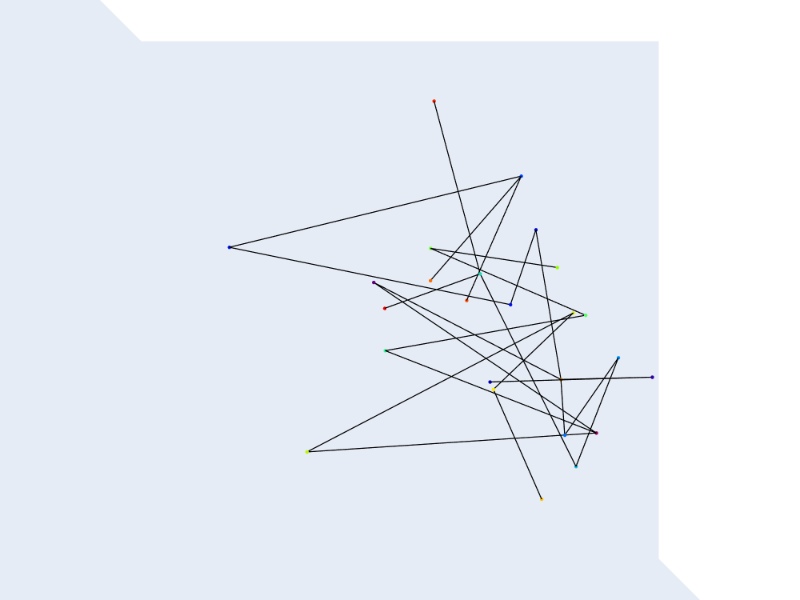}
         \includegraphics[width=.195\linewidth]{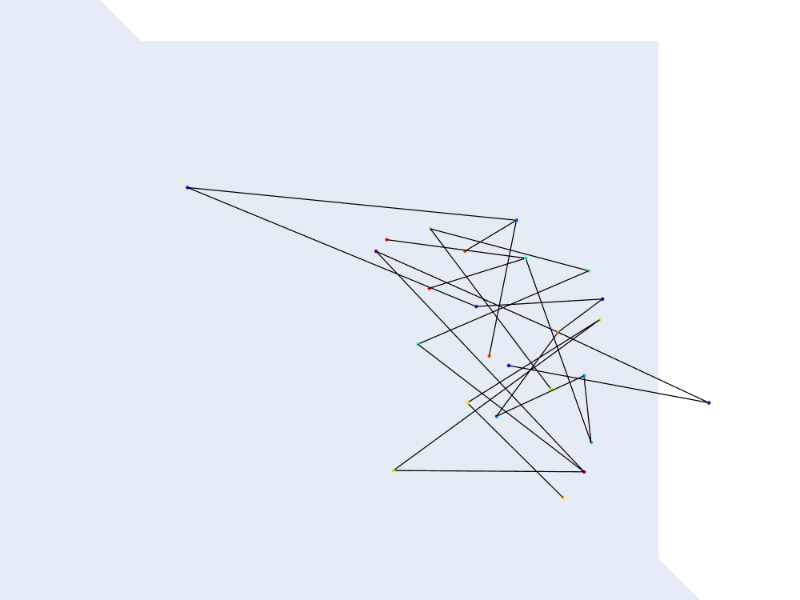}
         \includegraphics[width=.195\linewidth]{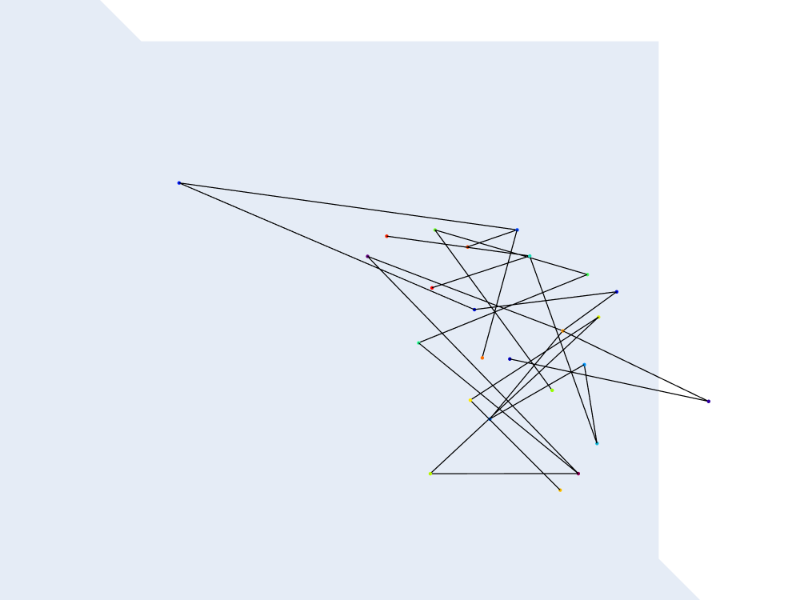}
         \includegraphics[width=.195\linewidth]{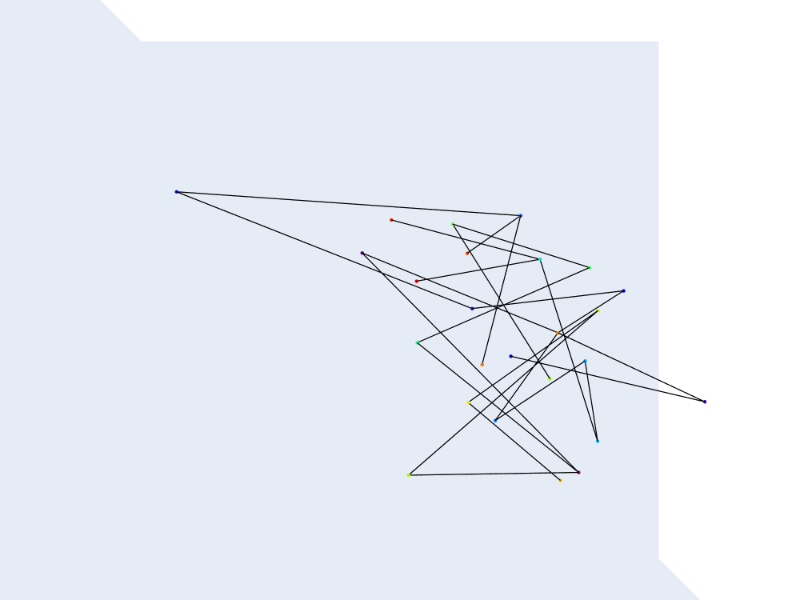}
         \caption{ResNet}
     \end{subfigure}

    \begin{subfigure}{\linewidth}
        \centering
        \includegraphics[width=.195\linewidth]{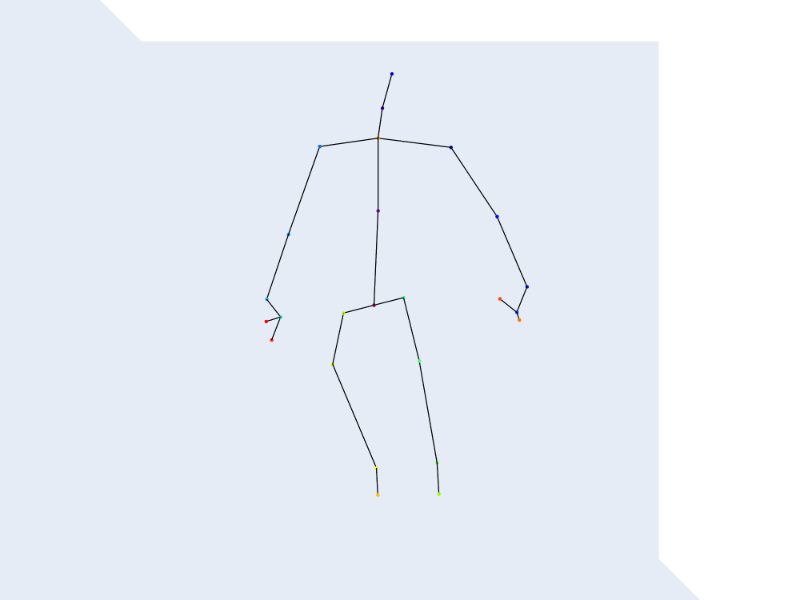}
        \includegraphics[width=.195\linewidth]{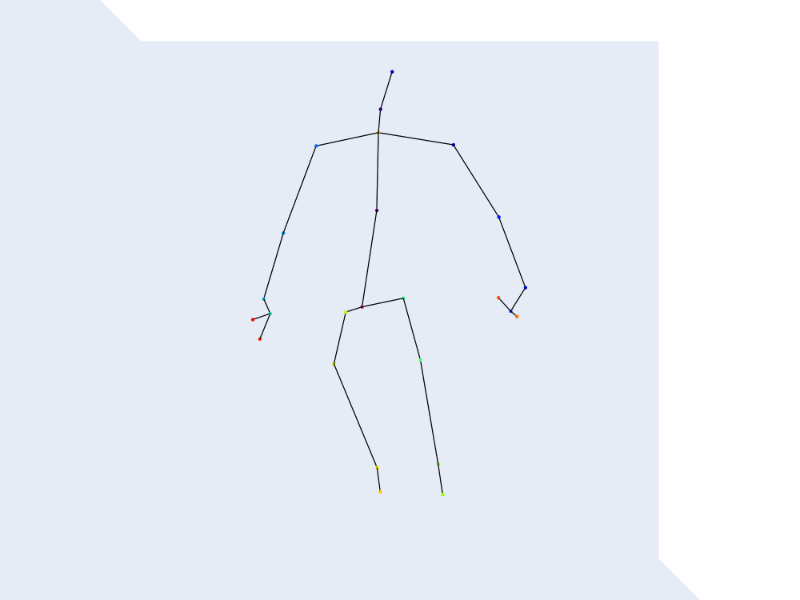}
        \includegraphics[width=.195\linewidth]{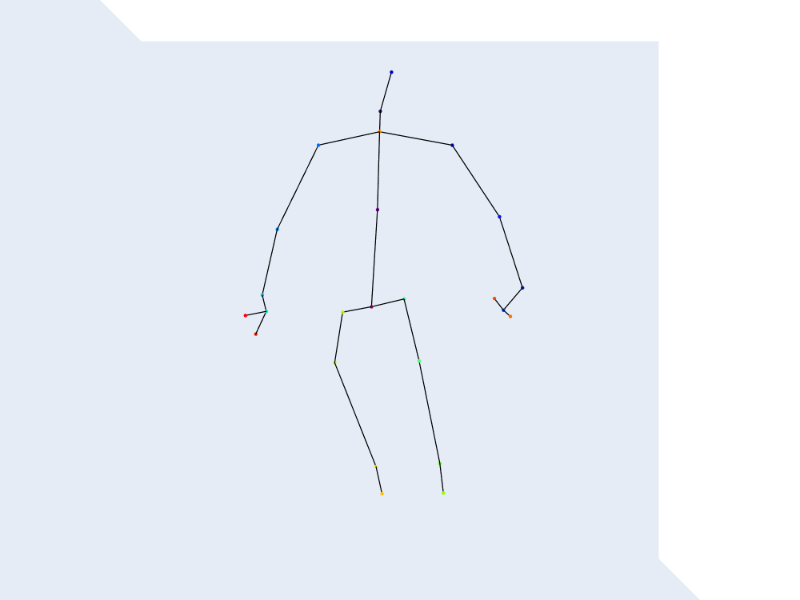}
        \includegraphics[width=.195\linewidth]{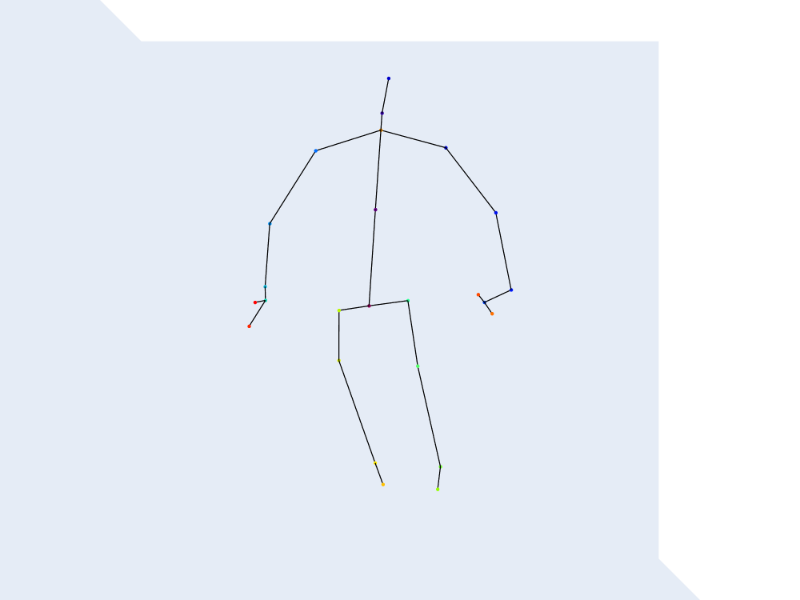}
        \includegraphics[width=.195\linewidth]{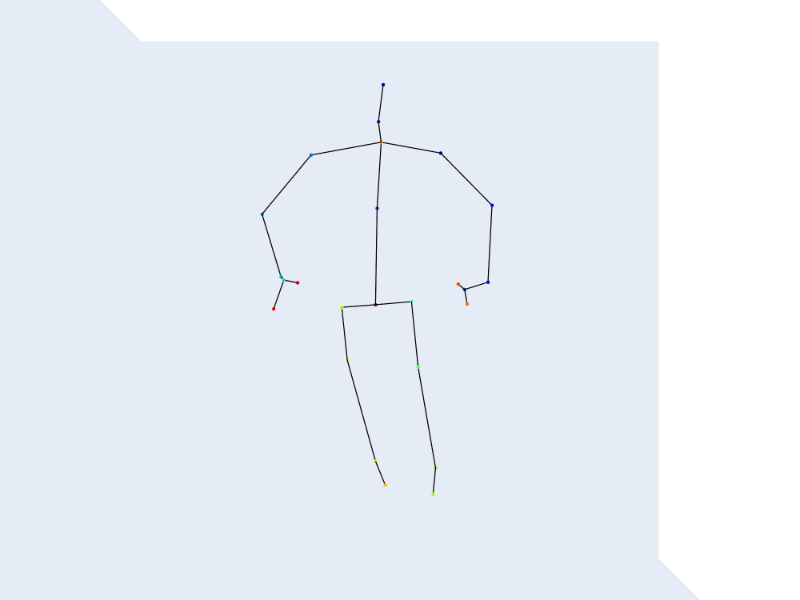}
        \caption{DMR}
    \end{subfigure}

    \begin{subfigure}{\linewidth}
        \centering
        \includegraphics[width=.195\linewidth]{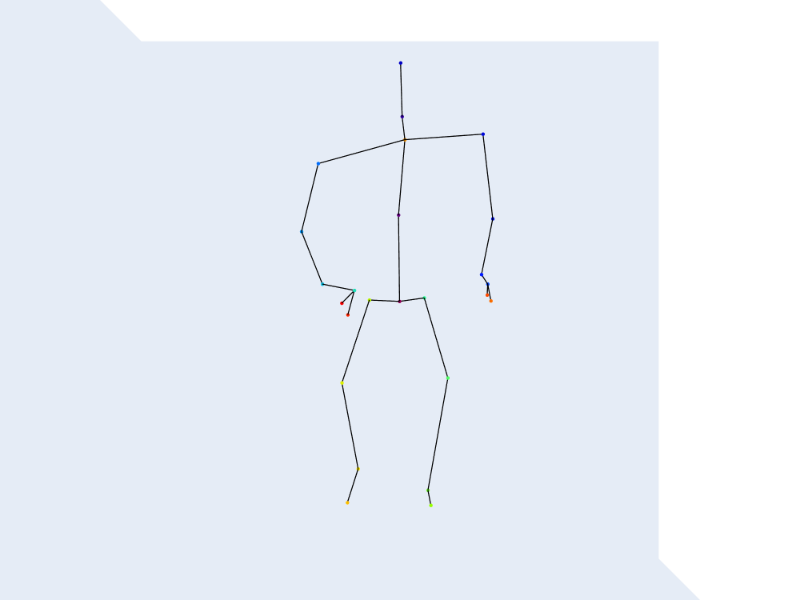}
        \includegraphics[width=.195\linewidth]{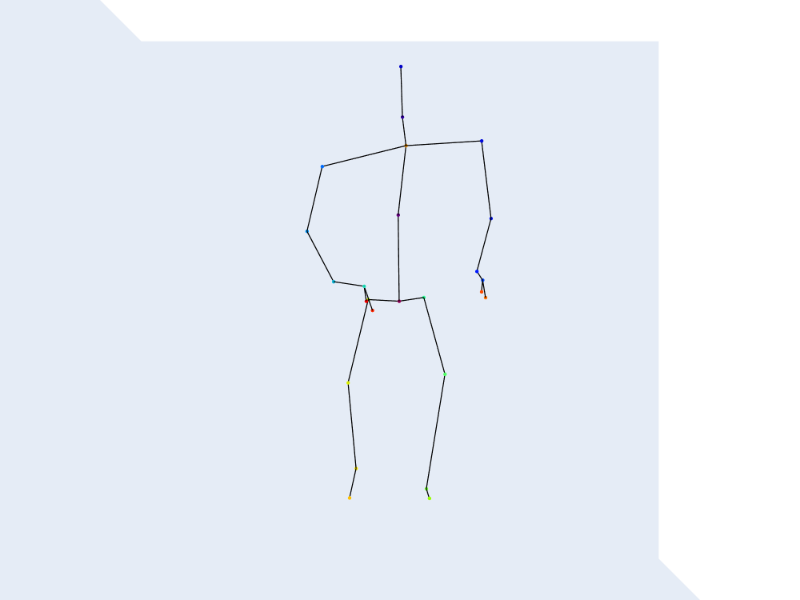}
        \includegraphics[width=.195\linewidth]{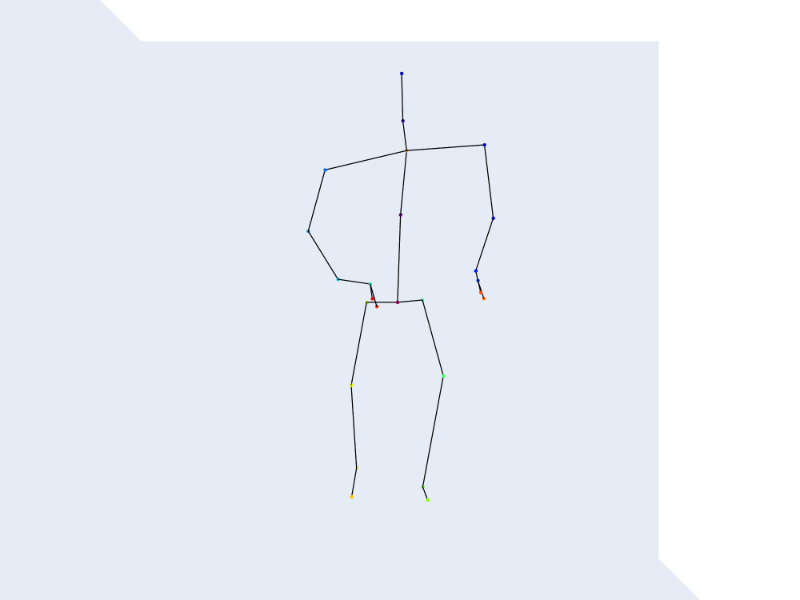}
        \includegraphics[width=.195\linewidth]{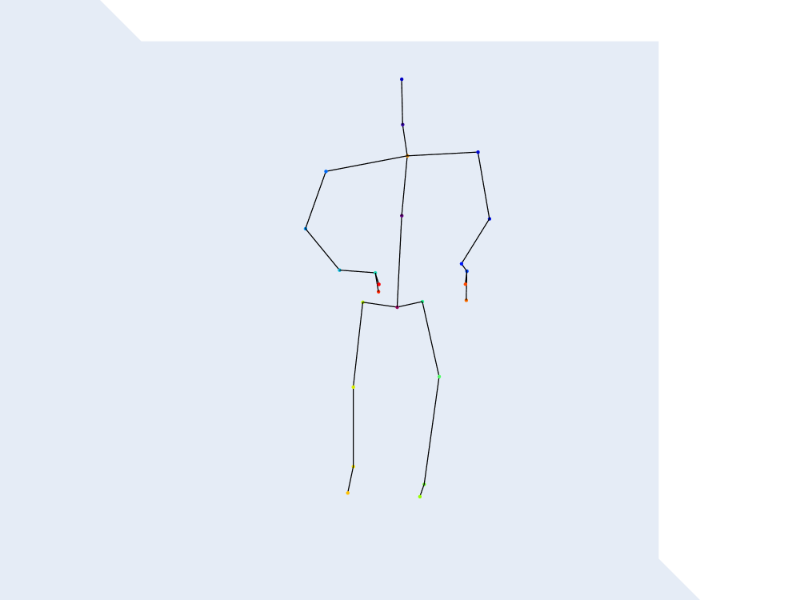}
        \includegraphics[width=.195\linewidth]{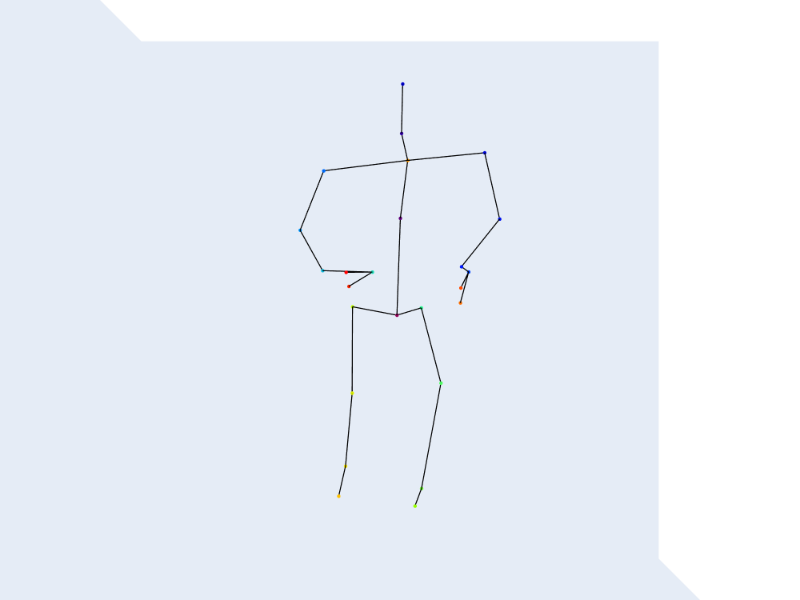}
        \caption{PMR}
    \end{subfigure}

    \caption{Example visualization: Actor 1 (Male) performing the ``Drink Water" action.}
    \label{fig:vis2}
\end{figure*}

\begin{figure*}[!htb]
    \centering

    \begin{subfigure}{\linewidth}
        \centering
        \includegraphics[width=.195\linewidth]{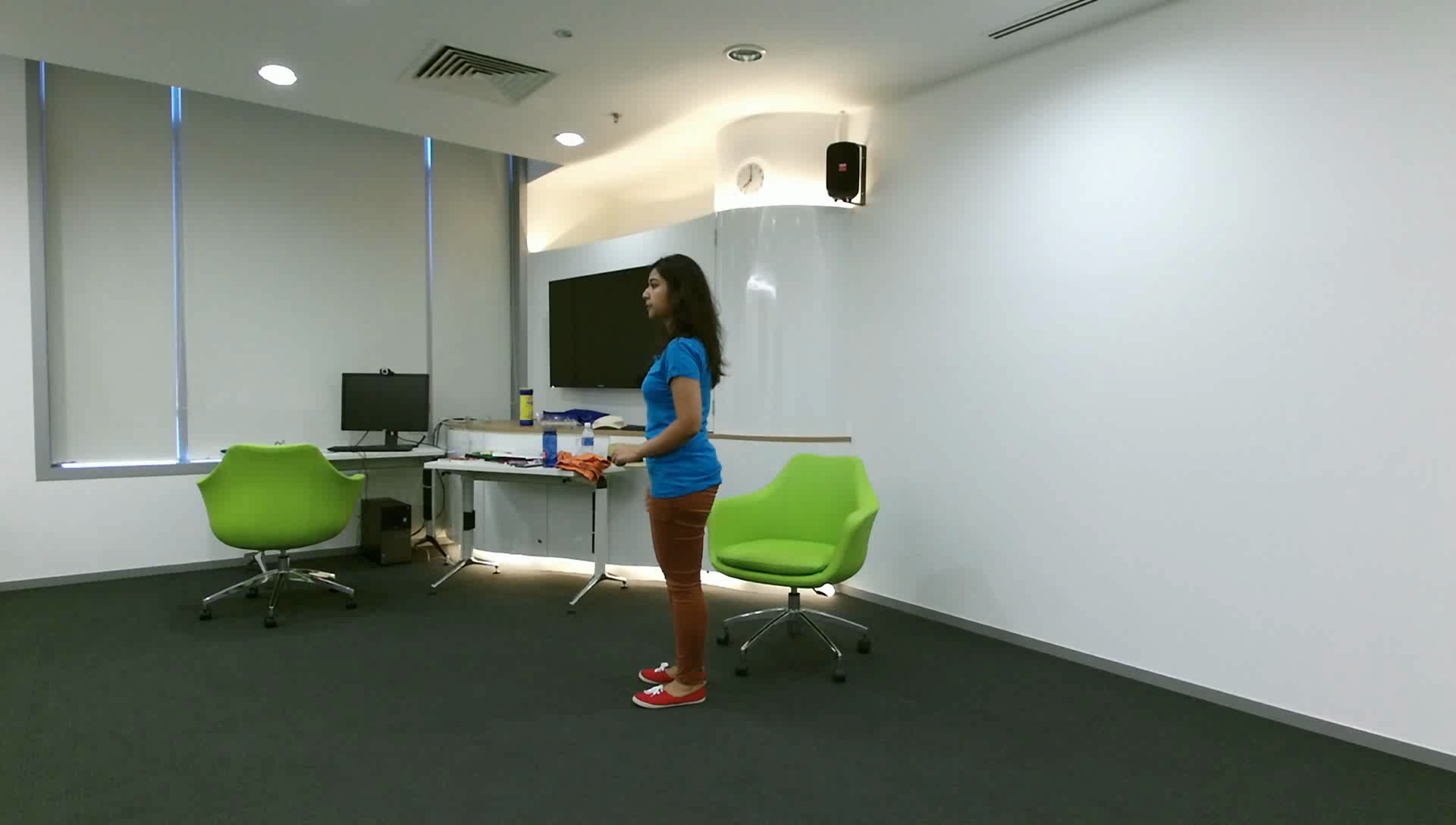}
        \includegraphics[width=.195\linewidth]{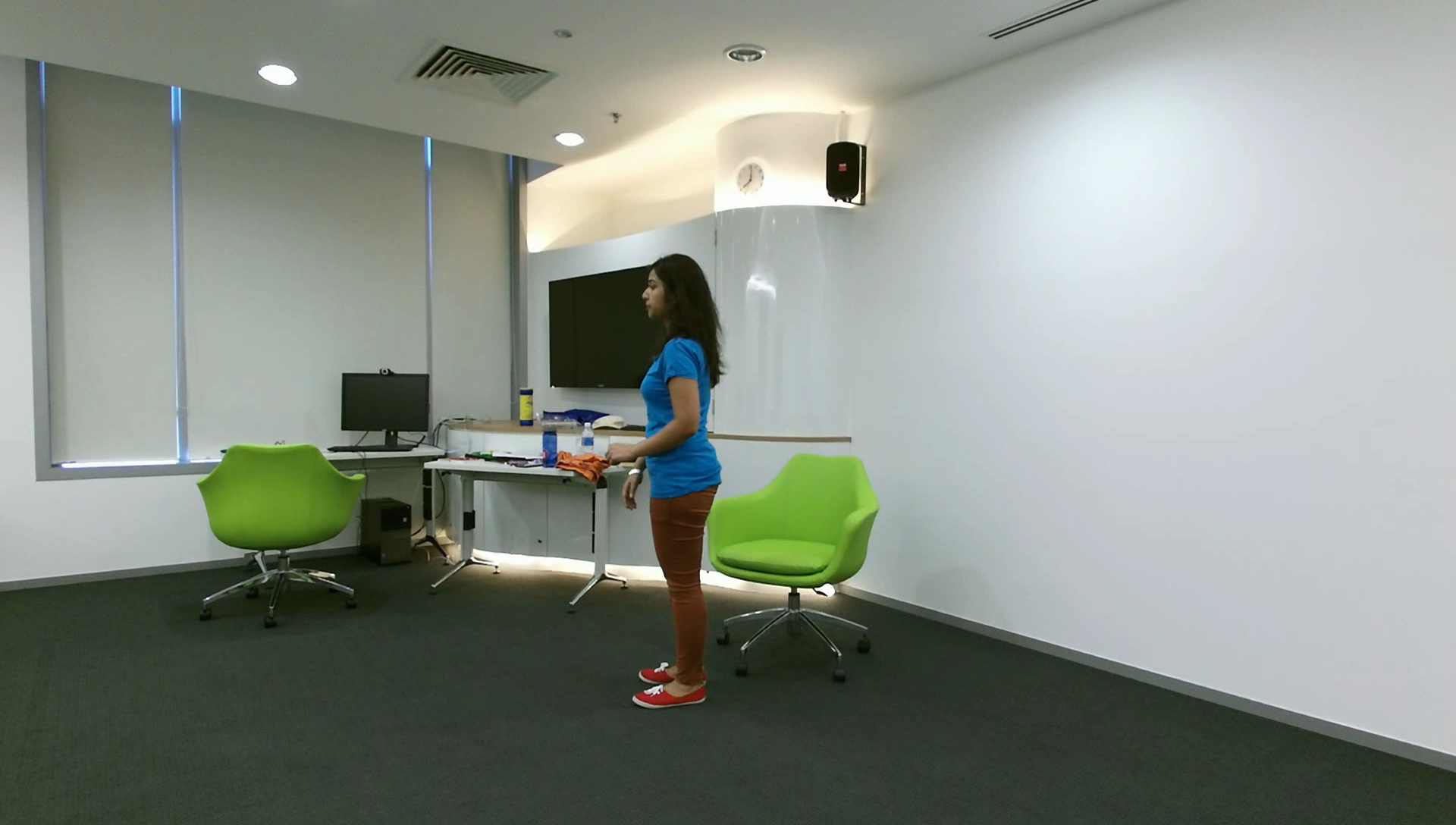}
        \includegraphics[width=.195\linewidth]{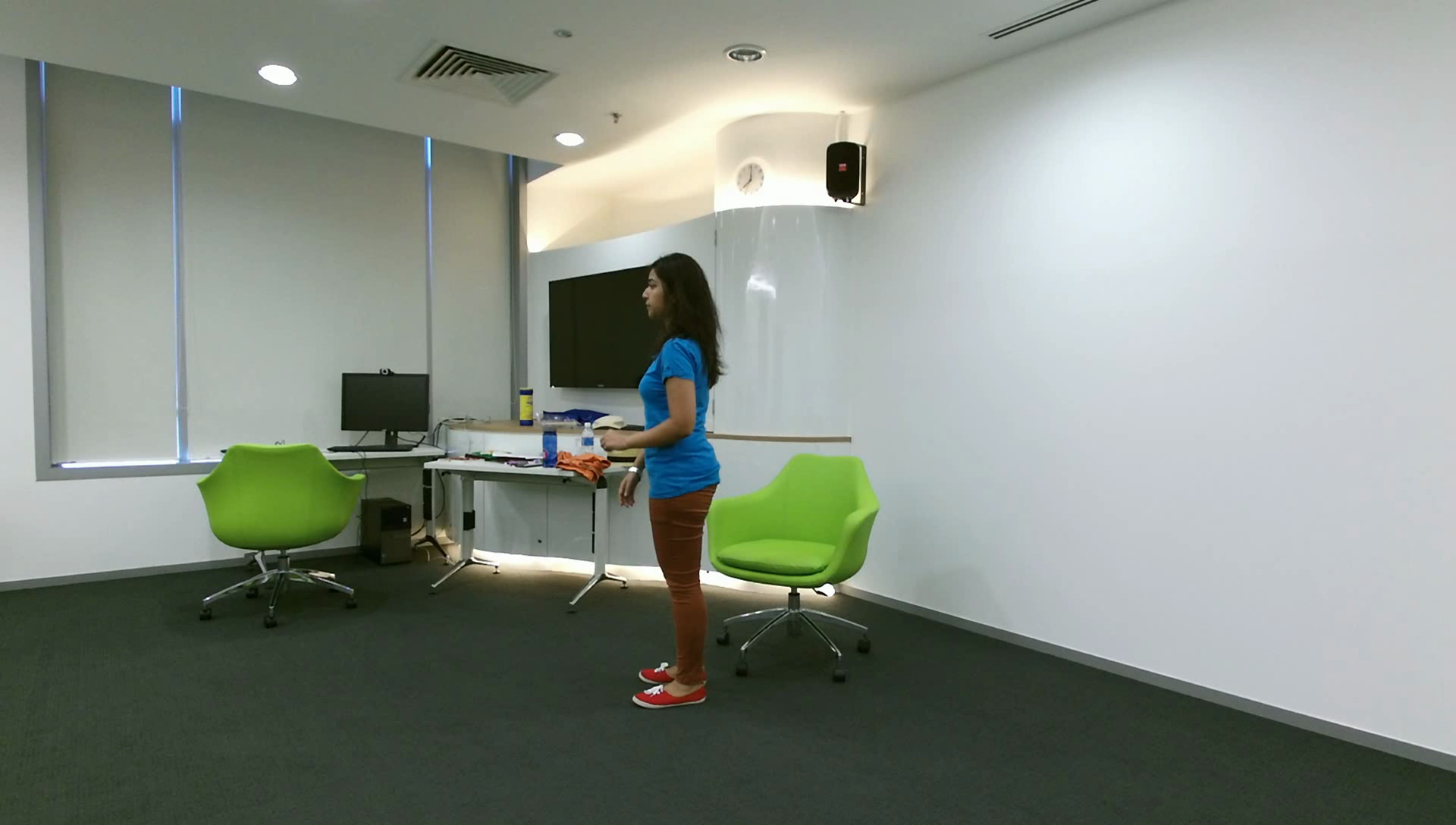}
        \includegraphics[width=.195\linewidth]{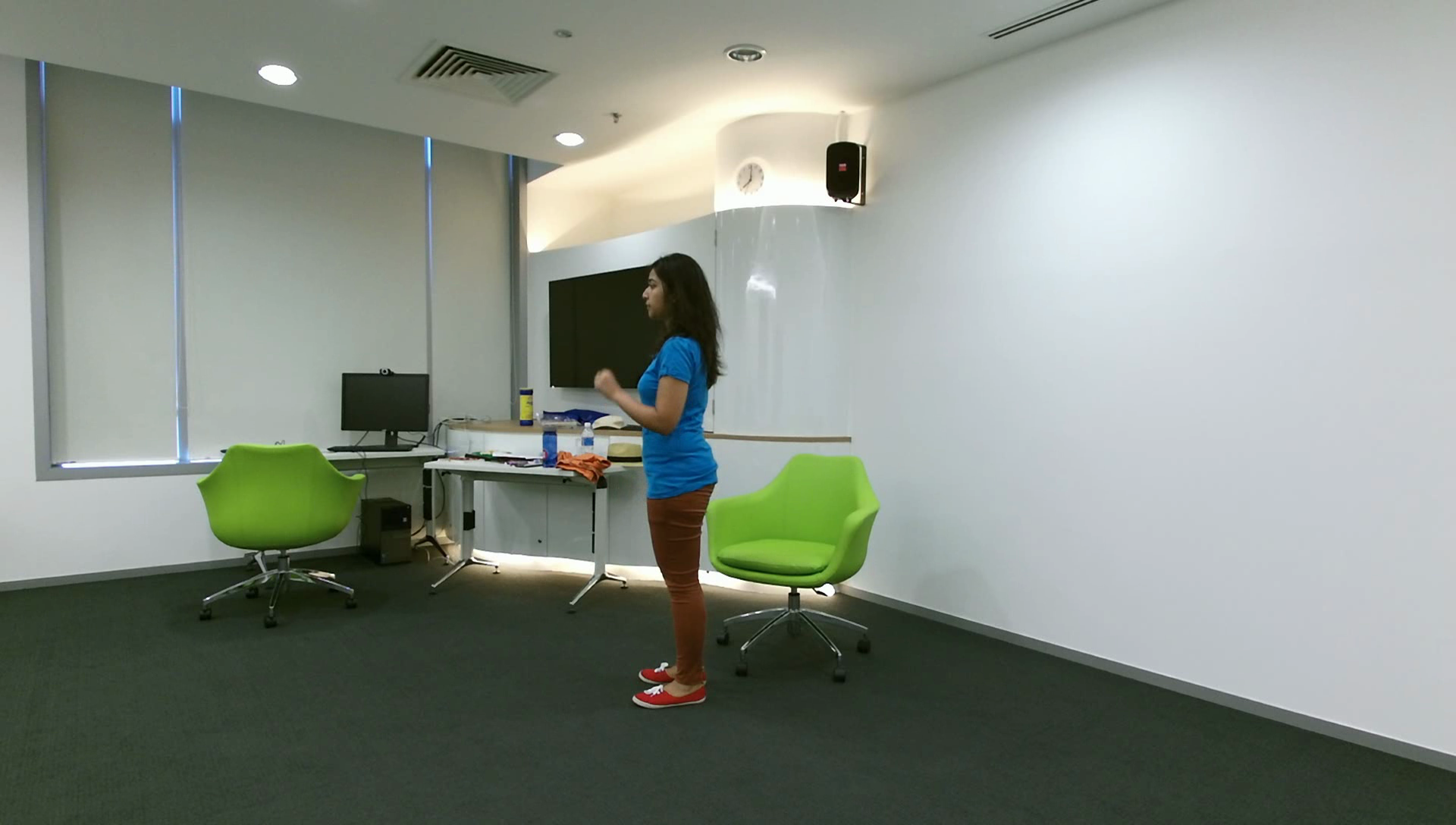}
        \includegraphics[width=.195\linewidth]{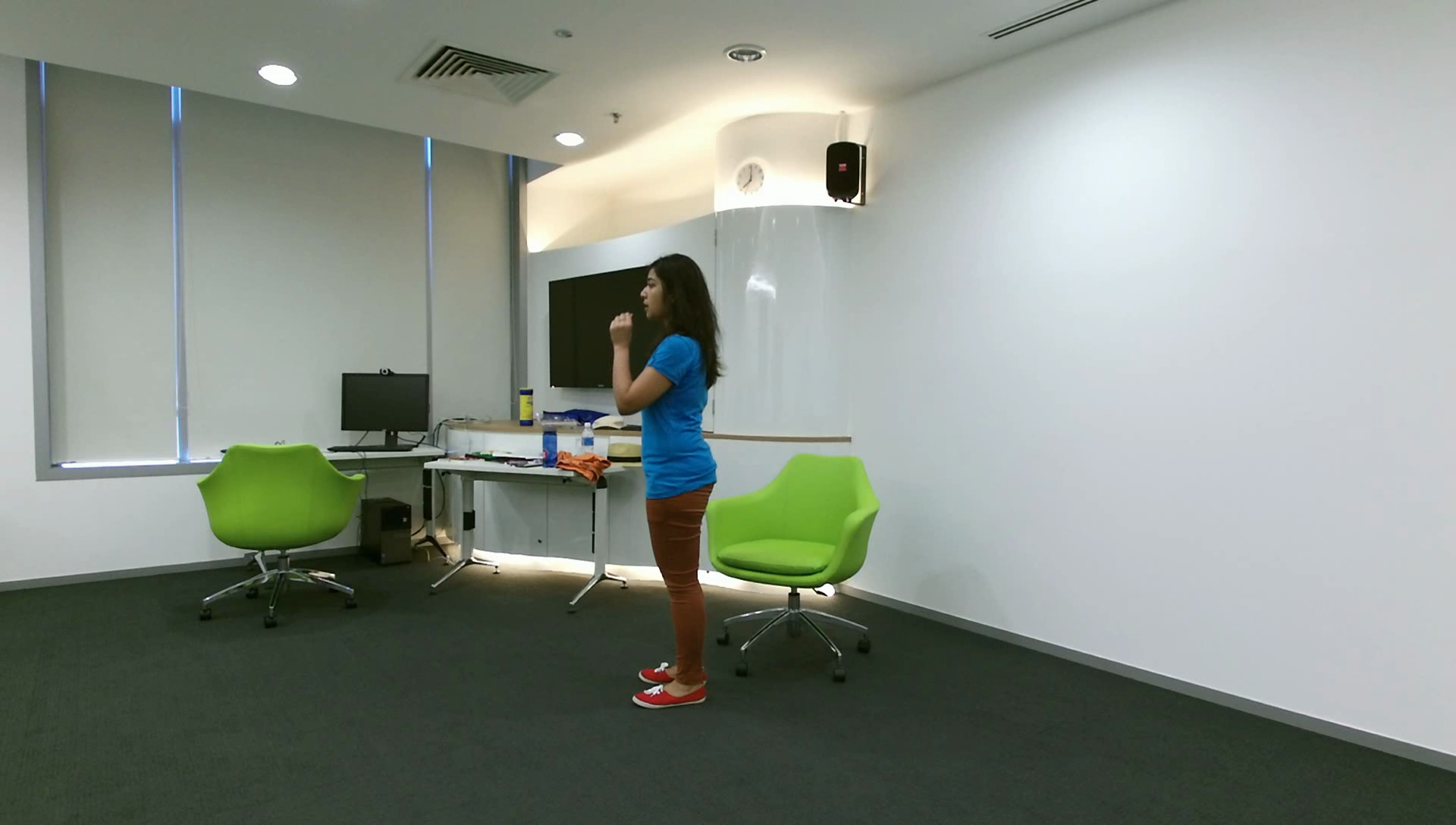}
        \caption{Original RGB Video}
    \end{subfigure}

    \begin{subfigure}{\linewidth}
        \centering
        \includegraphics[width=.195\linewidth]{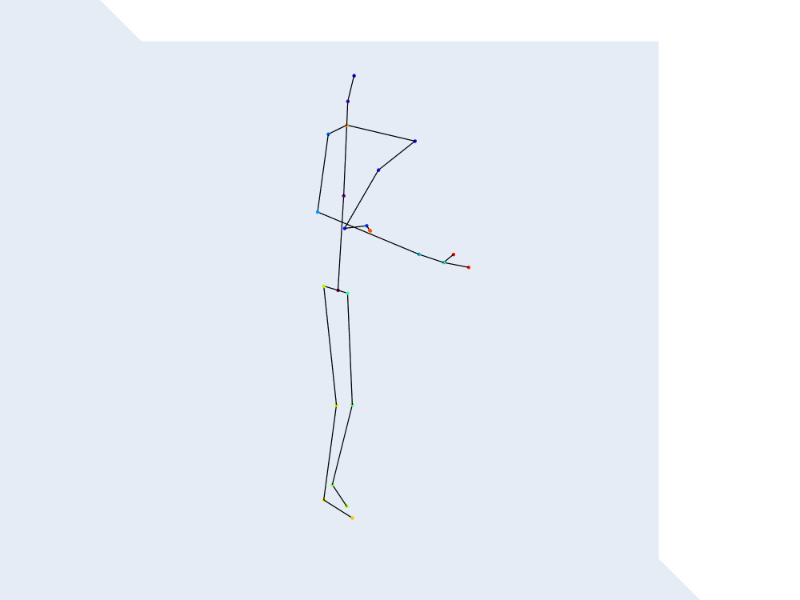}
        \includegraphics[width=.195\linewidth]{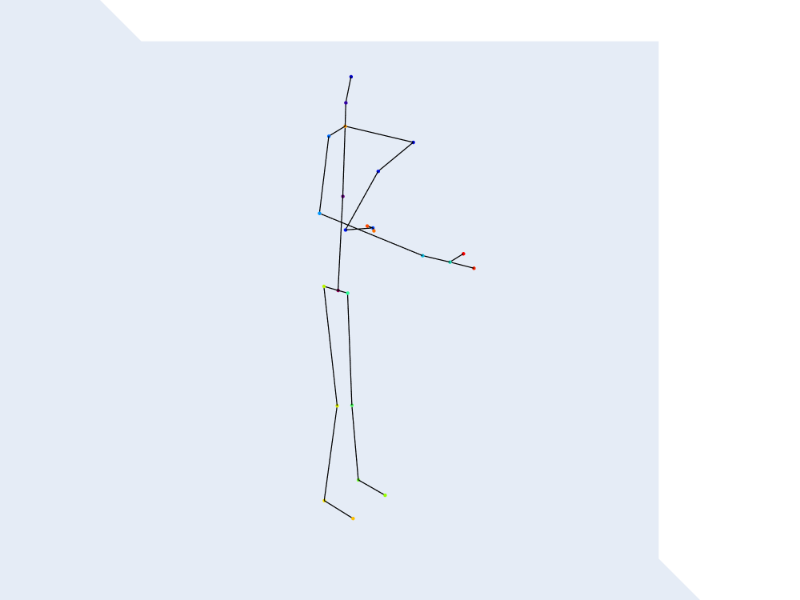}
        \includegraphics[width=.195\linewidth]{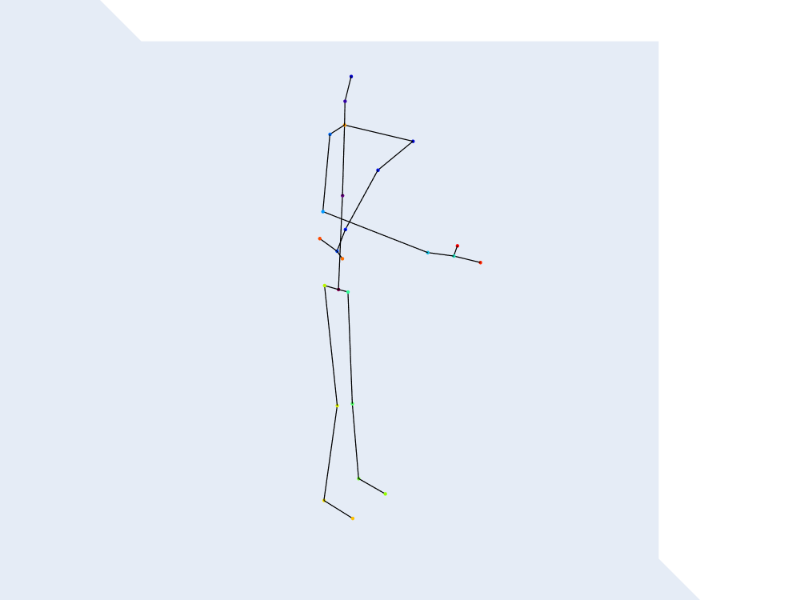}
        \includegraphics[width=.195\linewidth]{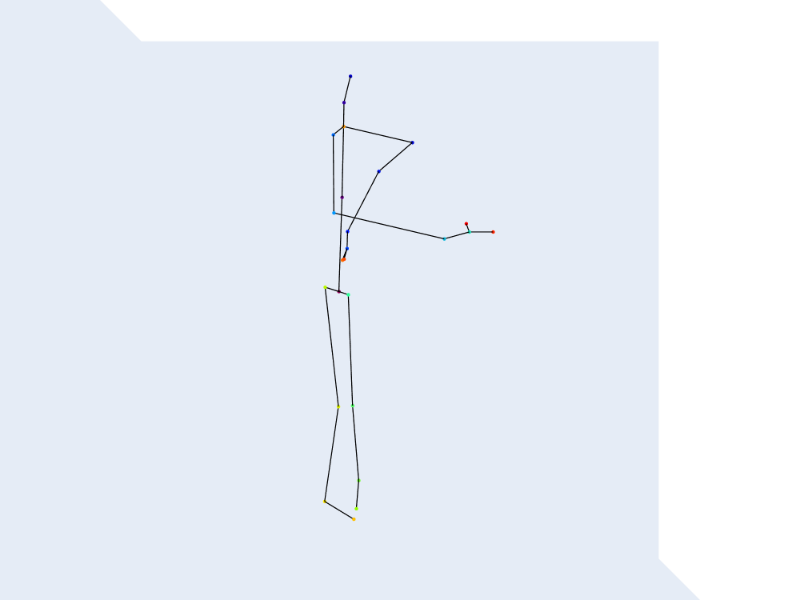}
        \includegraphics[width=.195\linewidth]{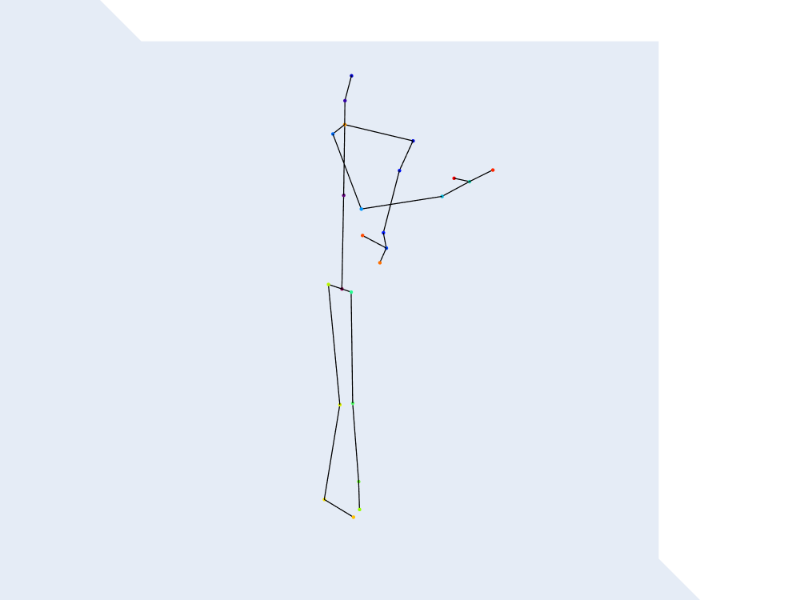}
        \caption{Original}
    \end{subfigure}

     \begin{subfigure}{\linewidth}
         \centering
         \includegraphics[width=.195\linewidth]{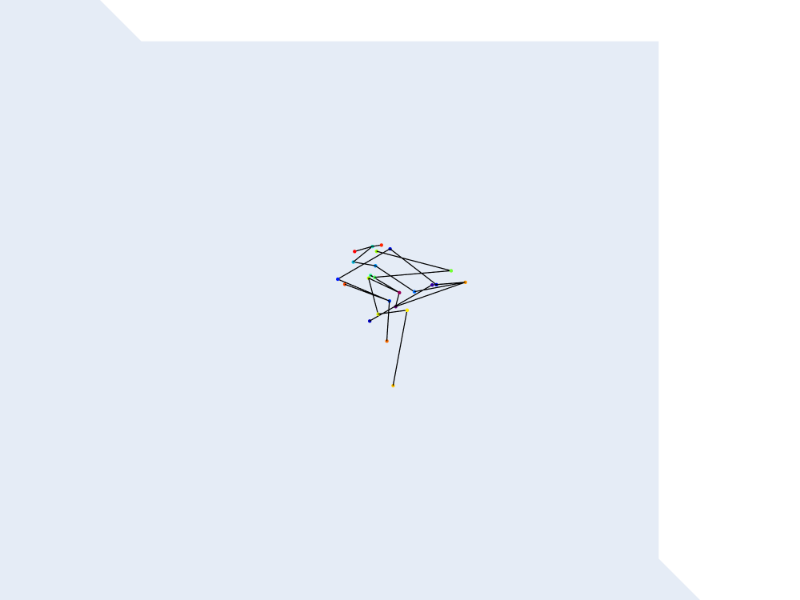}
         \includegraphics[width=.195\linewidth]{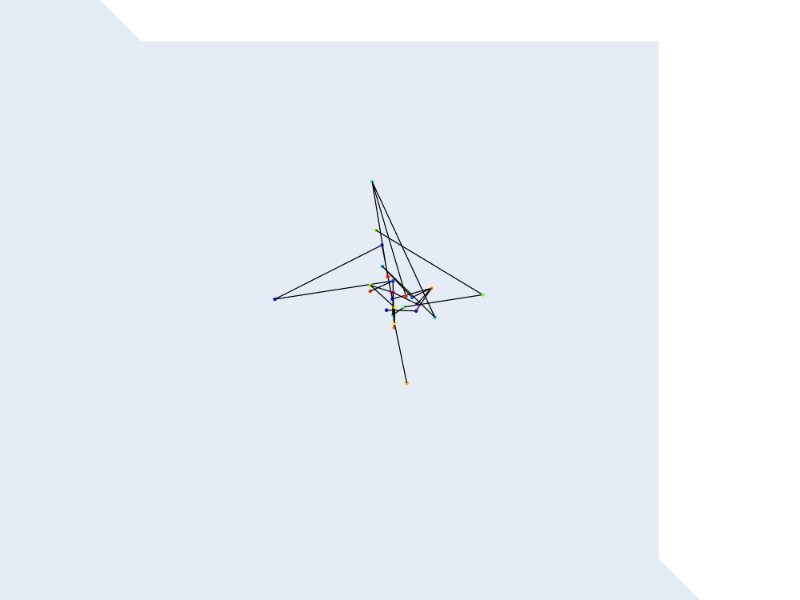}
         \includegraphics[width=.195\linewidth]{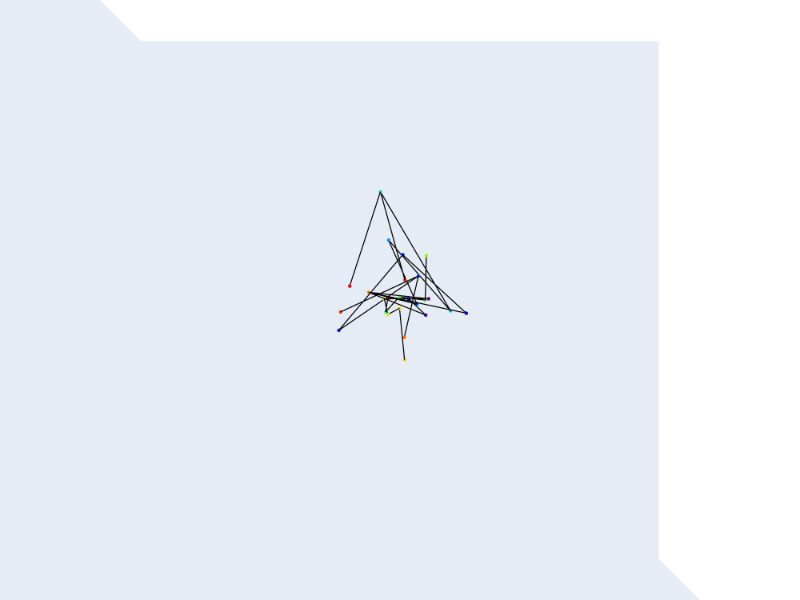}
         \includegraphics[width=.195\linewidth]{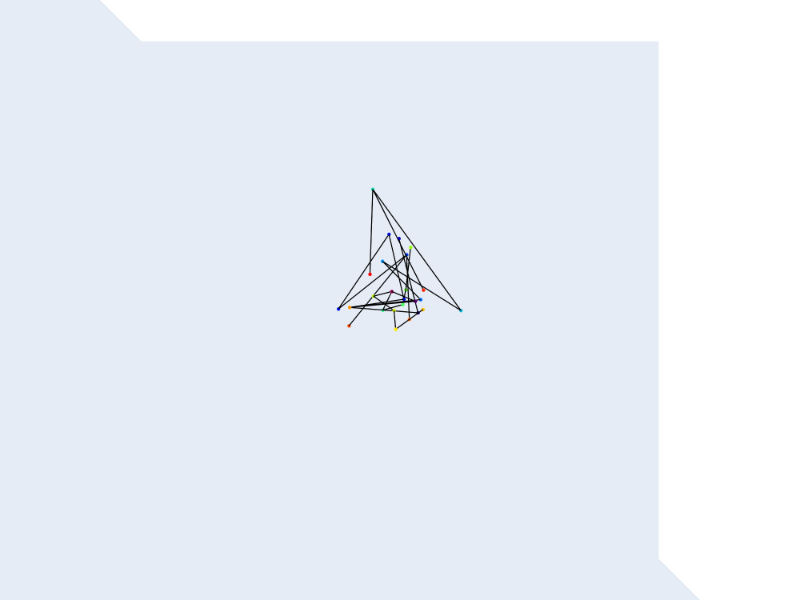}
         \includegraphics[width=.195\linewidth]{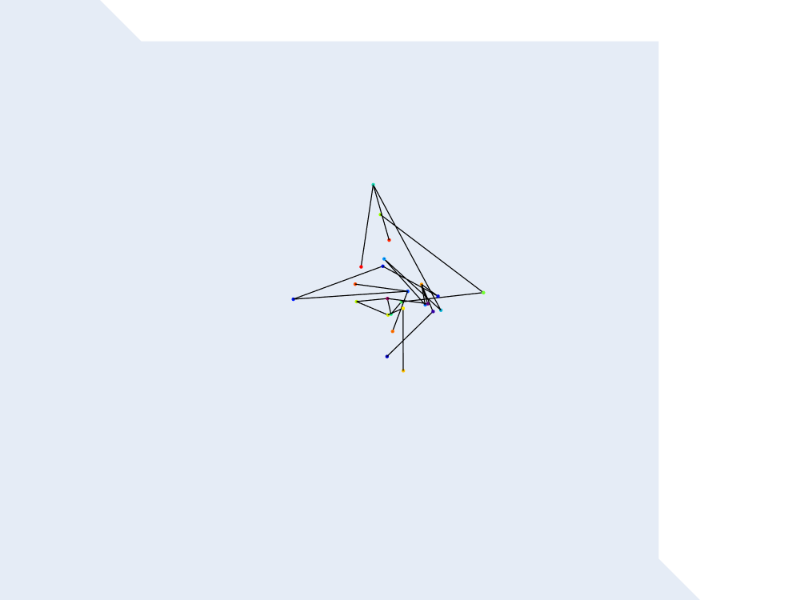}
         \caption{UNet}
     \end{subfigure}

     \begin{subfigure}{\linewidth}
         \centering
         \includegraphics[width=.195\linewidth]{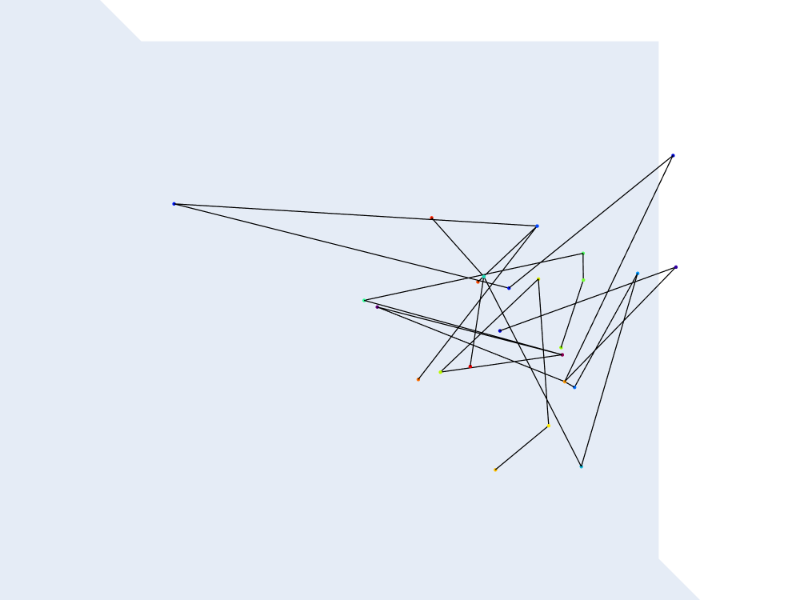}
         \includegraphics[width=.195\linewidth]{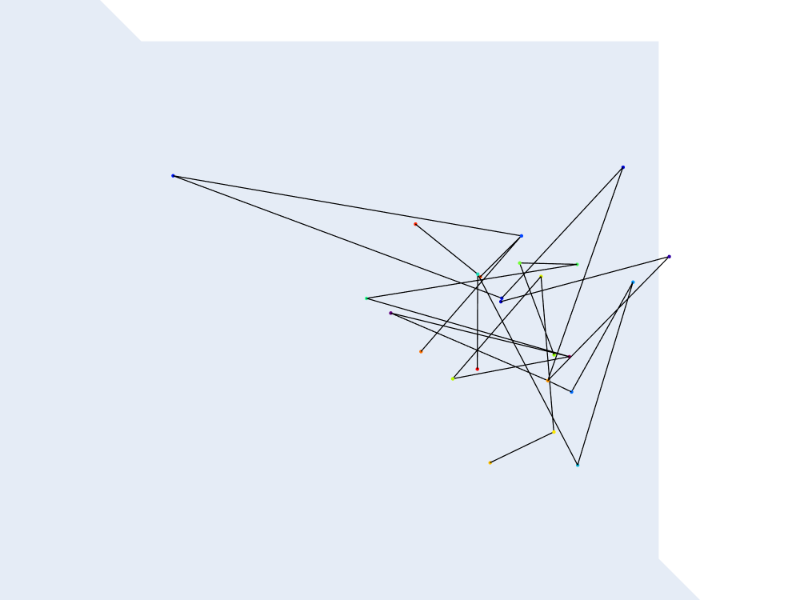}
         \includegraphics[width=.195\linewidth]{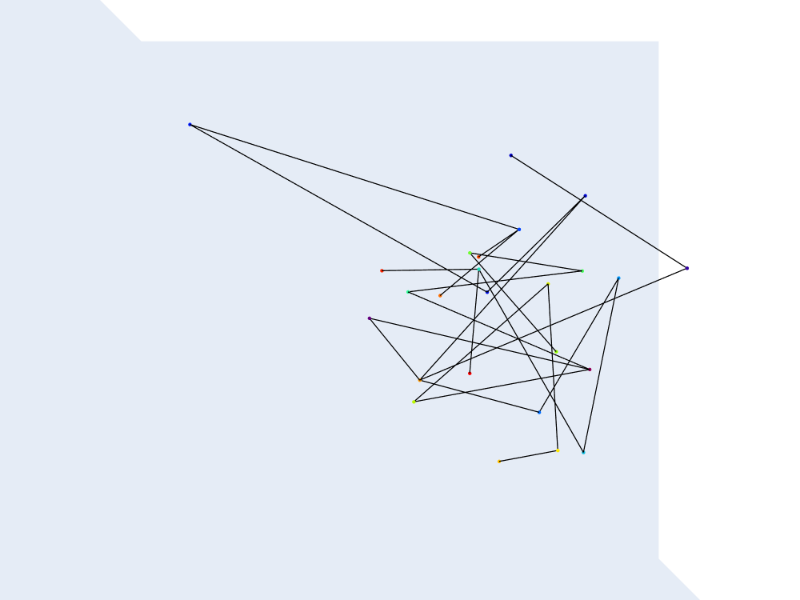}
         \includegraphics[width=.195\linewidth]{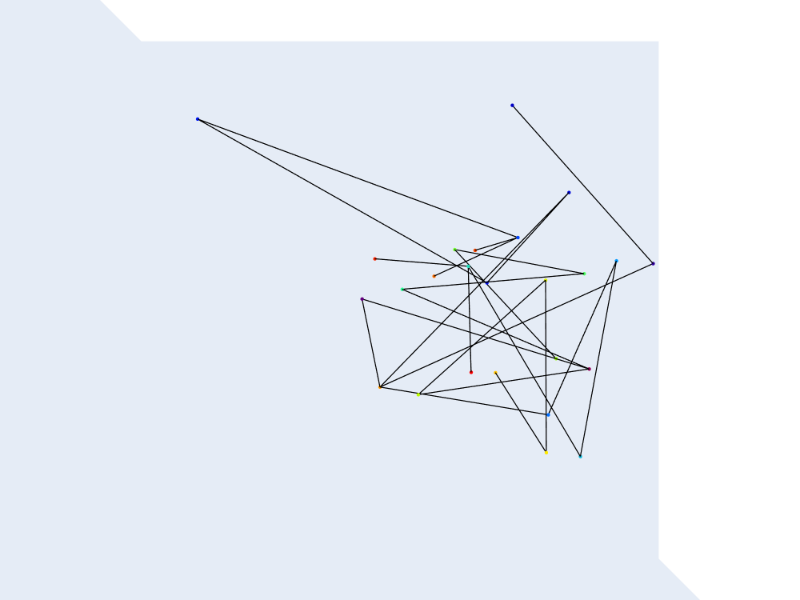}
         \includegraphics[width=.195\linewidth]{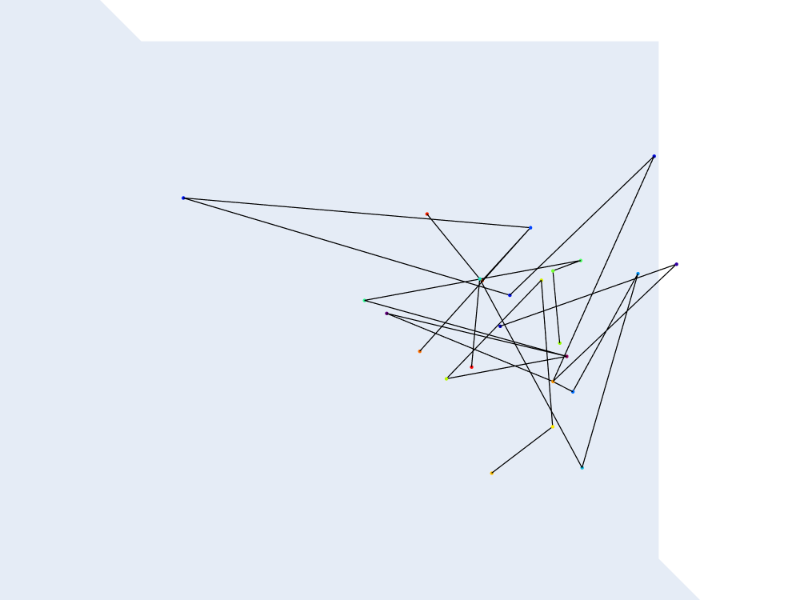}
         \caption{ResNet}
     \end{subfigure}

    \begin{subfigure}{\linewidth}
        \centering
        \includegraphics[width=.195\linewidth]{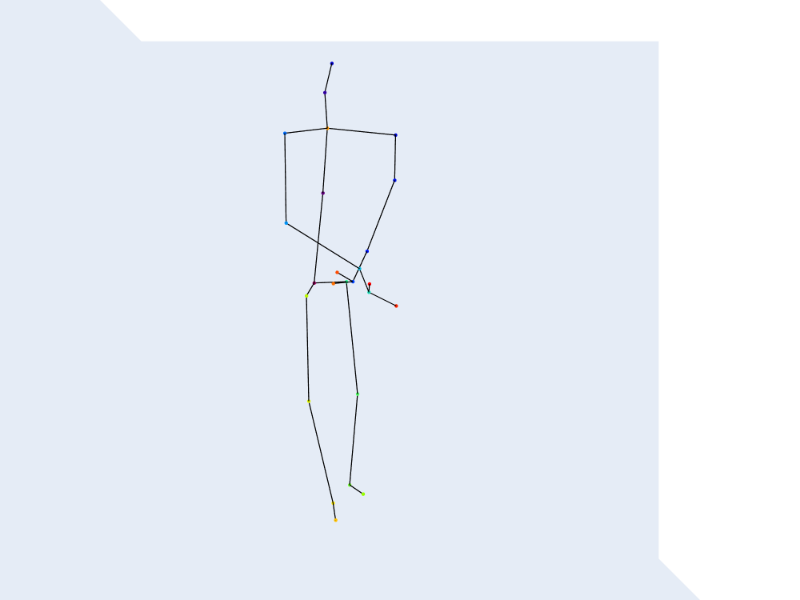}
        \includegraphics[width=.195\linewidth]{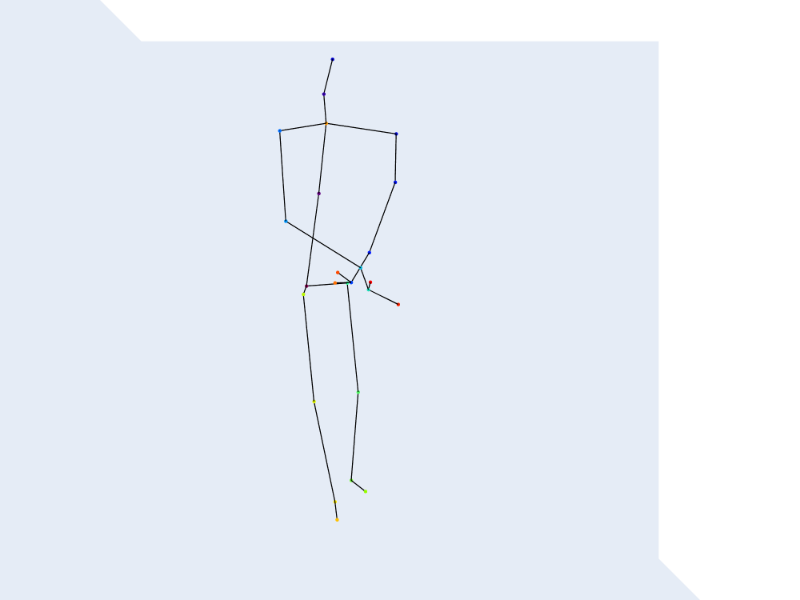}
        \includegraphics[width=.195\linewidth]{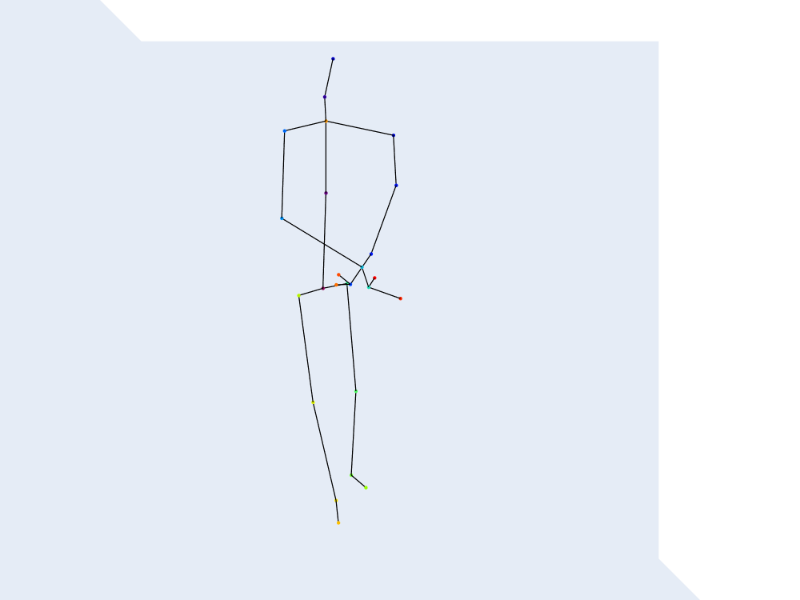}
        \includegraphics[width=.195\linewidth]{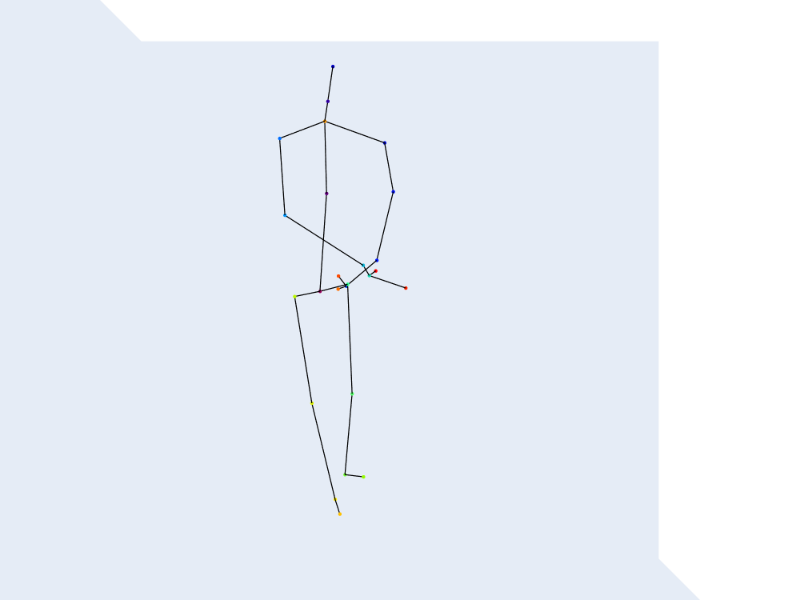}
        \includegraphics[width=.195\linewidth]{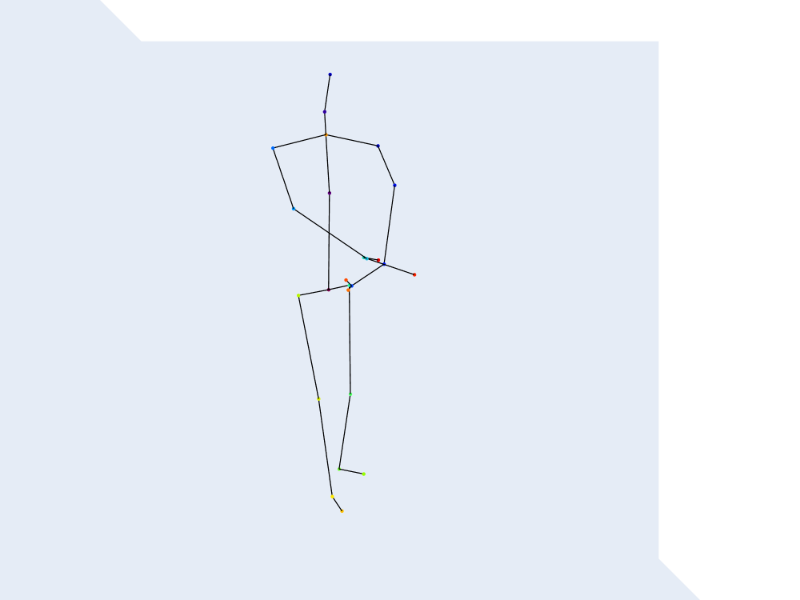}
        \caption{DMR}
    \end{subfigure}

    \begin{subfigure}{\linewidth}
        \centering
        \includegraphics[width=.195\linewidth]{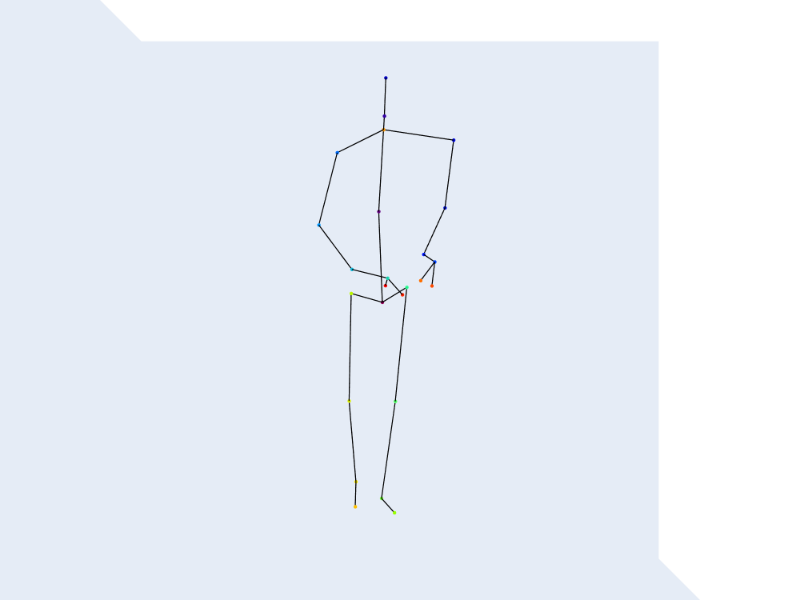}
        \includegraphics[width=.195\linewidth]{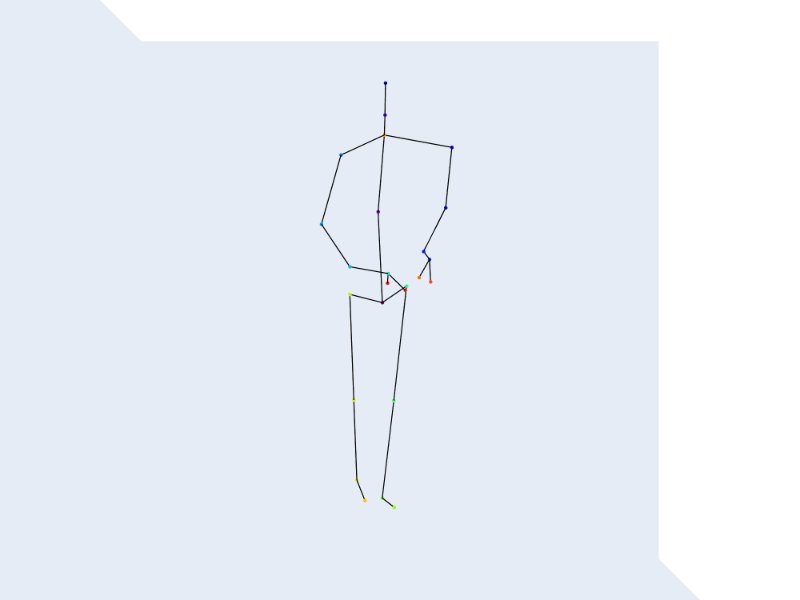}
        \includegraphics[width=.195\linewidth]{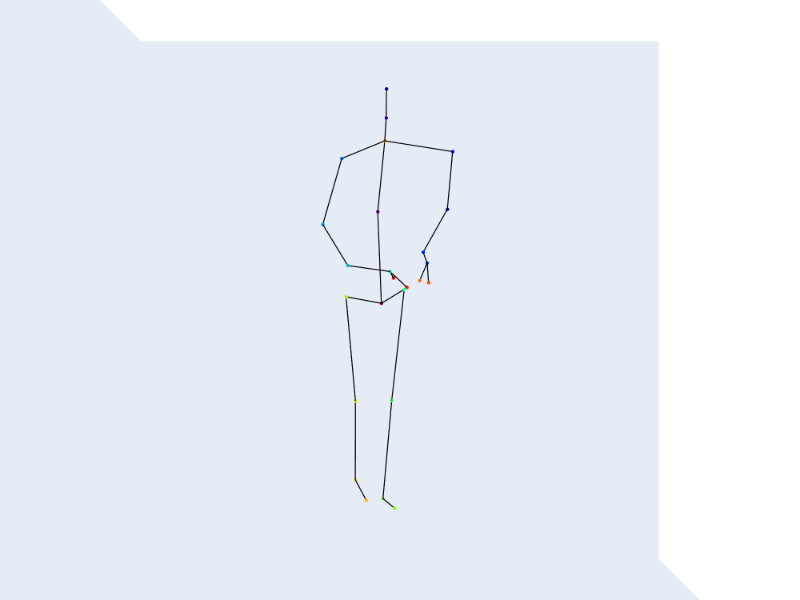}
        \includegraphics[width=.195\linewidth]{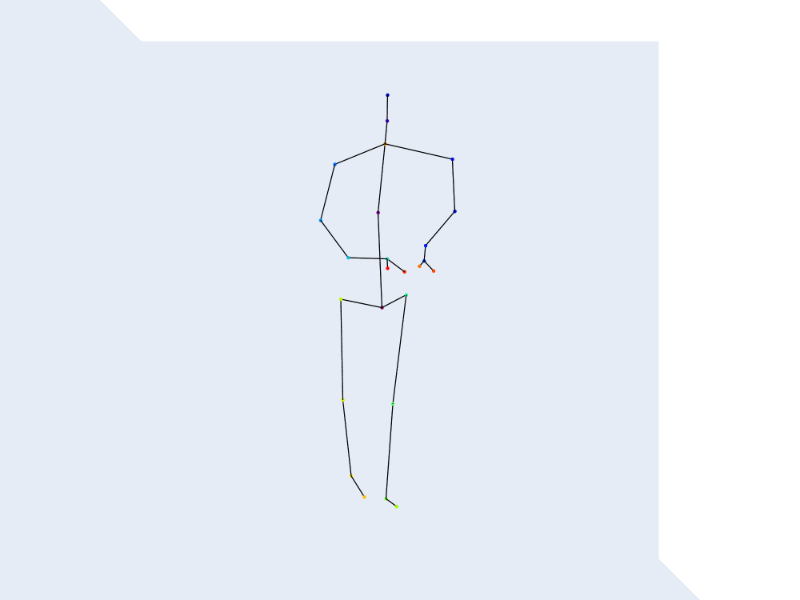}
        \includegraphics[width=.195\linewidth]{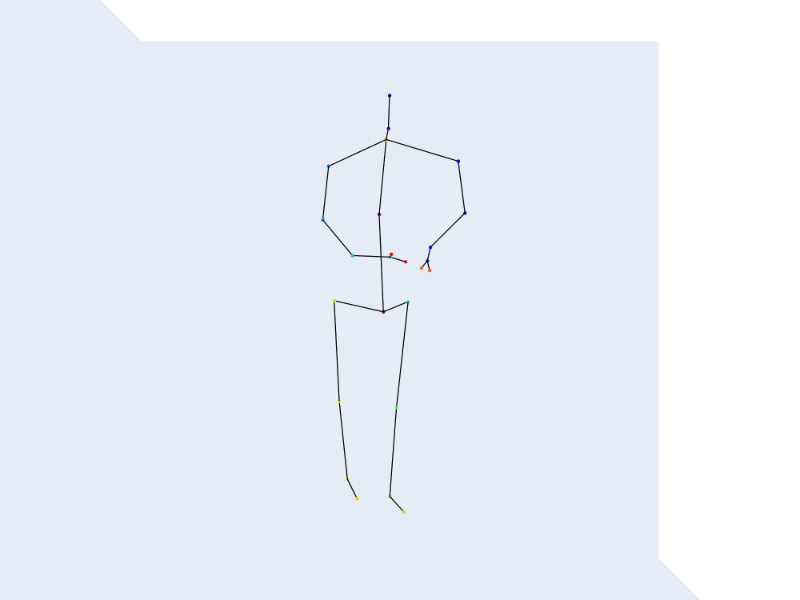}
        \caption{PMR}
    \end{subfigure}

    \caption{Example visualization: Actor 17 (Female) performing the ``Eat Meal" action.} 

    \label{fig:vis3}
\end{figure*}

\end{document}